\documentclass[twoside,11pt]{article}

\usepackage{enumitem}
\usepackage{jmlr2e}

\usepackage{amsmath}

\usepackage{algorithm}
\usepackage{algpseudocode}
\makeatletter
\renewcommand{\ALG@name}{Algorithm} 
\newcommand{\design}{x}

\newcommand{\indexvar}{\lambda}
\newcommand{\Index}{\Lambda}
\newcommand{\resp}{f}
\newcommand{\target}{f^{\ast}}

\usepackage{soul}
\usepackage{xcolor}
\sethlcolor{yellow}

\soulregister\cite7
\soulregister\ref7
\soulregister\eqref7
\usepackage{changes} 

\newcommand{\pa}[1]{{\color{orange} #1}}
\usepackage{subcaption}
\usepackage{booktabs}

\usepackage{lastpage}
\jmlrheading{}{2025}{1-\pageref{LastPage}}{--; Revised --}{--}{21-0000}{Ahadi, Marzban, Adibi and Paynabar}

\ShortHeadings{Min-Max Bayesian Optimization}{Ahadi, Marzban, Adibi and Paynabar}
\firstpageno{1}

\begin{document}

\title{Bayesian Optimization for Function-Valued Responses under Min–Max Criteria}

\author{
    \name Pouya Ahadi \email \href{mailto:pouya.ahadi@gatech.edu}{pouya.ahadi@gatech.edu} \\
    \addr School of Industrial and Systems Engineering\\
    Georgia Institute of Technology\\
    Atlanta, GA 30332, USA
    \AND
    \name Reza Marzban \email \href{mailto:mmarzban3@gatech.edu}{mmarzban3@gatech.edu} \\
    \addr School of Electrical and Computer Engineering\\
    Georgia Institute of Technology\\
    Atlanta, GA 30332, USA
    \AND
    \name Ali Adibi \email \href{mailto: ali.adibi@ece.gatech.edu}{ali.adibi@ece.gatech.edu} \\
    \addr School of Electrical and Computer Engineering\\
    Georgia Institute of Technology\\
    Atlanta, GA 30332, USA
    \AND
    \name Kamran Paynabar\thanks{Corresponding author} \email \href{mailto:kamran.paynabar@isye.gatech.edu}{kamran.paynabar@isye.gatech.edu}  \\
    \addr School of Industrial and Systems Engineering\\
    Georgia Institute of Technology\\
    Atlanta, GA 30332, USA
}



\maketitle

\begin{abstract}
Bayesian optimization is a standard tool for optimizing expensive black-box
functions, yet most existing methods target scalar-valued responses.
In many scientific and engineering applications, the output is functional,
such as curves over time or wavelength, and performance is naturally defined
by worst-case deviation from a target profile rather than an average criterion.
We propose \emph{Min-Max Functional Bayesian Optimization} (MM-FBO), a
framework that directly minimizes the maximum squared deviation of a
functional response from a target function. The response surface is
represented via a Mercer-based Karhunen--Lo\`{e}ve expansion, and
independent Gaussian process models are fitted to the resulting coefficient
processes. Exploiting the posterior deviation's scaled noncentral chi-squared
distribution, we derive closed-form moments yielding a tractable acquisition
function that balances worst-case exploitation and integrated exploration,
without Monte Carlo approximation. We establish theoretical guarantees
including bounds on truncation error, exact posterior reconstruction, the
distribution of the squared deviation, and a discretization bound, together
with a consistency result showing that the acquisition and its minimizers
converge to the true min-max solution. Experiments on synthetic benchmarks
and real-world case studies, including metasurface photonic design and vapor
phase infiltration, demonstrate that MM-FBO achieves lower worst-case regret,
faster convergence, and improved stability compared to scalar-objective
baselines.
\end{abstract}


\section{Introduction}

Bayesian optimization (BO) has become a mainstream strategy for optimizing expensive black box functions, owing to its high data efficiency and principled balance between exploration and exploitation \citep{wang2023survey,Shahriari2016}. The framework has achieved remarkable success across diverse scientific and engineering domains, including aeronautical design, drug discovery, materials science, and automated machine learning \citep{npj2021materialsBO,candelieri2021bo}. At its core, BO relies on probabilistic surrogate models and acquisition functions: \citet{kotthoff2021bo} emphasizes that Gaussian processes are typically employed to capture uncertainty about the objective, with various fast approximations developed to address the cubic computational cost in the number of observations \citep{chokhachian2026fast}, while acquisition functions such as Expected Improvement guide sequential evaluations \citep{candelieri2021bo,Jones1998}. This combination enables BO to identify promising solutions with far fewer evaluations than traditional search methods, making it especially suitable for applications where each experiment or simulation is costly.

Functional responses arise in many modern scientific and engineering problems where the output is not a single scalar but a curve defined over a continuous domain, such as time, wavelength, or angular response. In nanophotonics, for example, a device’s behavior might be fundamentally characterized by its spectral or angular response to an incident optical wavefront, which varies continuously across wavelength or incident angle and encodes rich underlying physics \citep{Yu2014, Kildishev2013, Chen2020Dispersion}. Because these responses must satisfy design objectives over an entire range rather than at a single operating point, inverse-design problems in photonics naturally take the form of functional optimization. Recent hybrid inverse-design frameworks have begun addressing this challenge by learning low-dimensional input-side representations \citep{marzban2025inverse} of the device geometry and then applying global or hybrid optimization methods to search this compressed design space for an optimal design \citep{chen2022high, marzban2025hilab, marzban2025gibs,kudyshev2020machine,radford2025inverse}. These approaches are particularly effective in broadband and multifunctional settings where one seeks a single device that performs well across a continuous range of wavelengths. A common objective in these problems is to optimize a worst case metric such as maximizing the minimum diffraction or transmission efficiency across a target spectral band, which leads naturally to the min–max formulation considered here \citep{choi2024multiwavelength,chen2022high}. Similar robust or worst case formulations have also been central in earlier topology optimized broadband photonic designs \citep{molesky2018inverse, pestourie2018inverse}. Beyond nanophotonics, function-valued outputs arise broadly in materials science, chemical reaction modeling, and complex engineering processes. In these applications, the full functional profile often carries crucial domain-specific information that cannot be captured by a single scalar summary \citep{Conti2010, ren2021reaction}. 

Existing approaches to functional data modeling and optimization often reduce function valued responses to scalar objectives by minimizing integrated deviation, typically in the form of mean squared deviation. While effective for capturing average performance, such scalarization can obscure variability across the functional domain and lead to solutions that are not robust to functional uncertainty. In BO of functional outputs for inverse problems, for example, directly modeling the function over its domain provides clearer guidance than collapsing to a single scalar target \citep{Huang2021Bayesian}. This perspective motivates methods that explicitly account for structure and uncertainty across the entire function.

A central focus of this work is the functional min-max formulation. Given a function $\resp(\design,\indexvar)$ defined over design variables $\design \in \mathcal X$ and functional index $\indexvar \in \Index$, the goal is to find
\[
\design^{\ast} \in \arg\min_{\design \in \mathcal X} \ \max_{\indexvar \in \Index} \ d(\design,\indexvar),
\quad \text{where } d(\design,\indexvar)=\big(\resp(\design,\indexvar)-\target(\indexvar)\big)^{2},
\] 
where $\target(\indexvar)$ is the desired response over the functional domain. This formulation contrasts with classical integrated deviation criteria, which emphasize average performance but may fail to capture the true worst case behavior of functional responses. Recent studies in BO have begun to address such min-max structures. For example, \citet{Weichert2021MinMaxBO} introduced information based strategies to efficiently locate robust optima under min-max objectives, while \citet{Paulson2022ARBO} developed adversarially robust BO for controller auto tuning where the goal is to minimize the maximum loss under uncertainty. More recently, entropy-based approaches \citep{Weichert2024RES} and scalable algorithms for arbitrary uncertain inputs \citep{Yang2023AIRBO} have further demonstrated the importance of directly modeling min-max criteria in Bayesian optimization. These works underscore the distinction between minimizing average deviation and directly targeting the worst case functional response.

Our work lies at the intersection of functional data analysis and BO. Functional data analysis provides a foundation for representing high-dimensional outputs as smooth curves. BO, on the other hand, has become the method of choice for optimizing expensive black box functions by combining Gaussian process surrogates with acquisition functions such as Expected Improvement \citep{Jones1998,Shahriari2016}. Notable examples include emulation of complex computer models with multi-output or dynamic responses \citep{Conti2010} and BO of functional outputs in inverse problems \citep{Huang2021Bayesian}. More recently, hybrid frameworks in nanophotonics have combined generative models (which learn an input-side representation) with BO to efficiently search for high-performance, multifunctional devices \citep{marzban2025hilab,keawmuang2025hybrid}. Our methodology formalizes the specific min-max objective function and provides the specialized acquisition strategy that is often implicit in these applied inverse-design challenges. 

The contributions of this work are threefold. First, we propose Min–Max Functional Bayesian Optimization (MM-FBO), a new formulation of functional BO that directly targets the minimization of the maximum squared deviation across the functional domain, distinguishing it from prior approaches that rely on integrated deviation criteria. Second, we develop a methodology that represents functional responses through a Mercer based expansion and models the resulting coefficient processes using Gaussian process surrogates, enabling the construction of an acquisition function that balances worst case exploitation with integrated exploration. Third, we provide a comprehensive theoretical foundation including bounds 
on truncation error, the distributional structure of the squared deviation, exact posterior reconstruction, a discretization bound for the worst case objective, and a consistency result showing that, as the surrogate becomes 
accurate and uncertainty vanishes, the acquisition converges to the true min-max objective. We then validate MM-FBO through extensive experiments on synthetic benchmarks and physics inspired case studies, demonstrating its value for BO when outputs are inherently functional rather than scalar.

We ground MM-FBO in two physics inspired studies that have function valued responses. First, in electromagnetics we design periodic metasurfaces that shape a scattering spectrum over a dense grid of wavelengths. Each candidate design is parameterized by a small set of geometric coefficients that define a supercell and the performance is the maximum squared deviation from a target spectrum across the wavelength range. This setting reflects common practice in metasurface design where spectra encode rich physical behavior and where dimensionality reduction and manifold ideas are natural tools \citep{Yu2014,Kildishev2013,Chen2020Dispersion, kiarashinejad2020deep, zandehshahvar2022manifold}. Second, in materials processing we study vapor phase infiltration for polymer to inorganic hybrids, a process whose outcome is naturally represented as a function of time or dose. We connect this study to BO of functional outputs in inverse problems \citep{Huang2021Bayesian} and to core references that formalize the chemistry and process mechanisms of vapor phase infiltration and sequential infiltration synthesis \citep{LengLosego2017,Waldman2019,Cara2021,ren2021reaction}.


\section{Problem Formulation}
\label{sec:formulation}

Many optimization tasks in science and engineering are characterized not by 
scalar outcomes but by functional responses. For example, in photonic design 
one may wish to match a desired scattering spectrum across the visible 
wavelength range; in electrochemistry the performance of a battery is observed 
as a capacity profile over time or cycles; and in energy harvesting, radiative 
cooling materials are evaluated by their emissivity curves across a band of 
wavelengths. Unlike scalar-valued objectives, these responses are continuous 
functions of an auxiliary index, and the optimization task must account for 
discrepancies over the entire functional domain rather than at a single point. 
It is not sufficient for a design to perform well on average if it fails in 
even a small region of the domain. These examples motivate the need for 
optimization criteria that go beyond average deviation minimization.

Formally, let $\mathcal{X} \subset \mathbb{R}^{d}$ denote the compact design 
space and let $\Lambda$ denote the compact functional domain. For each 
$x \in \mathcal{X}$ and $\lambda \in \Lambda$, the system returns a response 
$f:\mathcal{X} \times \Lambda \to \mathbb{R}$ defined by 
$(x,\lambda) \mapsto f(x,\lambda)$. Given a target function 
$f^{*}:\Lambda \to \mathbb{R}$, we define the pointwise squared deviation
\begin{equation}
\label{eq:deviation}
d(x,\lambda) 
\;=\; 
\bigl(f(x,\lambda) - f^{*}(\lambda)\bigr)^{2}, 
\qquad 
x \in \mathcal{X},\; \lambda \in \Lambda.
\end{equation}

Two standard scalarizations of~\eqref{eq:deviation} are the integrated 
deviation and the worst-case deviation. The integrated deviation averages 
discrepancies across the domain,
\[
\mathcal{I}(x) 
\;=\; 
\int_{\Lambda} d(x,\lambda) \, d\lambda,
\]
and rewards good average performance but can tolerate large local deviations. 
By contrast, the worst-case deviation focuses on the maximum discrepancy,
\begin{equation}
\label{eq:objective}
g(x) 
\;=\; 
\sup_{\lambda \in \Lambda} d(x,\lambda) 
\;=\; 
\sup_{\lambda \in \Lambda} 
\bigl(f(x,\lambda) - f^{*}(\lambda)\bigr)^{2},
\end{equation}
which requires uniform fidelity across the functional domain and penalizes any 
region with substantial departure from the target. 

We assume $f(x,\cdot)$ is continuous and square-integrable on $\Lambda$ for 
each $x \in \mathcal{X}$, and that both $\mathcal{X}$ and $\Lambda$ are 
compact. These assumptions ensure well-posedness of the optimization problem 
and enable the functional-analytic tools developed in Section~\ref{sec:method}.

Our goal is to solve the min-max problem
\begin{equation}
\label{eq:minmax-problem}
x^{*} \;\in\; \arg\min_{x \in \mathcal{X}} g(x),
\end{equation}
where $g(x)$ is defined in~\eqref{eq:objective}. This min-max formulation 
captures the requirement of uniform fidelity across the functional domain and 
differs fundamentally from minimizing the integrated deviation $\mathcal{I}(x)$, 
which prioritizes average performance. While $g(x)$ is a scalar-valued 
objective, directly optimizing it requires evaluating 
$\sup_{\lambda \in \Lambda} d(x,\lambda)$ at each candidate design by querying 
$f$ over a dense grid of $\lambda$ values, which is expensive. The following 
section develops a functional Gaussian process surrogate that models 
$f(x,\lambda)$ jointly across $\mathcal{X} \times \Lambda$, enabling tractable 
prediction of $g(x)$ and principled uncertainty quantification across the 
functional domain.

\section{Methodology}
\label{sec:method}

We now develop a Bayesian optimization framework for function-valued responses 
under the min-max objective~\eqref{eq:objective}. The methodology proceeds in five steps: we represent the unknown response 
surface $f(x,\lambda)$ as a Gaussian process over the joint domain 
$\mathcal{X} \times \Lambda$ and derive its Karhunen--Lo\`eve decomposition 
into independent coefficient processes 
(Section~\ref{subsec:functional-rep}), fit independent Gaussian process 
surrogates to these coefficient processes 
(Section~\ref{subsec:surrogate}), derive the closed-form distribution of 
the squared deviation from the target 
(Section~\ref{subsec:deviation-dist}), construct a tractable acquisition 
function that balances worst-case exploitation with integrated 
exploration (Section~\ref{subsec:acquisition}), and develop the numerical 
procedure for implementing the methodology in practice 
(Section~\ref{subsec:implementation}).

\subsection{Functional Representation of the Response}
\label{subsec:functional-rep}

We model the unknown response surface $f(x,\lambda)$ as a Gaussian process
defined over the joint domain $(x,\lambda)\in \mathcal{X}\times\Lambda$,
\begin{equation}
\label{eq:gp-model}
f(x,\lambda)
\;\sim\;
\mathcal{GP}\!\left(0,\; k\!\left((x,\lambda),(x',\lambda')\right)\right),
\end{equation}
where $k\bigl((x,\lambda),(x',\lambda')\bigr)$ is the covariance kernel.
We adopt a zero-mean prior for notational simplicity; in applications where 
the functional response has a known trend, a mean function can be estimated 
from an initial sample and subtracted before modeling, with no loss of 
generality.

To capture the dependence structure across the design space and functional 
domain simultaneously, we adopt a separable covariance kernel of the form
\begin{equation}
\label{eq:separable-kernel}
k\!\left((x,\lambda),(x',\lambda')\right)
=
k_{x}(x,x')\,k_{\lambda}(\lambda,\lambda'),
\end{equation}
where $k_{x}:\mathcal{X}\times\mathcal{X}\to\mathbb{R}$ encodes similarity
between design points and $k_{\lambda}:\Lambda\times\Lambda\to\mathbb{R}$
encodes correlation along the functional domain. The separable form reflects 
the assumption that proximity in $\mathcal{X}$ and proximity in $\Lambda$ 
contribute independently to the overall covariance structure, which is natural 
when design inputs and functional indices represent qualitatively different 
quantities (e.g., geometric parameters versus wavelength).

A key structural property of functional optimization problems is that the
functional domain is typically discretized at a much finer resolution than
the design space, so the dominant complexity of the response surface arises
from variation along $\lambda$. To exploit this structure, we analyze
$k_{\lambda}$ through its spectral decomposition. Assume $k_{\lambda}$ is 
continuous, positive definite, and strictly positive definite (i.e., the 
induced integral operator on $L^{2}(\Lambda)$ has trivial null space), so that 
it generates a complete orthonormal eigenbasis. Then Mercer's theorem 
guarantees the absolutely and uniformly convergent expansion
\begin{equation}
\label{eq:mercer-expansion}
k_{\lambda}(\lambda,\lambda')
=
\sum_{m=1}^{\infty}
\gamma_m \,\phi_m(\lambda)\,\phi_m(\lambda'),
\end{equation}
where $\gamma_1 \ge \gamma_2 \ge \cdots > 0$ are the eigenvalues and
$\{\phi_m\}_{m=1}^{\infty}$ are the corresponding orthonormal eigenfunctions
satisfying
\begin{equation}
\label{eq:orthonormality}
\int_{\Lambda} \phi_m(\lambda)\,\phi_{m'}(\lambda)\,d\lambda 
\;=\; 
\delta_{mm'},
\end{equation}
where $\delta_{mm'}$ denotes the Kronecker delta ($\delta_{mm'} = 1$ if 
$m = m'$, and $0$ otherwise). The eigenfunctions form a complete orthonormal 
basis for $L^{2}(\Lambda)$ and capture the dominant modes of correlation 
implied by $k_{\lambda}$. Standard smooth kernels (e.g., squared-exponential, 
Mat\'ern) satisfy these conditions.

Substituting the Mercer expansion~\eqref{eq:mercer-expansion} into the 
separable covariance~\eqref{eq:separable-kernel} yields
\begin{equation}
\label{eq:joint-expansion}
k\!\left((x,\lambda),(x',\lambda')\right)
=
\sum_{m=1}^{\infty}
\gamma_m\,
k_{x}(x,x')\,
\phi_m(\lambda)\,\phi_m(\lambda').
\end{equation}
This factored form expresses the joint covariance as a countable sum of 
separable terms, each indexed by a functional mode. We now show how this 
structural decomposition of the kernel induces a corresponding representation 
of the Gaussian process $f(x,\lambda)$ itself.

\paragraph{Karhunen--Lo\`eve representation of $f$.}
Because $\{\phi_m\}_{m=1}^\infty$ form a complete orthonormal basis for 
$L^{2}(\Lambda)$, and because the random curve $f(x,\cdot)$ lies in 
$L^{2}(\Lambda)$ almost surely under our finite-variance assumption, $f(x,\cdot)$ 
admits a unique series expansion in this basis,
\begin{equation}
\label{eq:kl-expansion}
f(x,\lambda)
\;=\;
\sum_{m=1}^{\infty}
\alpha_m(x)\,\phi_m(\lambda),
\end{equation}
where the coefficients $\{\alpha_m(x)\}_{m=1}^\infty$ are random functions of 
$x$ and the series converges in $L^{2}(\Lambda)$ for each $x$. The coefficients 
are uniquely determined by orthonormality: taking the inner product of both 
sides of~\eqref{eq:kl-expansion} with $\phi_{m'}(\lambda)$ and applying 
condition~\eqref{eq:orthonormality} isolates
\begin{equation}
\label{eq:coeff-projection}
\alpha_m(x) 
\;=\; 
\int_{\Lambda} f(x,\lambda)\,\phi_m(\lambda)\,d\lambda,
\qquad m = 1, 2, \ldots.
\end{equation}
The expansion~\eqref{eq:kl-expansion} together with~\eqref{eq:coeff-projection} 
constitutes the Karhunen--Lo\`eve representation of the Gaussian process 
$f(x,\lambda)$ under the separable covariance structure. It cleanly separates 
the two sources of variation in the response surface: the eigenfunctions 
$\phi_m(\lambda)$ encode the shape of each functional mode, while the 
coefficients $\alpha_m(x)$ govern how the amplitude of that mode varies across 
designs.

The distributional properties of the coefficient processes follow directly 
from~\eqref{eq:coeff-projection} and the Gaussianity of $f$. Because each 
$\alpha_m(x)$ is a linear functional of the Gaussian process $f$, the family 
$\{\alpha_m(x)\}_{m=1}^\infty$ is jointly Gaussian. The following proposition 
characterizes their mean and covariance structure, establishing that they are 
mutually independent Gaussian processes over $\mathcal{X}$.

\begin{proposition}[Coefficient Processes of the KL Expansion]
\label{prop:coeff-processes}
Under the separable covariance~\eqref{eq:separable-kernel} with 
Mercer expansion~\eqref{eq:mercer-expansion}, the coefficient processes 
$\{\alpha_m(x)\}_{m=1}^\infty$ defined by the 
projection~\eqref{eq:coeff-projection} are mutually independent zero-mean 
Gaussian processes over $\mathcal{X}$ with cross-covariance
\begin{equation}
\label{eq:coeff-cov}
\mathrm{Cov}\!\left(\alpha_m(x),\,\alpha_{m'}(x')\right)
\;=\;
\gamma_m\, k_{x}(x,x')\,\delta_{mm'}.
\end{equation}
Equivalently, each coefficient process is independently distributed as
\begin{equation}
\label{eq:coeff-prior}
\alpha_m(x)
\;\sim\;
\mathcal{GP}\!\left(0,\;\gamma_m\, k_{x}(x,x')\right),
\qquad m = 1, 2, \ldots.
\end{equation}
\end{proposition}

The proof is provided in Appendix~\ref{app:coeff-processes} and proceeds by computing the 
covariance~\eqref{eq:coeff-cov} directly from the projection 
definition~\eqref{eq:coeff-projection}, using the eigenfunction property
\begin{equation}
\label{eq:eigenfunction-property}
\int_\Lambda k_\lambda(\lambda,\lambda')\,\phi_m(\lambda')\,d\lambda' 
\;=\; 
\gamma_m\,\phi_m(\lambda),
\end{equation}
together with the orthonormality condition~\eqref{eq:orthonormality}.

Proposition~\ref{prop:coeff-processes} reveals the key structural insight 
underlying our methodology: under the separable covariance assumption, the 
infinite-dimensional functional GP $f(x,\lambda)$ decomposes into a countable 
collection of independent finite-dimensional GPs $\{\alpha_m(x)\}$ over the 
design space, each scaled by the spectral weight $\sqrt{\gamma_m}$. This 
decomposition transforms a problem requiring inference over a function-valued 
surface into one requiring parallel inference over scalar-valued surrogates, a 
substantial simplification that we exploit throughout the remainder of the 
methodology.

\paragraph{Truncation and finite-dimensional approximation.}
While the representation~\eqref{eq:kl-expansion} is theoretically exact, the 
series is infinite and therefore not directly computable. In practice, the 
eigenvalues $\{\gamma_m\}$ of smooth kernels (e.g., squared-exponential, 
Mat\'ern) decay rapidly, so higher-order modes contribute negligibly to the 
overall covariance structure. This motivates approximating the infinite 
expansion by retaining only the $M$ leading terms,
\begin{equation}
\label{eq:truncated-model}
f^{(M)}(x,\lambda)
\;=\;
\sum_{m=1}^{M}
\alpha_m(x)\,\phi_m(\lambda),
\qquad
\alpha_m(x)\;\sim\;\mathcal{GP}\!\left(0,\,\gamma_m k_{x}(x,x')\right),
\quad m=1,\dots,M.
\end{equation}
The truncation level $M$ controls the trade-off between approximation
fidelity and computational cost, and is chosen via the cumulative explained
variance ratio
\begin{equation}
\label{eq:variance-ratio}
r_M
\;=\;
\frac{\sum_{m=1}^{M}\gamma_m}{\sum_{m=1}^{\infty}\gamma_m},
\end{equation}
where the denominator equals $\int_\Lambda k_\lambda(\lambda,\lambda)\,d\lambda$, the trace of the integral operator induced by $k_\lambda$ on $L^{2}(\Lambda)$, 
and is finite for any trace-class kernel. One selects the smallest $M$ 
satisfying $r_M \geq \tau$ for a user-specified threshold $\tau\in(0,1)$ close 
to one (e.g., $\tau=0.99$), ensuring that the truncated model captures the 
dominant structure of the functional response while discarding modes whose 
contribution to the total variance is negligible. The following proposition 
quantifies the mean squared error incurred by this truncation.

\begin{proposition}[Mercer Truncation Error]
\label{prop:truncation}
Let $f(x,\lambda)$ be the Gaussian process with separable 
covariance~\eqref{eq:separable-kernel}, admitting the Karhunen--Lo\`eve 
representation~\eqref{eq:kl-expansion}, and let $f^{(M)}(x,\lambda)$ be its 
$M$-term truncation defined in~\eqref{eq:truncated-model}. Then for every 
$x\in\mathcal{X}$,
\begin{equation}
\label{eq:truncation-error}
\mathbb{E}\!\left[
\int_{\Lambda}
\bigl(f(x,\lambda)-f^{(M)}(x,\lambda)\bigr)^2
\,d\lambda
\right]
\;=\;
k_x(x,x)
\sum_{m=M+1}^{\infty}
\gamma_m
\;=\;
k_x(x,x)\!\left(\sum_{m=1}^{\infty}\gamma_m\right)(1-r_M),
\end{equation}
where $r_M$ is the variance ratio~\eqref{eq:variance-ratio}. Consequently, the 
mean squared truncation error is bounded above by 
$k_x(x,x)(\sum_{m=1}^{\infty}\gamma_m)(1-\tau)$ whenever $r_M\ge\tau$.
\end{proposition}

Proposition~\ref{prop:truncation} shows that the threshold rule $r_M\ge\tau$
directly controls the mean squared approximation error, with the residual
error vanishing as $\tau\to 1$ or equivalently as the tail eigenvalues
$\gamma_{M+1},\gamma_{M+2},\ldots$ become negligible. A proof is provided in 
Appendix~\ref{app:truncation}.

Under the truncation~\eqref{eq:truncated-model}, the infinite-dimensional 
functional GP reduces to a finite collection of $M$ independent scalar 
Gaussian processes $\{\alpha_m(x)\}_{m=1}^M$ over the design space 
$\mathcal{X}$. Bayesian optimization of the original functional response 
therefore reduces to inference over these coefficient processes, given 
observations of the form $\alpha_m(x_i) = \int_{\Lambda} f(x_i,\lambda)\,
\phi_m(\lambda)\,d\lambda$ at design points $\{x_1,\ldots,x_n\}$. 
Section~\ref{subsec:surrogate} develops this inference, treating the 
coefficient observations as available; the numerical procedure for computing 
them from observed functional responses is deferred to 
Section~\ref{subsec:implementation}.
\subsection{Posterior Inference via Coefficient Surrogates}
\label{subsec:surrogate}

Section~\ref{subsec:functional-rep} reduced the functional Gaussian process 
$f(x,\lambda)$ to a finite collection of $M$ mutually independent scalar 
Gaussian processes $\{\alpha_m(x)\}_{m=1}^M$ over the design space 
$\mathcal{X}$, related to the original surface through the 
expansion~\eqref{eq:truncated-model} and the 
projection~\eqref{eq:coeff-projection}. We now develop posterior inference 
for these coefficient processes given a collection of design points and the 
corresponding coefficient observations, then reconstruct the posterior of 
the full functional response from the resulting coefficient posteriors.

Suppose we have evaluated the response at $n$ design points 
$\{x_1,\ldots,x_n\} \subset \mathcal{X}$, and have access to the 
corresponding coefficient values 
$\{\alpha_m(x_i)\}_{i=1}^n$ for each mode $m = 1,\ldots,M$. 
In practice these values are not obtained exactly: they are computed by 
numerical projection of observed functional responses onto the basis 
$\{\phi_m\}$, a procedure detailed in Section~\ref{subsec:implementation}. 
The projection introduces small errors arising from quadrature, basis 
truncation, and any measurement noise present in the original functional 
observations. We model these errors collectively as additive zero-mean 
Gaussian noise that is independent across modes,
\begin{equation}
\label{eq:obs-model}
\widetilde\alpha_m(x_i) 
\;=\; 
\alpha_m(x_i) + \epsilon_{im},
\qquad
\epsilon_{im} \overset{\text{iid}}{\sim} \mathcal{N}\!\left(0,\;\sigma_{n,m}^2\right),
\qquad
i = 1,\ldots,n,\;\; m = 1,\ldots,M,
\end{equation}
where $\widetilde\alpha_m(x_i)$ denotes the noisy observation actually used in 
inference and $\sigma_{n,m}^2$ is a mode-specific noise variance estimated 
from data. The mode-independence assumption holds exactly when the projection 
is computed on a uniform grid (the cross-mode noise covariance vanishes by 
orthonormality) or when the original responses are noise-free, and serves as 
a natural approximation otherwise. Collecting the observations for mode $m$ 
into the vector
\begin{equation}
\label{eq:obs-vector}
\boldsymbol{\alpha}_m 
\;=\; 
\bigl[\widetilde\alpha_m(x_1),\,\widetilde\alpha_m(x_2),\,\ldots,\,
\widetilde\alpha_m(x_n)\bigr]^\top 
\;\in\; \mathbb{R}^{n},
\end{equation}
we treat $\boldsymbol{\alpha}_m$ as the training data for an independent 
Gaussian process regression on the $m$-th coefficient process.

From Proposition~\ref{prop:coeff-processes}, each coefficient process carries 
the prior~\eqref{eq:coeff-prior},
\begin{equation*}
\alpha_m(x) 
\;\sim\; 
\mathcal{GP}\!\left(0,\;\gamma_m\,k_x(x,x')\right),
\qquad m = 1,\ldots,M,
\end{equation*}
where $k_x:\mathcal{X}\times\mathcal{X}\to\mathbb{R}$ is a user-specified 
stationary covariance kernel parameterized by hyperparameters $\vartheta_m$. 
Define the $n\times n$ training kernel matrix
\begin{equation}
\label{eq:Kx-matrix}
K_x \;=\; \bigl[k_x(x_i,x_j)\bigr]_{i,j=1}^{n},
\end{equation}
and the cross-covariance vector between a new design point $x$ and the 
training inputs,
\begin{equation}
\label{eq:kx-vector}
\mathbf{k}_x(x) 
\;=\; 
\bigl[k_x(x,x_1),\,\ldots,\,k_x(x,x_n)\bigr]^{\top} \;\in\; \mathbb{R}^{n}.
\end{equation}
Combining the prior~\eqref{eq:coeff-prior} with the observation 
model~\eqref{eq:obs-model}, the joint distribution of the training 
observations $\boldsymbol{\alpha}_m$ and the coefficient value $\alpha_m(x)$ 
at a new design point $x$ is jointly Gaussian, with marginal training 
covariance $\gamma_m K_x + \sigma_{n,m}^2 I_n$ and cross-covariance 
$\gamma_m \mathbf{k}_x(x)$. Standard Gaussian conditioning then yields the 
posterior
\begin{equation}
\label{eq:coeff-posterior}
\alpha_m(x) \mid \boldsymbol{\alpha}_m 
\;\sim\; 
\mathcal{N}\!\left(\mu_m(x),\;\sigma_m^2(x)\right),
\end{equation}
with posterior mean and variance
\begin{align}
\mu_m(x) 
&\;=\; 
\gamma_m\,\mathbf{k}_x(x)^{\top}
\bigl(\gamma_m K_x + \sigma_{n,m}^2 I_n\bigr)^{-1}
\boldsymbol{\alpha}_m, 
\label{eq:coeff-post-mean}\\[4pt]
\sigma_m^2(x) 
&\;=\; 
\gamma_m\,k_x(x,x) 
\;-\; 
\gamma_m^2\,\mathbf{k}_x(x)^{\top}
\bigl(\gamma_m K_x + \sigma_{n,m}^2 I_n\bigr)^{-1}
\mathbf{k}_x(x).
\label{eq:coeff-post-var}
\end{align}
Because the coefficient processes are mutually independent under the 
prior~\eqref{eq:coeff-prior}, and the noise terms in~\eqref{eq:obs-model} are 
independent across modes, the joint posterior over 
$\{\alpha_m(x)\}_{m=1}^M$ factorizes: the $M$ coefficient posteriors are 
mutually independent and can be computed in parallel.

\paragraph{Reconstructing the functional posterior.}
The posterior over the full functional response surface is recovered from the 
$M$ coefficient posteriors through the basis expansion~\eqref{eq:truncated-model}. 
For any design--index pair $(x,\lambda) \in \mathcal{X}\times\Lambda$, define 
the data 
$\mathcal{D} = \{(x_i, \boldsymbol{\alpha}_{1:M}(x_i))\}_{i=1}^n$ where 
$\boldsymbol{\alpha}_{1:M}(x_i) = (\widetilde\alpha_1(x_i),\ldots,\widetilde\alpha_M(x_i))$. 
The posterior mean and variance of $f^{(M)}(x,\lambda)$ at $(x,\lambda)$ are
\begin{equation}
\label{eq:post-f}
\mu_f(x,\lambda) 
\;=\; 
\sum_{m=1}^{M} \mu_m(x)\,\phi_m(\lambda),
\qquad
\sigma_f^{2}(x,\lambda) 
\;=\; 
\sum_{m=1}^{M} \sigma_m^{2}(x)\,\phi_m(\lambda)^{2}.
\end{equation}
and the marginal posterior at any $(x,\lambda)$ is Gaussian,
\begin{equation}
\label{eq:functional-posterior}
f^{(M)}(x,\lambda) \mid \mathcal{D} 
\;\sim\; 
\mathcal{N}\!\left(\mu_f(x,\lambda),\;\sigma_f^{2}(x,\lambda)\right).
\end{equation}
The following proposition formalizes this reconstruction and additionally 
characterizes the full posterior covariance structure across the functional 
domain, going beyond the marginal variance to describe how uncertainty at 
different functional locations $\lambda$ and $\lambda'$ co-varies.

\begin{proposition}[Posterior Reconstruction of the Functional Response]
\label{prop:posterior}
Let $\mu_m(x)$ and $\sigma_m^2(x)$ denote the posterior mean and variance of 
the $m$-th coefficient process given by~\eqref{eq:coeff-post-mean} 
and~\eqref{eq:coeff-post-var}. Under the truncated separable GP 
model~\eqref{eq:truncated-model}, the marginal posterior of the functional 
response at any $(x,\lambda) \in \mathcal{X}\times\Lambda$ is Gaussian with 
mean and variance given by~~\eqref{eq:post-f}. 
Furthermore, the posterior covariance between the functional response at two 
index locations $\lambda$ and $\lambda'$ is
\begin{equation}
\label{eq:post-cov-f}
\mathrm{Cov}\!\left(f^{(M)}(x,\lambda),\,f^{(M)}(x,\lambda') \,\middle|\, 
\mathcal{D}\right)
\;=\;
\sum_{m=1}^{M} \sigma_m^{2}(x)\,\phi_m(\lambda)\,\phi_m(\lambda').
\end{equation}
\end{proposition}

A proof is provided in Appendix~\ref{app:posterior}. Proposition~\ref{prop:posterior} shows 
that the posterior uncertainty over the entire functional response surface is 
fully determined by the $M$ coefficient posteriors and the basis functions, 
with the covariance between any two functional locations factoring cleanly 
through the shared basis. The original infinite-dimensional functional 
inference problem thereby reduces to $M$ parallel scalar GP regressions over 
$\mathcal{X}$, each computationally no more expensive than standard Bayesian 
optimization. The kernel hyperparameters $\vartheta_m$ and noise variance $\sigma_{n,m}^2$ 
appearing in~\eqref{eq:coeff-post-mean}--\eqref{eq:coeff-post-var} are 
estimated by standard type-II maximum likelihood applied to each mode 
independently \citep{Rasmussen2006GP}, and re-estimated at each Bayesian 
optimization iteration as new observations are incorporated.

\subsection{Squared Deviation Distribution}
\label{subsec:deviation-dist}

Section~\ref{subsec:surrogate} established the Gaussian posterior of the 
truncated functional response $f^{(M)}(x,\lambda)$ given observed data. We 
now use this posterior to characterize the distribution of the pointwise 
squared deviation from the target,
\begin{equation*}
d(x,\lambda) \;=\; \bigl(f(x,\lambda) - f^{*}(\lambda)\bigr)^{2},
\end{equation*}
introduced in Section~\ref{sec:formulation}. Because the posterior of $f$ is 
Gaussian, the posterior distribution of $d(x,\lambda)$ admits a closed-form 
characterization with explicit moments, which serve as the building blocks of 
the acquisition function in Section~\ref{subsec:acquisition}.

\paragraph{Posterior deviation process.}
The squared deviation $d(x,\lambda)$ depends on the unknown response 
$f(x,\lambda)$ and is therefore itself a posterior random variable under the 
GP surrogate. To characterize its distribution, define the posterior 
deviation process
\begin{equation}
\label{eq:deviation-process}
h(x,\lambda) 
\;=\; 
f^{(M)}(x,\lambda) - f^{*}(\lambda),
\end{equation}
which is a deterministic shift of the Gaussian posterior of $f^{(M)}$ by the 
known target $f^{*}(\lambda)$. By Proposition~\ref{prop:posterior}, $h(x,\lambda)$ 
inherits Gaussianity: for every $(x,\lambda)\in\mathcal{X}\times\Lambda$,
\begin{equation}
\label{eq:h-distribution}
h(x,\lambda) \;\sim\; 
\mathcal{N}\!\left(\mu_h(x,\lambda),\;\sigma_h^{2}(x,\lambda)\right),
\end{equation}
where the posterior mean and variance are
\begin{equation}
\label{eq:h-moments}
\mu_h(x,\lambda) \;=\; \mu_f(x,\lambda) - f^{*}(\lambda),
\qquad
\sigma_h^{2}(x,\lambda) \;=\; \sigma_f^{2}(x,\lambda),
\end{equation}
with $\mu_f$ and $\sigma_f^{2}$ given by~\eqref{eq:post-f}. The mean 
$\mu_h(x,\lambda)$ measures the expected discrepancy between the surrogate 
prediction and the target at functional location $\lambda$, while the 
variance $\sigma_h^{2}(x,\lambda)$ reflects residual posterior uncertainty in 
the functional response at that location.

\paragraph{Distribution of the squared deviation.}
Since $d(x,\lambda) = h(x,\lambda)^{2}$ is the square of a Gaussian random 
variable, its posterior distribution follows directly from~\eqref{eq:h-distribution}. 
Standardizing $h(x,\lambda)$ by its standard deviation yields a normal random 
variable with unit variance and mean $\mu_h(x,\lambda)/\sigma_h(x,\lambda)$, 
whose square is, by definition, a noncentral chi-squared random variable 
with one degree of freedom. Specifically,
\begin{equation}
\label{eq:d-distribution}
d(x,\lambda) 
\;\sim\; 
\sigma_h^{2}(x,\lambda)\,\chi_1'^{\,2}\!\left(\delta(x,\lambda)\right),
\qquad
\delta(x,\lambda) \;=\; \frac{\mu_h^{2}(x,\lambda)}{\sigma_h^{2}(x,\lambda)},
\end{equation}
where $\chi_1'^{\,2}(\delta)$ denotes the noncentral chi-squared distribution 
with one degree of freedom and noncentrality parameter $\delta$, and the 
scaling factor $\sigma_h^{2}(x,\lambda)$ accounts for the variance of the 
underlying Gaussian. The distribution admits an explicit probability density 
function via the folded-normal representation: for $y > 0$,
\begin{equation}
\label{eq:d-density}
p_d(y \mid x,\lambda) 
\;=\; 
\frac{1}{\sigma_h(x,\lambda)\sqrt{2\pi y}}\,
\exp\!\left(
-\frac{y + \mu_h^{2}(x,\lambda)}{2\sigma_h^{2}(x,\lambda)}
\right)
\cosh\!\left(
\frac{\mu_h(x,\lambda)\sqrt{y}}{\sigma_h^{2}(x,\lambda)}
\right).
\end{equation}
The density makes explicit how the two posterior quantities shape the 
distribution: $\mu_h$ controls the location of the mode, while $\sigma_h$ 
governs the spread.

Constructing an acquisition function for the min--max objective requires 
summarizing the distribution~\eqref{eq:d-distribution} through its moments. 
Both moments admit clean closed-form expressions, formalized in the 
following proposition.

\begin{proposition}[Distribution and Moments of the Posterior Squared Deviation]
\label{prop:deviation}
Suppose $h(x,\lambda)\sim\mathcal{N}\!\left(\mu_h(x,\lambda),\sigma_h^{2}(x,\lambda)\right)$ 
for fixed $(x,\lambda)\in\mathcal{X}\times\Lambda$, with 
$\sigma_h^{2}(x,\lambda) > 0$. Then the squared posterior deviation 
$d(x,\lambda) = h(x,\lambda)^{2}$ follows the scaled noncentral chi-squared 
distribution~\eqref{eq:d-distribution}, and its first two moments are given 
in closed form by
\begin{equation}
\label{eq:d-mean}
\mu_d(x,\lambda) 
\;=\; 
\mathbb{E}\bigl[d(x,\lambda)\bigr] 
\;=\; 
\mu_h^{2}(x,\lambda) + \sigma_h^{2}(x,\lambda),
\end{equation}
\begin{equation}
\label{eq:d-var}
\sigma_d^{2}(x,\lambda) 
\;=\; 
\mathrm{Var}\bigl(d(x,\lambda)\bigr) 
\;=\; 
2\sigma_h^{4}(x,\lambda) + 4\mu_h^{2}(x,\lambda)\,\sigma_h^{2}(x,\lambda).
\end{equation}
\end{proposition}

A proof is provided in Appendix~\ref{app:deviation}. The expressions in 
Proposition~\ref{prop:deviation} have a natural interpretation: the posterior 
mean $\mu_d(x,\lambda)$ combines the squared bias $\mu_h^{2}(x,\lambda)$ and 
the posterior variance $\sigma_h^{2}(x,\lambda)$ at each functional location, 
while the posterior variance $\sigma_d^{2}(x,\lambda)$ captures the 
uncertainty in how large the deviation will be. Crucially, both 
quantities~\eqref{eq:d-mean}--\eqref{eq:d-var} are available analytically from 
the GP posteriors of Section~\ref{subsec:surrogate}, with no Monte Carlo 
approximation required.

The key implication is that the exploration and exploitation terms in the 
acquisition function constructed in Section~\ref{subsec:acquisition} are not 
heuristic quantities but arise from an exact distributional characterization 
of the squared deviation. The posterior standard deviation 
$\sigma_d(x,\lambda) = \sqrt{\sigma_d^{2}(x,\lambda)}$ provides a local 
measure of epistemic uncertainty at each $\lambda$, and 
$\mu_d(x,\lambda)$ together with $\sigma_d(x,\lambda)$ supply the two 
ingredients needed to define a tractable min--max acquisition function in 
the next subsection.

\subsection{Acquisition Function and Consistency}
\label{subsec:acquisition}

The ideal next evaluation point minimizes the true min--max objective 
$g(x) = \sup_{\lambda\in\Lambda} d(x,\lambda)$ defined in~\eqref{eq:objective}. 
Under the GP surrogate, $d(x,\lambda)$ is a posterior random variable for 
each $\lambda$ (Section~\ref{subsec:deviation-dist}), so the natural 
surrogate for $g(x)$ would be $\sup_{\lambda} d(x,\lambda)$, the supremum of 
a noncentral chi-squared process indexed by $\lambda$. Characterizing the 
distribution of this supremum analytically is intractable, and 
simulation-based approximations such as posterior sampling over dense grids 
become computationally prohibitive in higher dimensions. We therefore design 
a closed-form acquisition function that preserves the min--max structure 
while remaining fully tractable.

\paragraph{Acquisition function.}
Our acquisition combines two complementary terms: the worst-case expected 
squared deviation, which drives exploitation, and an integrated measure of 
predictive uncertainty, which drives exploration. Formally, for a 
trade-off parameter $\kappa > 0$, define
\begin{equation}
\label{eq:acquisition}
\alpha(x) 
\;=\; 
\sup_{\lambda\in\Lambda}\,\mu_d(x,\lambda) 
\;-\; 
\kappa \int_{\Lambda} \sigma_d(x,\lambda)\,d\lambda,
\end{equation}
where $\mu_d(x,\lambda)$ and $\sigma_d(x,\lambda) = \sqrt{\sigma_d^{2}(x,\lambda)}$ 
are the posterior mean and standard deviation of $d(x,\lambda)$ given by 
Proposition~\ref{prop:deviation}. The next evaluation point is obtained by 
minimizing $\alpha(x)$ over $\mathcal{X}$,
\begin{equation}
\label{eq:next-point}
x_{\mathrm{next}} 
\;=\; 
\arg\min_{x\in\mathcal{X}}\,\alpha(x).
\end{equation}

The first term, $\sup_{\lambda} \mu_d(x,\lambda)$, is the predicted 
worst-case squared deviation across the functional domain; minimizing it 
focuses the search on designs that reduce the largest expected discrepancy 
from the target, directly targeting the min--max objective. The second 
term, $\int_{\Lambda} \sigma_d(x,\lambda)\,d\lambda$, aggregates pointwise 
uncertainty in the squared deviation over the entire domain; subtracting 
it encourages evaluation at designs where the deviation process is poorly 
characterized, preventing the search from becoming trapped at spurious 
local minima of the mean predictor. The trade-off parameter $\kappa$ governs 
the balance between these two competing objectives.

\paragraph{Consistency of the acquisition.}
Together, the four propositions established so far form the theoretical 
foundation of the methodology: 
Proposition~\ref{prop:coeff-processes} characterizes the coefficient 
processes, Proposition~\ref{prop:truncation} controls the truncation error, 
Proposition~\ref{prop:posterior} reconstructs the functional posterior 
exactly, and Proposition~\ref{prop:deviation} provides closed-form moments 
for the squared deviation. Taken together, they ensure that the acquisition 
function~\eqref{eq:acquisition} is constructed from distributional quantities 
rather than heuristic surrogates. The following theorem provides the 
capstone guarantee: as the GP surrogate becomes accurate and posterior 
uncertainty vanishes, the acquisition converges uniformly to the true 
min--max objective, and any sequence of approximate minimizers converges to 
the true min--max optimizer.

\begin{theorem}[Consistency of the Acquisition Function and Its Minimizers]
\label{thm:consistency}
Suppose $\mathcal{X}$ and $\Lambda$ are compact, $g(x) = \sup_{\lambda\in\Lambda} 
d(x,\lambda)$ is continuous on $\mathcal{X}$, and let 
\begin{equation*}
\alpha_t(x) 
\;=\; 
\sup_{\lambda\in\Lambda}\,\mu_{d,t}(x,\lambda) 
\;-\; 
\kappa_t \int_{\Lambda} \sigma_{d,t}(x,\lambda)\,d\lambda
\end{equation*}
denote the acquisition function~\eqref{eq:acquisition} at iteration $t$, with 
$\mu_{d,t}$, $\sigma_{d,t}$, and $\kappa_t$ the corresponding posterior mean, 
posterior standard deviation, and trade-off parameter. Suppose that as 
$t\to\infty$:
\begin{enumerate}[label=(\roman*)]
    \item the surrogate mean converges uniformly: 
          $\sup_{x\in\mathcal{X}} \sup_{\lambda\in\Lambda} 
          |\mu_{d,t}(x,\lambda) - d(x,\lambda)| \to 0$,
    \item the integrated posterior uncertainty vanishes: 
          $\sup_{x\in\mathcal{X}} \int_{\Lambda} \sigma_{d,t}(x,\lambda)\,d\lambda \to 0$, 
    \item the trade-off parameter remains bounded: 
          $\sup_{t} \kappa_t < \infty$.
\end{enumerate}
Then the acquisition converges uniformly to the true objective,
\begin{equation}
\label{eq:uniform-convergence}
\sup_{x\in\mathcal{X}}\,\bigl|\alpha_t(x) - g(x)\bigr| \;\to\; 0,
\end{equation}
and any sequence $\{x_t\}$ satisfying $\alpha_t(x_t) \le \inf_x \alpha_t(x) + 
\varepsilon_t$ with $\varepsilon_t \to 0$ has every limit point in 
$\arg\min_{x\in\mathcal{X}} g(x)$.
\end{theorem}

A proof is provided in Appendix~\ref{app:consistency}. Theorem~\ref{thm:consistency} establishes 
that the proposed methodology is asymptotically correct: provided the GP 
surrogates become accurate and posterior uncertainty is reduced through 
sufficient exploration, the algorithm converges to the true min--max 
solution. The hypotheses are natural: condition~(i) holds when the GP 
posterior mean is a consistent estimator of $d(x,\lambda)$, condition~(ii) 
holds when the design space is sufficiently explored over the course of the 
optimization, and condition~(iii) is enforced by the adaptive schedule on 
$\kappa$ described in the next section.
\subsection{Numerical Procedure and Algorithm}
\label{subsec:implementation}

The previous subsections developed the methodology in fully continuous form: 
the response surface $f(x,\lambda)$ is represented by a Karhunen--Lo\`eve 
expansion in $L^2(\Lambda)$, the coefficient processes are inferred via 
Gaussian process regression, the squared deviation distribution is 
characterized exactly, and the acquisition function~\eqref{eq:acquisition} is 
defined as a continuous functional of $\mu_d$ and $\sigma_d$ over the 
functional domain $\Lambda$. Implementing this methodology in practice 
requires three numerical components: extracting the basis $\{\phi_m\}_{m=1}^M$ 
from a discretized version of the kernel $k_\lambda$, projecting observed 
functional responses onto this basis to obtain coefficient observations, 
and evaluating the acquisition function on a finite grid for optimization 
within the Bayesian optimization loop. We address each in turn, then state 
the discrete acquisition formally and conclude with the algorithm.

\paragraph{Discrete basis extraction.}
The Mercer eigenfunctions $\{\phi_m\}$ are not available in closed form for 
general kernels and must be approximated numerically. Suppose functional 
responses are recorded on a common grid 
$\{\lambda_1,\ldots,\lambda_T\}\subset\Lambda$ with associated quadrature 
weights $\Delta_1,\ldots,\Delta_T > 0$ satisfying 
$\sum_{j=1}^T \Delta_j \approx |\Lambda|$, where $|\Lambda|$ denotes the 
length of the functional domain. For a uniform grid with spacing $\Delta$, 
all weights equal $\Delta$; for a nonuniform grid, standard choices include 
trapezoidal or Simpson weights. Collect the weights into the diagonal matrix 
$W = \mathrm{diag}(\Delta_1,\ldots,\Delta_T)\in\mathbb{R}^{T\times T}$ and 
construct the $T\times T$ kernel matrix
\begin{equation}
\label{eq:K-lambda-matrix}
K_\lambda \;=\; \bigl[k_\lambda(\lambda_i,\lambda_j)\bigr]_{i,j=1}^{T}.
\end{equation}
The discrete basis is obtained from the eigendecomposition of the 
symmetrically weighted matrix
\begin{equation}
\label{eq:weighted-eigendecomp}
W^{1/2}\,K_\lambda\,W^{1/2} \;=\; U\,\mathbf{D}\,U^{\top},
\end{equation}
where $\mathbf{D} = \mathrm{diag}(d_1,\ldots,d_T)$ contains the eigenvalues 
in decreasing order $d_1 \ge d_2 \ge \cdots \ge d_T > 0$, and 
$U \in \mathbb{R}^{T\times T}$ has orthonormal columns. The basis matrix is 
then defined as
\begin{equation}
\label{eq:Phi-matrix}
\Phi \;=\; W^{-1/2}\,U_M \;\in\; \mathbb{R}^{T\times M},
\qquad
\Phi_{jm} \;=\; \Delta_j^{-1/2}\,u_{jm},
\end{equation}
where $U_M$ retains the first $M$ columns of $U$. By construction, $\Phi$ 
satisfies the weighted orthonormality condition
\begin{equation}
\label{eq:weighted-orthonormality}
\Phi^{\top} W \Phi \;=\; I_M,
\end{equation}
which is the discrete counterpart of the continuous $L^2(\Lambda)$ 
orthonormality condition~\eqref{eq:orthonormality}. The truncation level $M$ 
is selected via the variance threshold $r_M \ge \tau$ 
from~\eqref{eq:variance-ratio}, applied to the empirical eigenvalues 
$\{d_m\}$. Under the Nystr\"om approximation, $d_m \approx \gamma_m$ when the 
weights $\{\Delta_j\}$ sum to $|\Lambda|$, so $\{d_m\}$ serve as consistent 
estimators of $\{\gamma_m\}$ as $T$ grows. The basis $\Phi$ is computed once 
from the fixed kernel $k_\lambda$ and grid, and is held constant throughout 
the Bayesian optimization procedure.

\paragraph{Coefficient projection.}
Let $Y\in\mathbb{R}^{n\times T}$ denote the matrix of observed functional 
responses, with entries $(Y)_{ij} = f(x_i,\lambda_j)$. The coefficient 
$\alpha_m(x_i) = \int_{\Lambda} f(x_i,\lambda)\,\phi_m(\lambda)\,d\lambda$ 
defined in~\eqref{eq:coeff-projection} is approximated numerically via 
quadrature,
\begin{equation}
\label{eq:projection-quadrature}
\alpha_m(x_i) 
\;\approx\; 
\sum_{j=1}^{T} (Y)_{ij}\,\phi_m(\lambda_j)\,\Delta_j,
\qquad i = 1,\ldots,n,\;\; m = 1,\ldots,M.
\end{equation}
Collecting these projections into the matrix 
$A \in \mathbb{R}^{n\times M}$ with entries $A_{im} = \alpha_m(x_i)$, the 
projection takes the compact form
\begin{equation}
\label{eq:projection-compact}
A \;=\; Y\,W\,\Phi.
\end{equation}
The columns of $A$ provide the coefficient observation vectors 
$\boldsymbol{\alpha}_m = [\alpha_m(x_1),\ldots,\alpha_m(x_n)]^{\top}$ used as 
training data for the GP regressions of 
Section~\ref{subsec:surrogate}. The errors introduced by the quadrature 
approximation in~\eqref{eq:projection-quadrature}, together with any 
measurement noise present in $Y$ and residual truncation effects, are 
absorbed into the mode specific noise variances $\sigma_{n,m}^2$ of 
the observation model~\eqref{eq:obs-model}, and estimated jointly with the 
kernel hyperparameters via type II maximum likelihood.

\paragraph{Discrete acquisition function.}
Algorithm~\ref{alg:mmfbo} optimizes the discrete counterpart of the 
continuous acquisition~\eqref{eq:acquisition}, evaluated on the same grid 
$\{\lambda_1,\ldots,\lambda_T\}$ used for basis extraction:
\begin{equation}
\label{eq:acq-discrete}
\alpha^{T}(x) 
\;=\; 
\max_{1\le j\le T}\,\mu_d(x,\lambda_j) 
\;-\; 
\kappa\,\sum_{j=1}^{T}\sigma_d(x,\lambda_j)\,\Delta_j,
\end{equation}
in which the supremum over $\lambda$ becomes a maximum over grid points 
and the integral becomes a quadrature sum with weights $\Delta_j$. The 
following proposition bounds the error introduced by this discretization. 
The result is stated in general form so it applies uniformly to any 
continuous function on $\Lambda$, in particular to $\mu_d(x,\cdot)$ and 
$\sigma_d(x,\cdot)$ that appear in the acquisition.

\begin{proposition}[Discretization Error for the Supremum]
\label{prop:discretization}
Let $\psi:\mathcal{X}\times\Lambda\to\mathbb{R}$ admit a modulus of 
continuity $\omega$ in $\lambda$ uniformly in $x$, i.e.,
\begin{equation}
\label{eq:modulus-continuity}
\bigl|\psi(x,\lambda) - \psi(x,\lambda')\bigr| 
\;\le\; 
\omega\!\left(|\lambda - \lambda'|\right)
\qquad
\text{for all } x\in\mathcal{X},\;\lambda,\lambda'\in\Lambda,
\end{equation}
where $\omega$ is nondecreasing with $\omega(0) = 0$. Define the continuous 
and discrete suprema
\begin{equation}
\label{eq:psi-suprema}
\psi^{*}(x) \;=\; \sup_{\lambda\in\Lambda} \psi(x,\lambda),
\qquad
\psi_T^{*}(x) \;=\; \max_{1\le t\le T} \psi(x,\lambda_t),
\end{equation}
over a finite collection $\{\lambda_1,\ldots,\lambda_T\}\subset\Lambda$ with 
fill distance 
$h_T = \sup_{\lambda\in\Lambda} \min_{1\le t\le T}|\lambda - \lambda_t|$. Then 
for every $x\in\mathcal{X}$,
\begin{equation}
\label{eq:discretization-bound}
0 \;\le\; \psi^{*}(x) - \psi_T^{*}(x) \;\le\; \omega(h_T).
\end{equation}
If, in addition, $\psi(x,\lambda)$ is $L$ Lipschitz in $\lambda$ uniformly 
in $x$, then $\psi^{*}(x) - \psi_T^{*}(x) \le L\,h_T$, and 
$\psi_T^{*}(x) \to \psi^{*}(x)$ uniformly in $x$ as $h_T \to 0$.
\end{proposition}

A proof is provided in Appendix~\ref{app:discretization}. The continuity hypothesis 
in~\eqref{eq:modulus-continuity} is mild and holds whenever $f$ and $f^{*}$ 
are continuous in $\lambda$ and the eigenfunctions $\{\phi_m\}$ are 
continuous, which covers all kernels considered in this work. Combining 
Theorem~\ref{thm:consistency} with Proposition~\ref{prop:discretization} 
applied to $\mu_d$ and to $\sigma_d$, the discrete 
acquisition~\eqref{eq:acq-discrete} converges uniformly to the continuous 
acquisition $\alpha_t$ as the grid densifies, and therefore inherits the 
consistency guarantee of Theorem~\ref{thm:consistency} provided $h_T\to 0$.

\paragraph{Adaptive trade off and acquisition optimization.}
A fixed value of $\kappa$ in~\eqref{eq:acq-discrete} can over emphasize 
exploration early in the search or prematurely exploit a poorly fitted 
surrogate. We therefore adapt $\kappa$ across iterations using a simple 
three phase schedule: $\kappa$ is set large in early iterations to ensure 
broad coverage of $\mathcal{X}$, reduced as the surrogate stabilizes to 
focus on exploitation, and temporarily increased if progress stagnates to 
escape local traps. This schedule requires no additional tuning parameters 
beyond the initial and minimum values of $\kappa$ and a stagnation 
threshold.

The discrete acquisition $\alpha^{T}(x)$ is non convex in $x$ in general 
and is therefore optimized via multi start local search. A global 
candidate set is formed by combining a Sobol sequence over $\mathcal{X}$ 
with a local pool centered around the current best design, and L BFGS is 
applied from each candidate. Diversity constraints discard restarts that 
are too close to previously evaluated points. The complete procedure is 
summarized in Algorithm~\ref{alg:mmfbo}.

\begin{algorithm}[t]
\caption{Min Max Functional Bayesian Optimization (MM FBO)}
\label{alg:mmfbo}
\begin{algorithmic}[1]
\Require Initial set size $n_0$, total budget $N$, basis threshold $\tau$, 
         initial trade off $\kappa_0$
\State \textbf{Initialize:} evaluate $f(x_j,\cdot)$ at $n_0$ space filling 
       designs $\{x_j\}_{j=1}^{n_0}$; form $Y\in\mathbb{R}^{n_0\times T}$
\State \textbf{Basis extraction:} compute the eigendecomposition 
       $W^{1/2}K_\lambda W^{1/2} = U\,\mathbf{D}\,U^{\top}$; select $M$ via 
       $r_M\ge\tau$; set $\Phi = W^{-1/2}U_M$
\State \textbf{Project:} $A = Y W\Phi$; obtain $\boldsymbol{\alpha}_m$ for 
       $m = 1,\ldots,M$
\State \textbf{Fit surrogates:} for each $m$, maximize the log marginal 
       likelihood to obtain $\hat\vartheta_m$, $\hat\sigma_{n,m}^{2}$; 
       compute coefficient posteriors $\mu_m(x)$, $\sigma_m^{2}(x)$
\While{total evaluations $< N$}
  \State Compute $\mu_d(x,\lambda)$ and $\sigma_d(x,\lambda)$ from 
         Proposition~\ref{prop:deviation}
  \State Form $\alpha^{T}(x)$ as in~\eqref{eq:acq-discrete}
  \State $x_{\mathrm{next}} \leftarrow \arg\min_{x\in\mathcal{X}}\,\alpha^{T}(x)$ 
         via multi start L BFGS with diversity constraints
  \State Evaluate $f(x_{\mathrm{next}},\cdot)$; append to $Y$
  \State Update $A = Y W\Phi$; re estimate hyperparameters; update 
         posteriors $\mu_m,\;\sigma_m^{2}$
  \State Adapt $\kappa$
\EndWhile
\State \Return $x^{*} = \arg\min_{x\in\{\text{evaluated}\}}\, 
       \max_{1\le j\le T}\,\mu_d(x,\lambda_j)$
\end{algorithmic}
\end{algorithm}


\section{Experiments}

We now evaluate the proposed methodology through a combination of controlled benchmarks and physics-inspired case studies. The experimental study is designed to test both accuracy and robustness under different conditions, with particular emphasis on functional responses where reliability across the entire domain is critical. Each experiment follows the same overall protocol: functional responses are generated or collected over a domain $\Lambda$, an initial set of space-filling designs is evaluated, and sequential optimization proceeds under a fixed budget. The proposed acquisition with adaptive strategies is compared against established baselines to assess improvements in convergence speed, stability, and worst-case performance.

Our evaluation proceeds in two stages. First, we consider synthetic benchmark functions, where the true optimum $g^\ast$ and the exact deviation process can be computed. These functions enable systematic exploration across dimensions and replication to quantify variability. Second, we turn to physics-inspired case studies drawn from real engineering problems, including photonic metasurface design to provide a desired scattering spectrum and polymer–inorganic hybrid formation via vapor phase infiltration. In both settings functional responses arise naturally and robustness to local deviations is essential. To contextualize performance we compare MM-FBO against a set of baselines: (i) Latin Hypercube Design (LHD), a static space-filling design (SFD) that provides a non-adaptive benchmark; (ii) standard Bayesian optimization, which fits a GP surrogate to the scalarized objective $g(x)=\max_{\lambda} d(x,\lambda)$ and selects new points via Expected Improvement. In practice we also employ simple variants of $\alpha(x)$ that adapt across phases of the search, beginning with broad exploration, transitioning to a balanced criterion, and ending with refinement that leverages local uncertainty in the squared deviation process.

Performance is quantified through both trajectory-based and summary metrics. The primary trajectory measure is the regret curve
\[
r_k \;=\; g(x_k^{\text{best}}) - g^{\ast},
\qquad 
x_k^{\text{best}} \;=\; \arg\min_{j \leq k} g(x_j),
\]
which records how the best observed design improves over $k$ iterations. Median regret with interquartile ranges is reported across replications to capture stability. To summarize overall efficiency we compute the area under the normalized regret curve (AUOC),
\[
\mathrm{AUOC} \;=\; \frac{1}{B}\sum_{k=1}^{B} \frac{r_k}{r_0},
\]
where $B$ is the evaluation budget and $r_0$ is the initial regret. Lower AUOC indicates faster and more consistent convergence. We also report the distribution of final regrets at budget $B$, and the time-to-threshold (TT) statistic
\[
\mathrm{TT}_\epsilon \;=\; \min \left\{k : \frac{r_k}{r_0} \leq \epsilon \right\},
\]
which measures the iteration count required to reach accuracy $\epsilon$. Together, these metrics provide a comprehensive view of convergence speed, robustness across replications, and worst-case accuracy.

\subsection{Simulation Study}
We evaluate the proposed method on a suite of time-dependent simulation oracles that generate function-valued responses over a dense grid of the index variable $\lambda$, interpreted as time. The problems span a range of dimensions and dynamic behaviors, providing a diverse testbed. For each oracle we define a reference response $f^{\ast}(\lambda)$ at a known target parameter $x^{\ast}$ and measure performance through the worst-case squared deviation objective
\[
g(x) \;=\; \sup_{\lambda \in \Lambda} \big( f(x,\lambda) - f^{\ast}(\lambda) \big)^{2}.
\]

The first oracle is the \textbf{mass–spring–damper system}, a classical model in vibration analysis where damping and natural frequency determine stability and oscillatory behavior. The design vector is $x=(\zeta,\omega_n)\in\mathbb R^2$ and the index is time $\lambda=t$. The displacement $y(t;\zeta,\omega_n)$ under a unit step input satisfies
\[
\ddot{y}(t) + 2\zeta \omega_n \dot{y}(t) + \omega_n^{2} y(t) = 1,\qquad y(0)=0,\;\dot{y}(0)=0.
\]
We define the functional response as
\[
f(x,t) \;=\; y(t;\zeta,\omega_n),
\]
so the objective measures the displacement curve’s deviation from a desired target response $f^{\ast}(t)$.

The second oracle is the \textbf{susceptible–infected–recovered (SIR) epidemic model}, a standard tool in epidemiology for characterizing disease spread. The design vector is $x=(\beta,\gamma,I_0)\in\mathbb R^3$, and the index is time $\lambda=t$. With initial conditions $S(0)=1-I_0$, $I(0)=I_0$ and $R(0)=0$, the dynamics are
\[
\frac{dS}{dt}=-\beta SI,\qquad \frac{dI}{dt}=\beta SI-\gamma I,\qquad \frac{dR}{dt}=\gamma I.
\]
The functional response is defined as
\[
f(x,t) \;=\; I(t;\beta,\gamma,I_0),
\]
the infection trajectory whose shape is critical in public health modeling.

The third oracle is the \textbf{Lotka–Volterra predator–prey system}, a nonlinear model from ecology describing interactions between predator and prey populations. The design vector is $x=(\alpha,\beta,\delta,\gamma)\in\mathbb R^4$, and the index is time $\lambda=t$. With initial conditions $x(0)=y(0)=1$, the dynamics are
\[
\frac{dx}{dt}=\alpha x-\beta xy,\qquad \frac{dy}{dt}=\delta xy-\gamma y.
\]
The functional response is defined as
\[
f(x,t) \;=\; x(t;\alpha,\beta,\delta,\gamma),
\]
capturing prey population trajectories across time.

The fourth oracle is the \textbf{one–dimensional heat diffusion model}, a partial differential equation (PDE) that is fundamental in thermal engineering and materials science. The design vector is $x=(\kappa,L,T_L,T_R,q,\alpha,\beta)\in\mathbb R^7$, with $\lambda=t$ indexing time. On $x\in[0,L]$, the temperature satisfies
\[
\frac{\partial u}{\partial t} = \kappa \frac{\partial^2 u}{\partial x^2}+q,\qquad u(0,t)=T_L,\;u(L,t)=T_R,
\]
with initial condition $u(x,0)=\alpha+\beta\sin(\pi x/L)$. The functional response is taken at mid depth,
\[
f(x,t) \;=\; u(L/2,t;\kappa,L,T_L,T_R,q,\alpha,\beta),
\]
yielding a one-dimensional time series that reflects transient thermal behavior.

\begin{figure*}[t]
\centering
\begin{subfigure}[b]{.24\textwidth}
  \includegraphics[width=\linewidth]{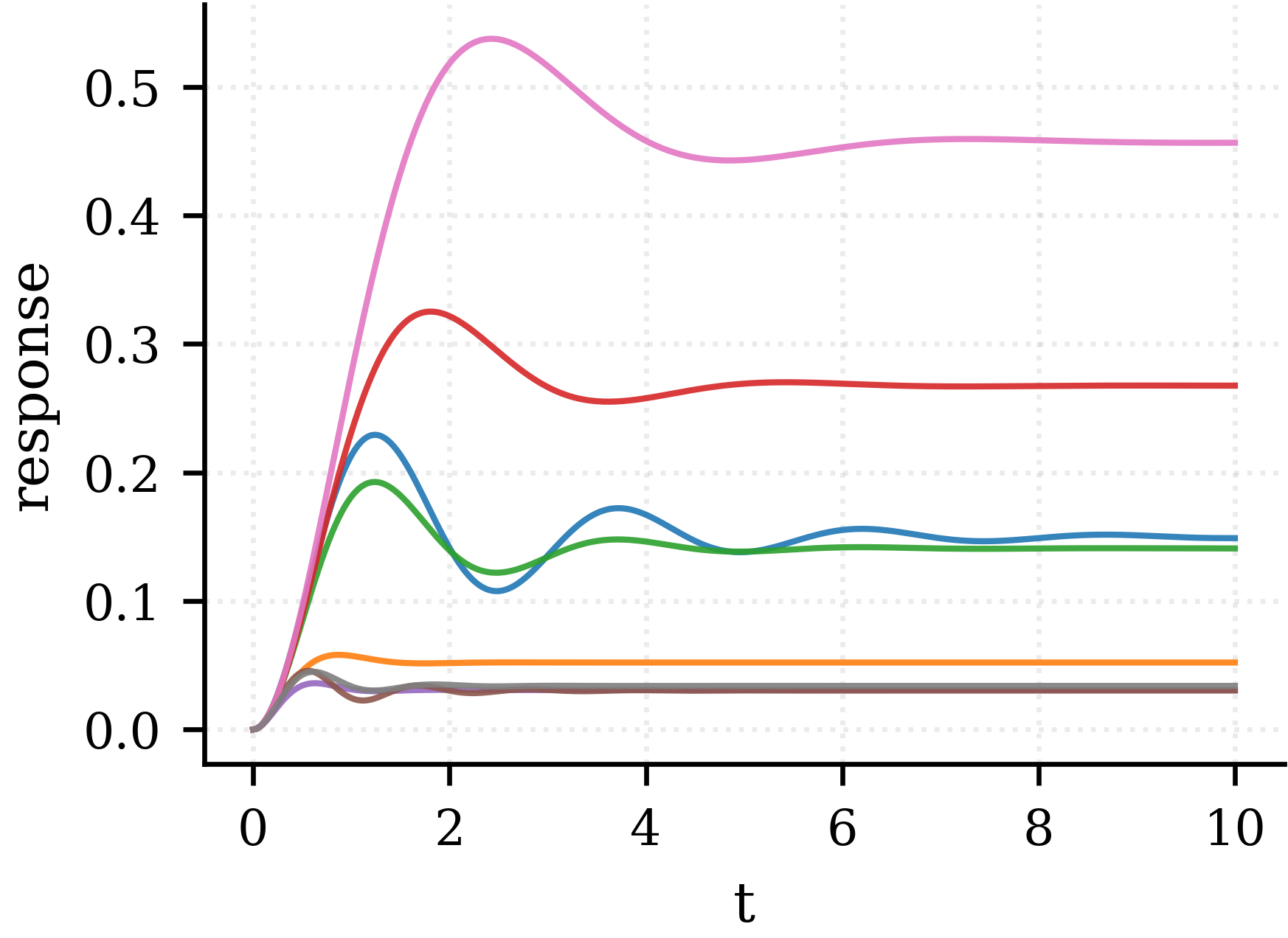}
  \caption{Mass spring damper}
\end{subfigure}\hfill
\begin{subfigure}[b]{.24\textwidth}
  \includegraphics[width=\linewidth]{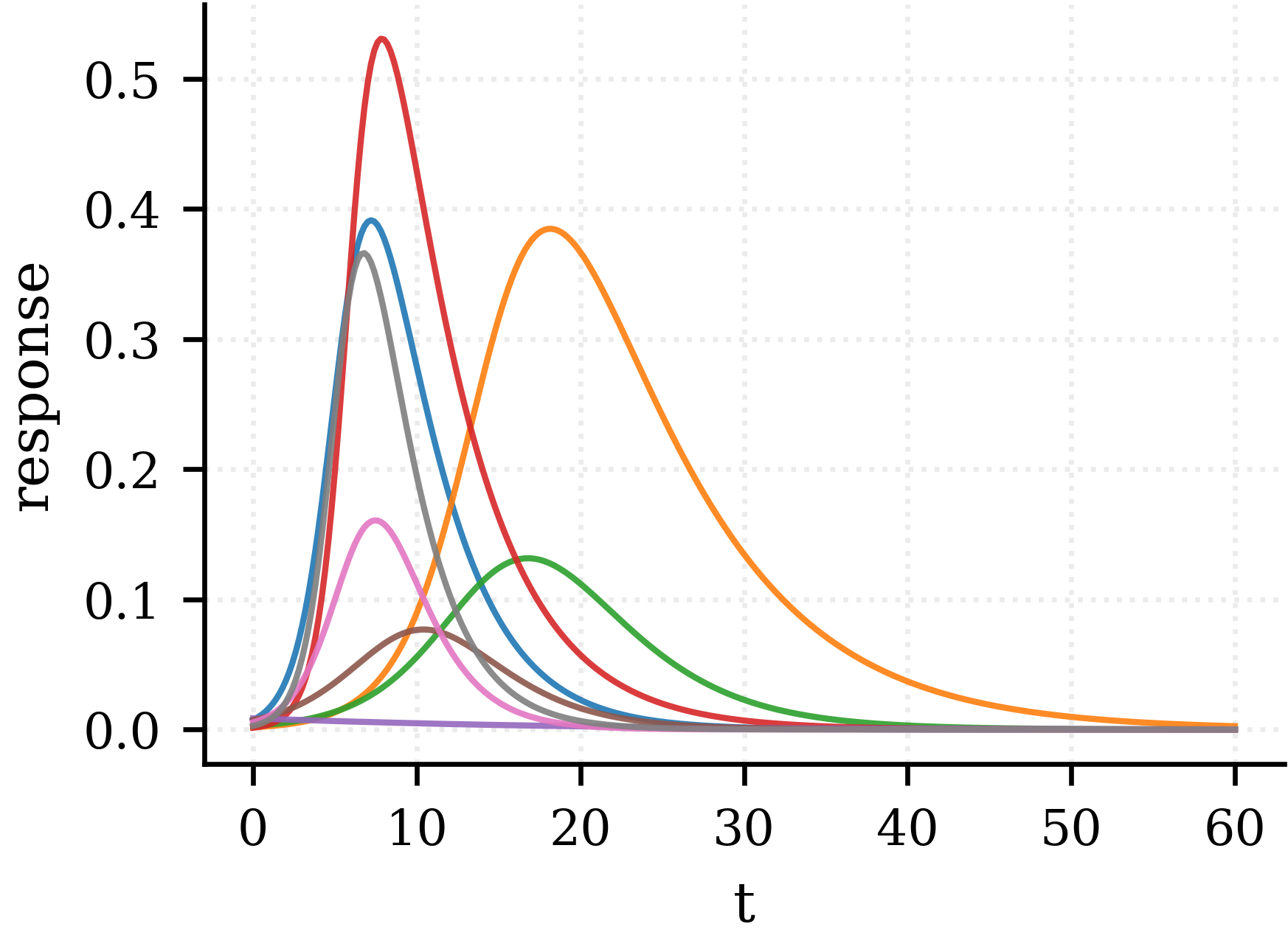}
  \caption{SIR}
\end{subfigure}\hfill
\begin{subfigure}[b]{.24\textwidth}
  \includegraphics[width=\linewidth]{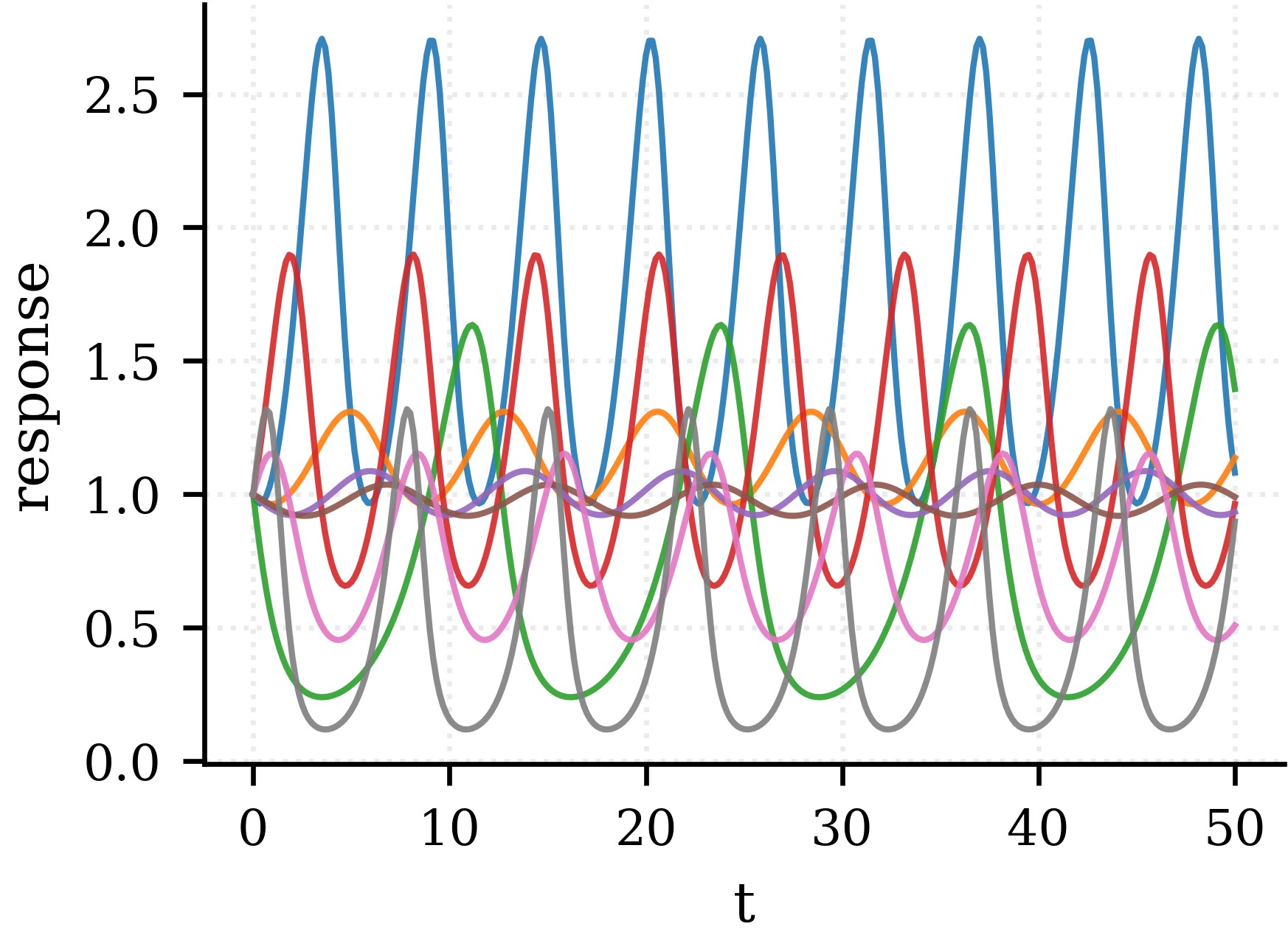}
  \caption{Lotka Volterra}
\end{subfigure}\hfill
\begin{subfigure}[b]{.24\textwidth}
  \includegraphics[width=\linewidth]{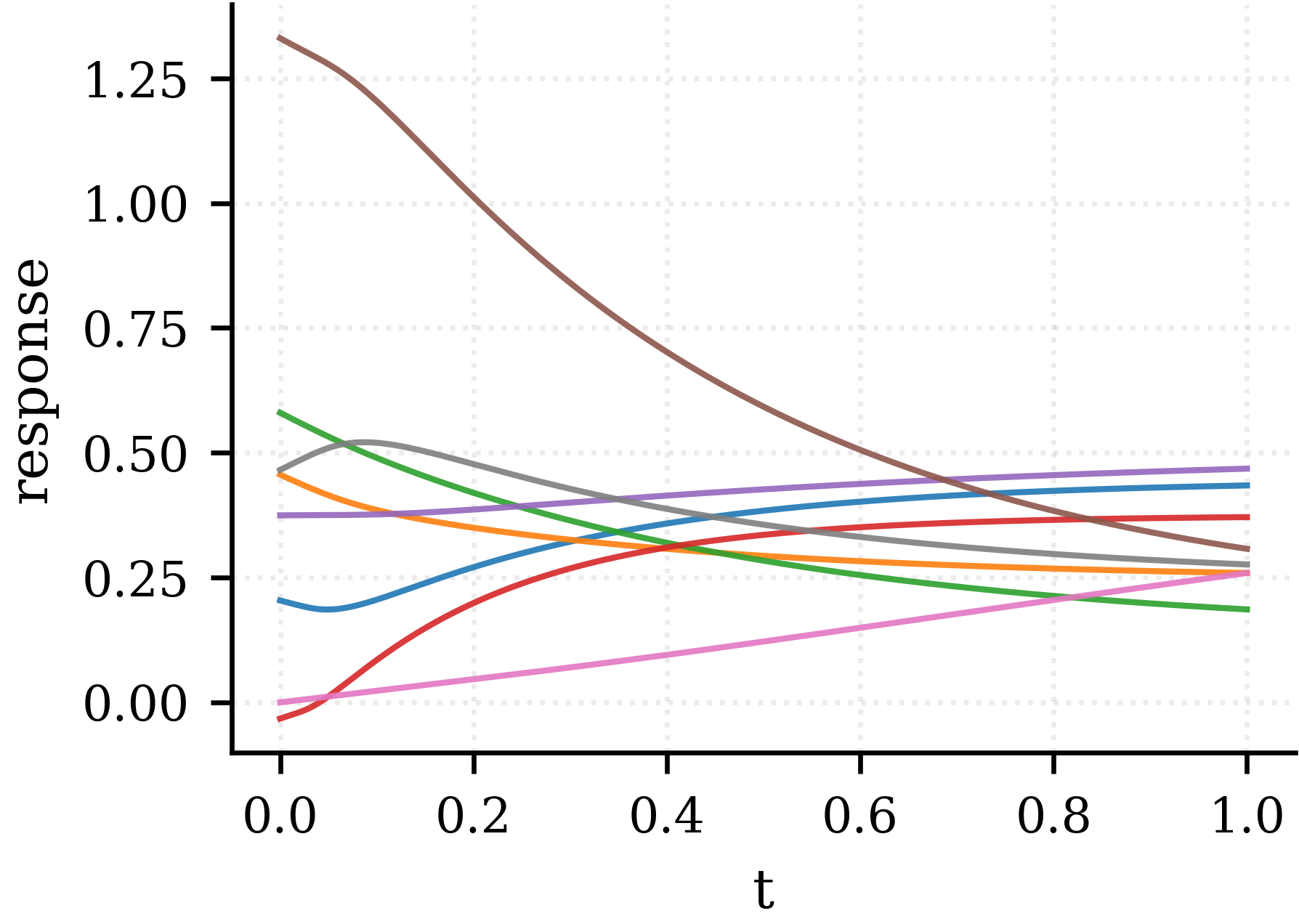}
  \caption{Heat diffusion}
\end{subfigure}
\caption{Representative trajectories drawn from each simulator. Each panel shows multiple random draws of the response over time to illustrate functional variability across the design space.}
\label{fig:functional_samples}
\end{figure*}

Together these oracles test the method across diverse mathematical structures: linear ordinary differential equations (ODEs), nonlinear compartmental models, nonlinear predator–prey dynamics, and parabolic PDEs, with varying dimensionality and temporal complexity. We visualize typical responses from each oracle. Figure~\ref{fig:functional_samples} shows multiple random trajectories for each system across the time domain, illustrating the variety of shapes and scales that motivate a worst-case objective.

All experiments share a common protocol. A budget of $m=50$ sequential evaluations is allowed after an initial set of $n_{0}$ seeds obtained via Latin hypercube sampling. At each iteration we form a candidate set by combining a global Sobol pool with a local pool around the current best design. Candidates are first ranked by an exploitation score, after which the integrated uncertainty acquisition based on the mean and standard deviation of the squared deviation process selects the next evaluation, with occasional pure exploitation steps to accelerate progress. We compare against two standard baselines: a Gaussian process fit directly to the scalar objective $g(x)$ with expected improvement, and a progressive space-filling design that is not adaptive. Each configuration is run for $R=50$ independent replications with shared seeds and candidate pools, enabling paired comparisons. Performance is assessed through four complementary metrics: regret trajectories, normalized regret trajectories, final regret distributions, and AUOC. To further quantify sample efficiency, we also report the time to threshold (TT) defined as the iteration at which regret first falls below $\epsilon$.

\begin{figure*}[t]
\centering
\begin{subfigure}[b]{.24\textwidth}
  \includegraphics[width=\linewidth]{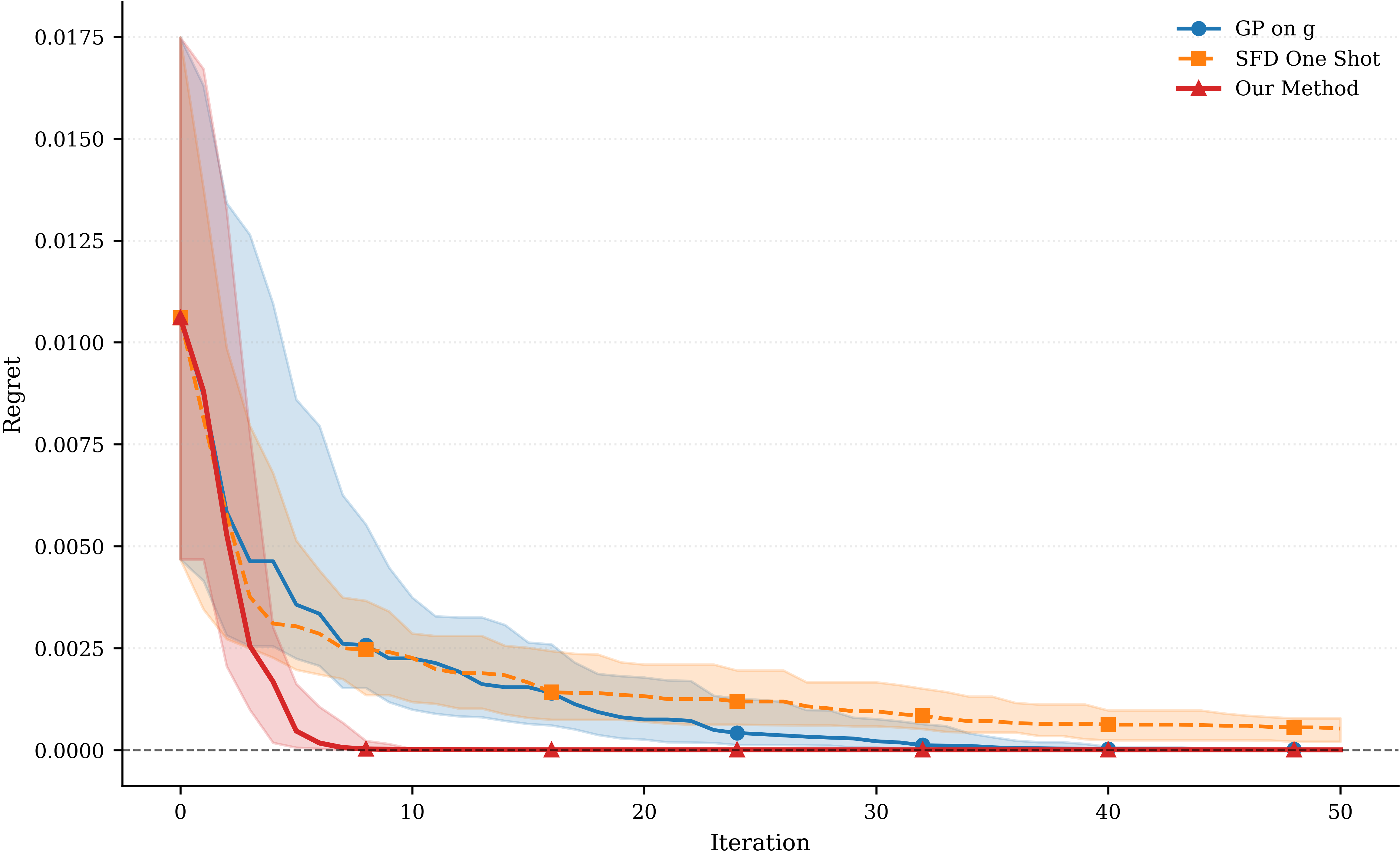}
  \caption{Mass-spring damper}
\end{subfigure}\hfill
\begin{subfigure}[b]{.24\textwidth}
  \includegraphics[width=\linewidth]{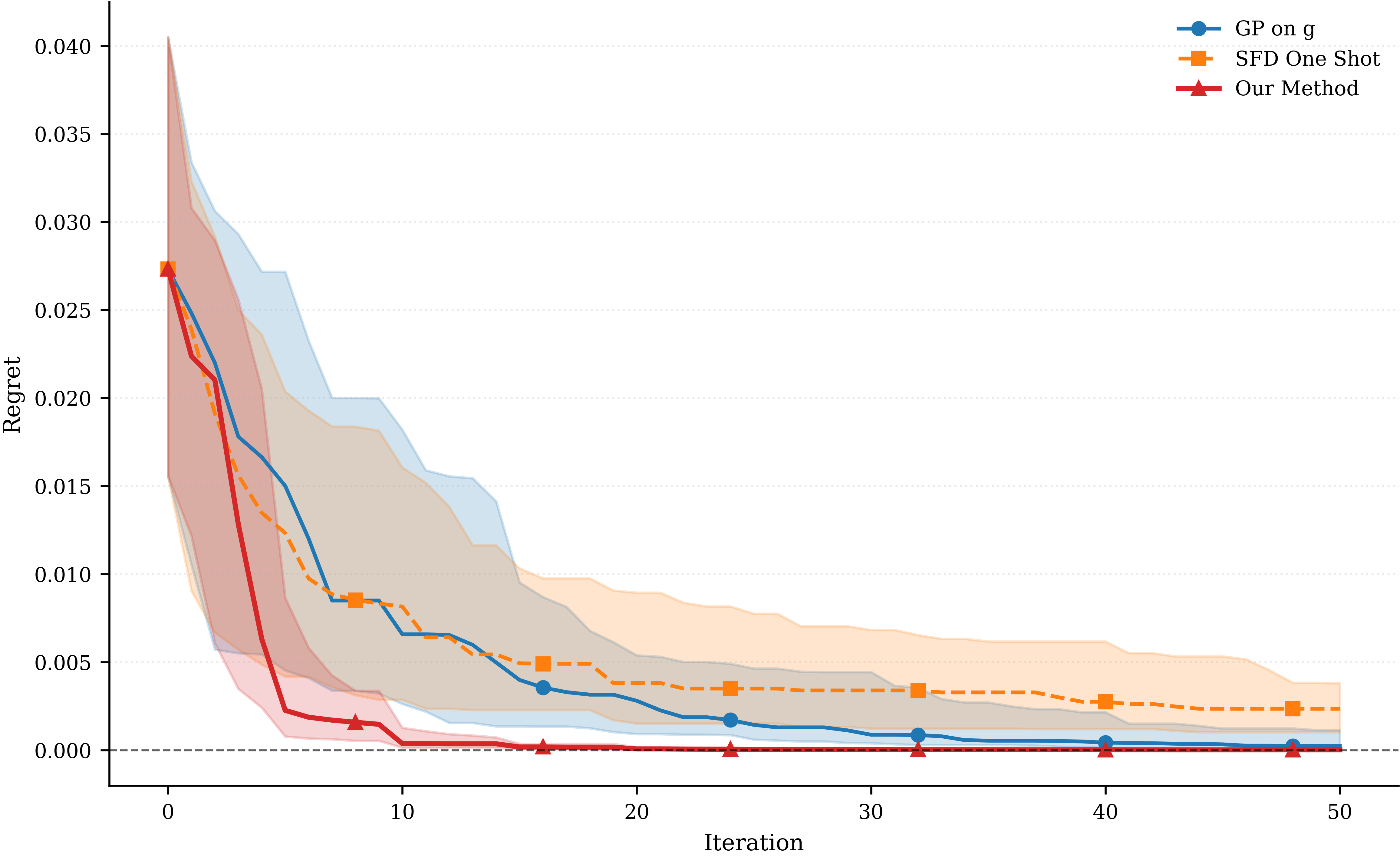}
  \caption{SIR}
\end{subfigure}\hfill
\begin{subfigure}[b]{.24\textwidth}
  \includegraphics[width=\linewidth]{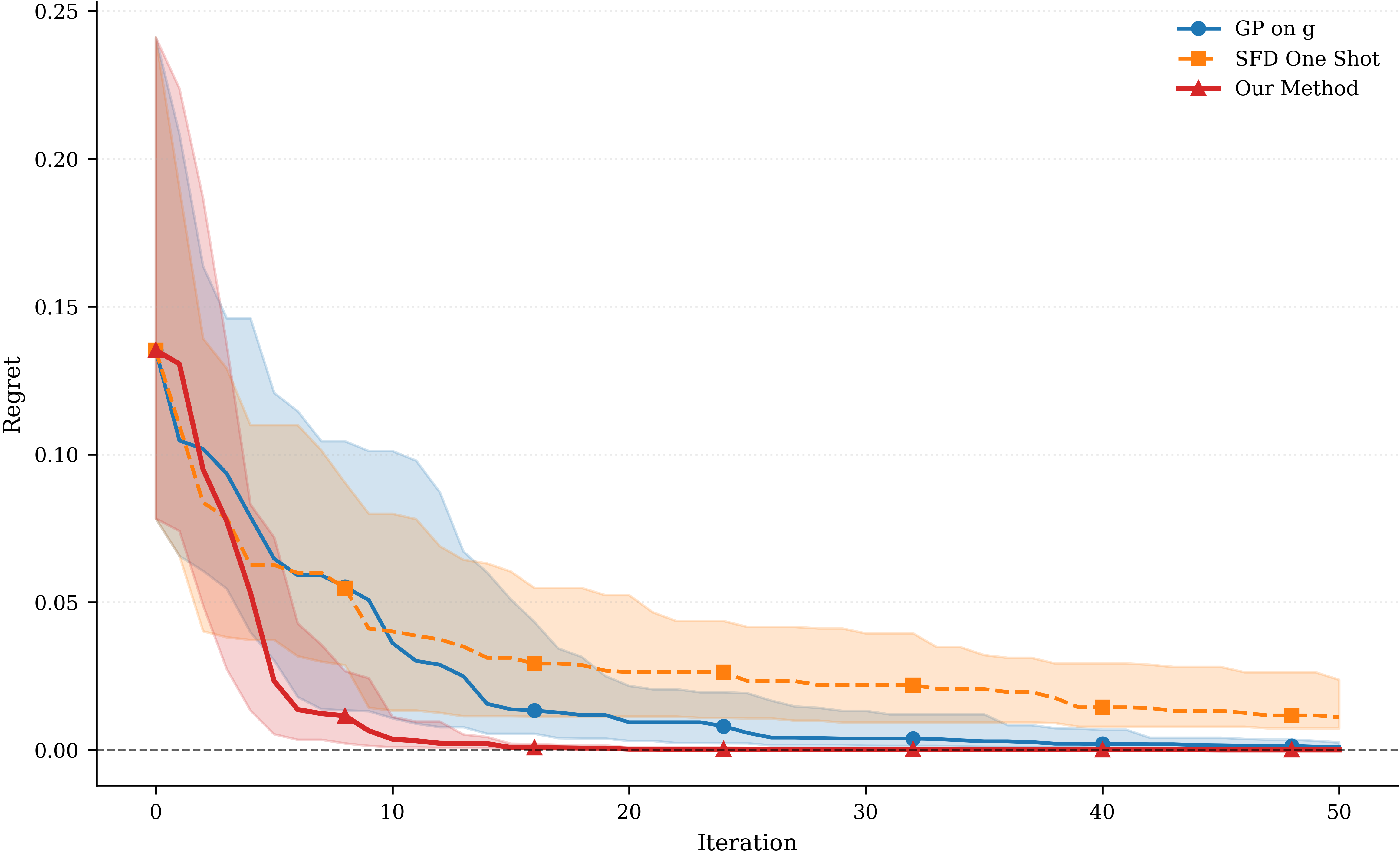}
  \caption{Lotka-Volterra}
\end{subfigure}\hfill
\begin{subfigure}[b]{.24\textwidth}
  \includegraphics[width=\linewidth]{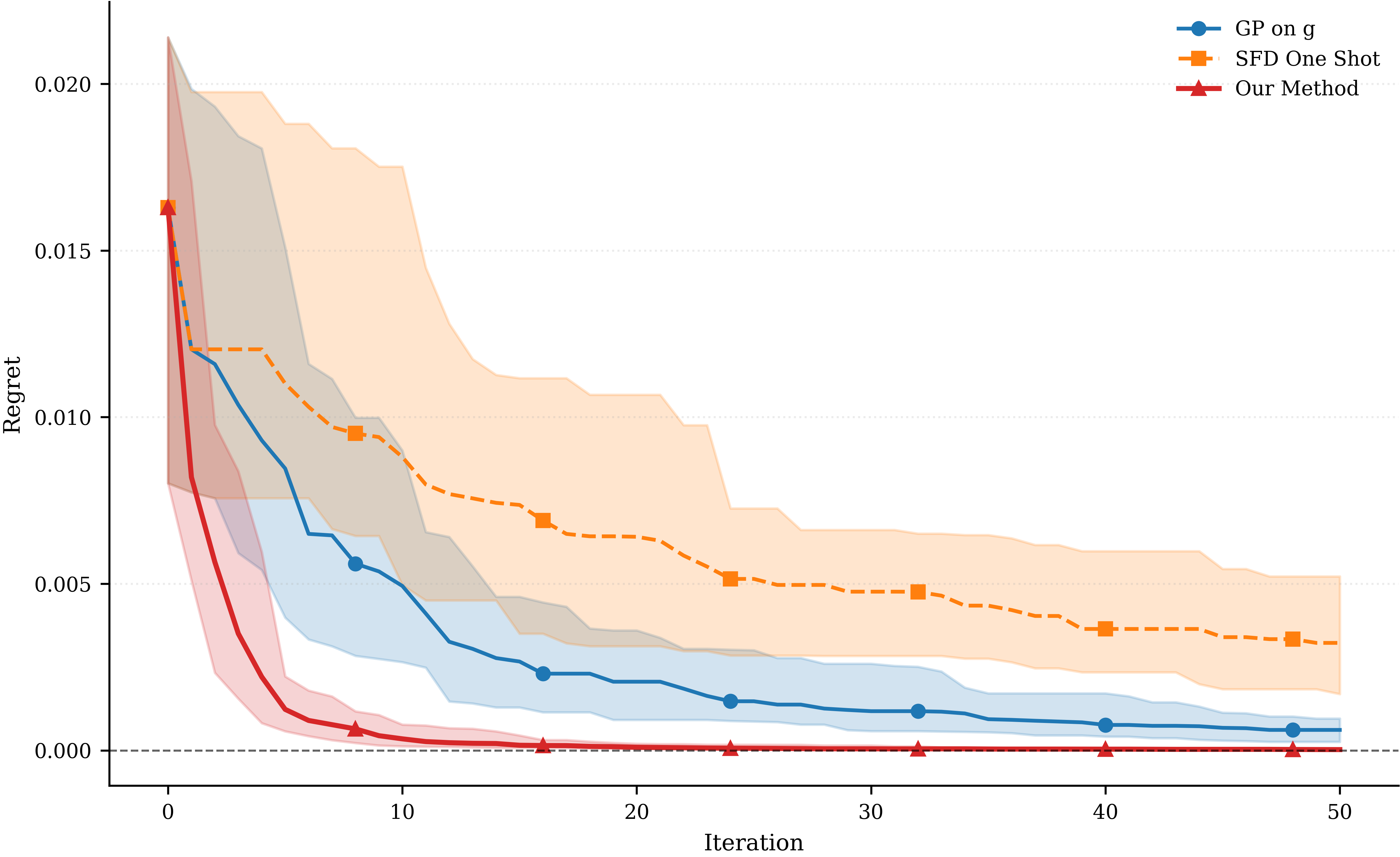}
  \caption{Heat diffusion}
\end{subfigure}
\caption{Regret versus iteration with median and interquartile bands.}
\label{fig:sim_regret}
\end{figure*}

\begin{figure*}[t]
\centering
\begin{subfigure}[b]{.24\textwidth}
  \includegraphics[width=\linewidth]{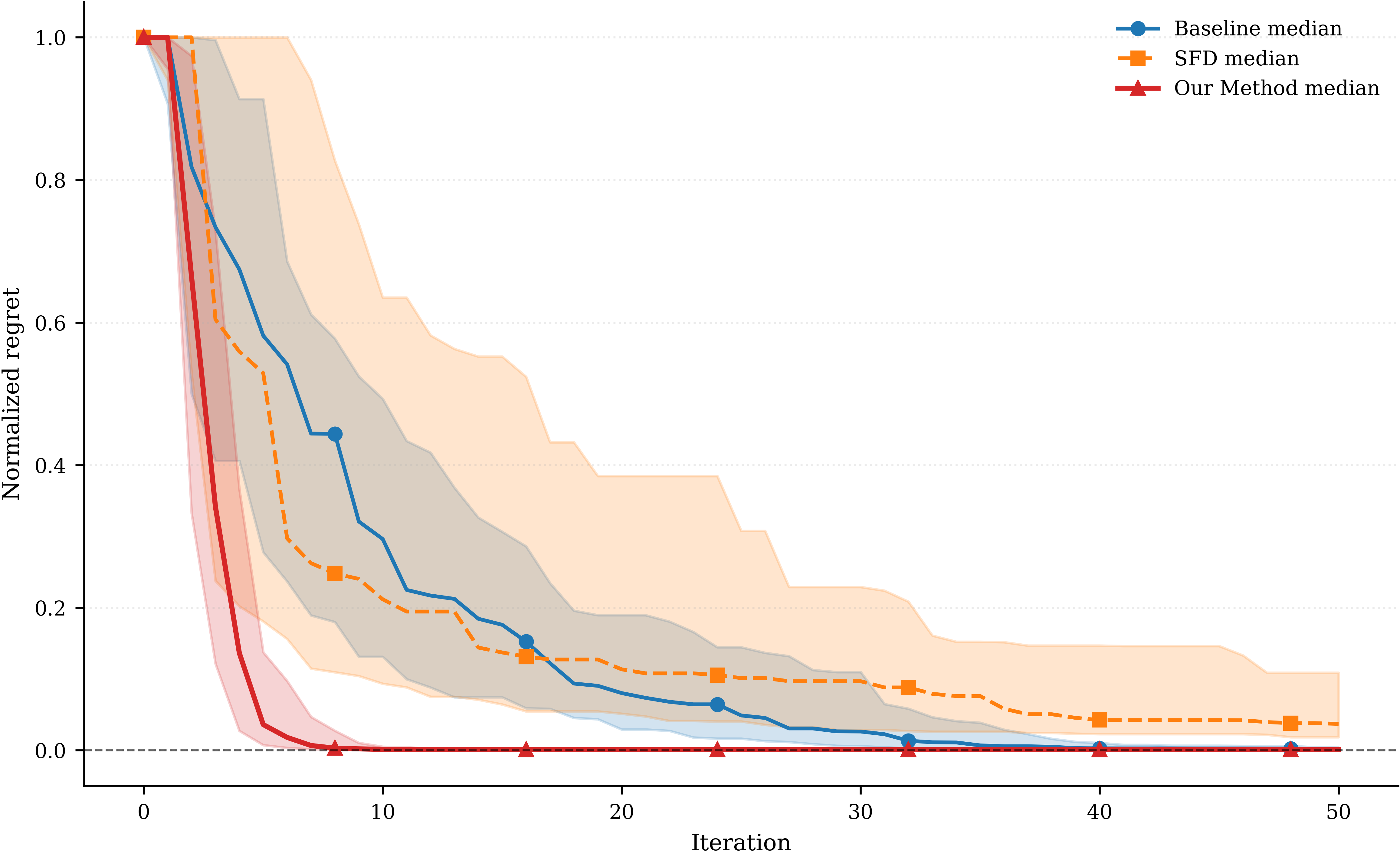}
  \caption{Mass-spring damper}
\end{subfigure}\hfill
\begin{subfigure}[b]{.24\textwidth}
  \includegraphics[width=\linewidth]{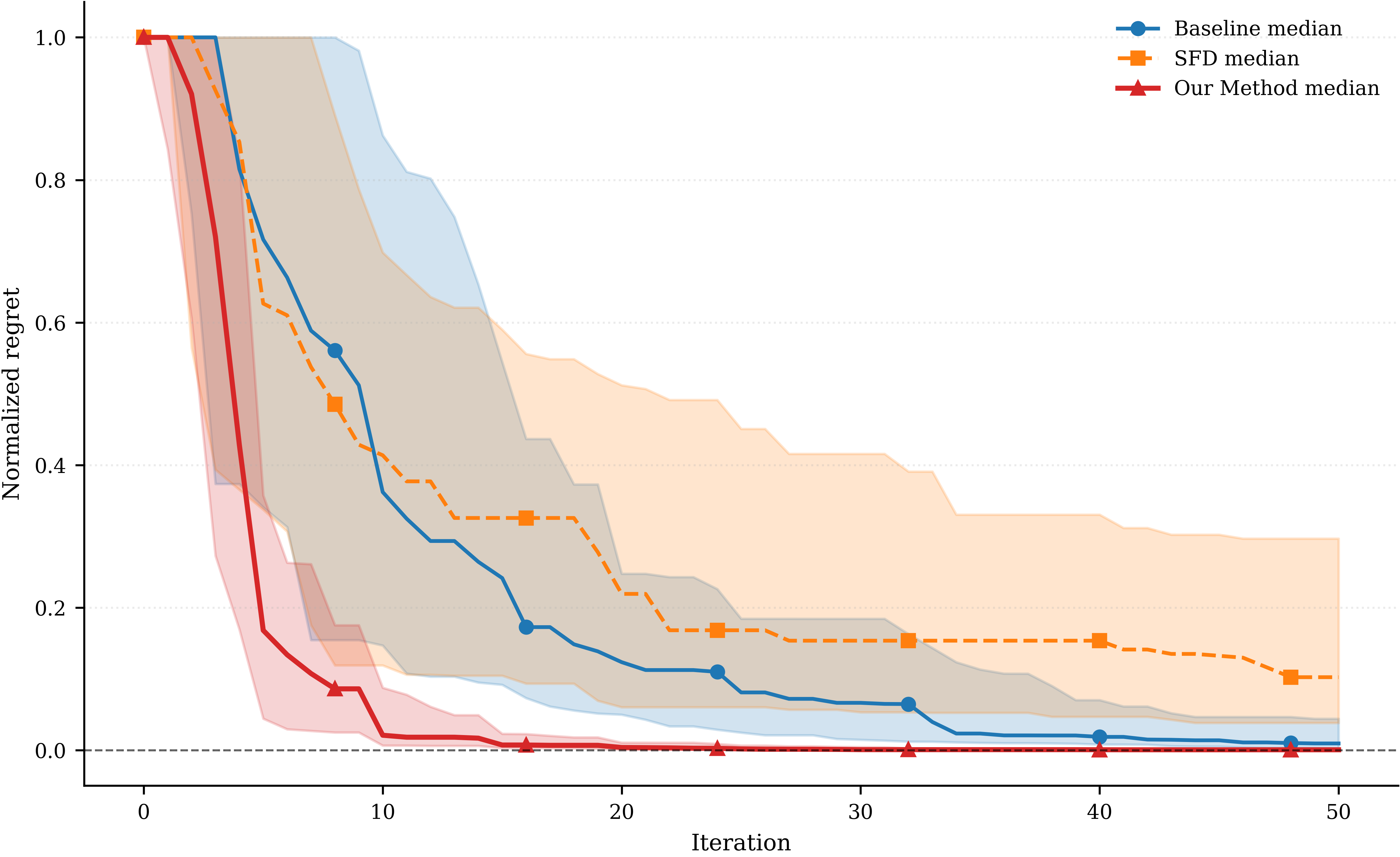}
  \caption{SIR}
\end{subfigure}\hfill
\begin{subfigure}[b]{.24\textwidth}
  \includegraphics[width=\linewidth]{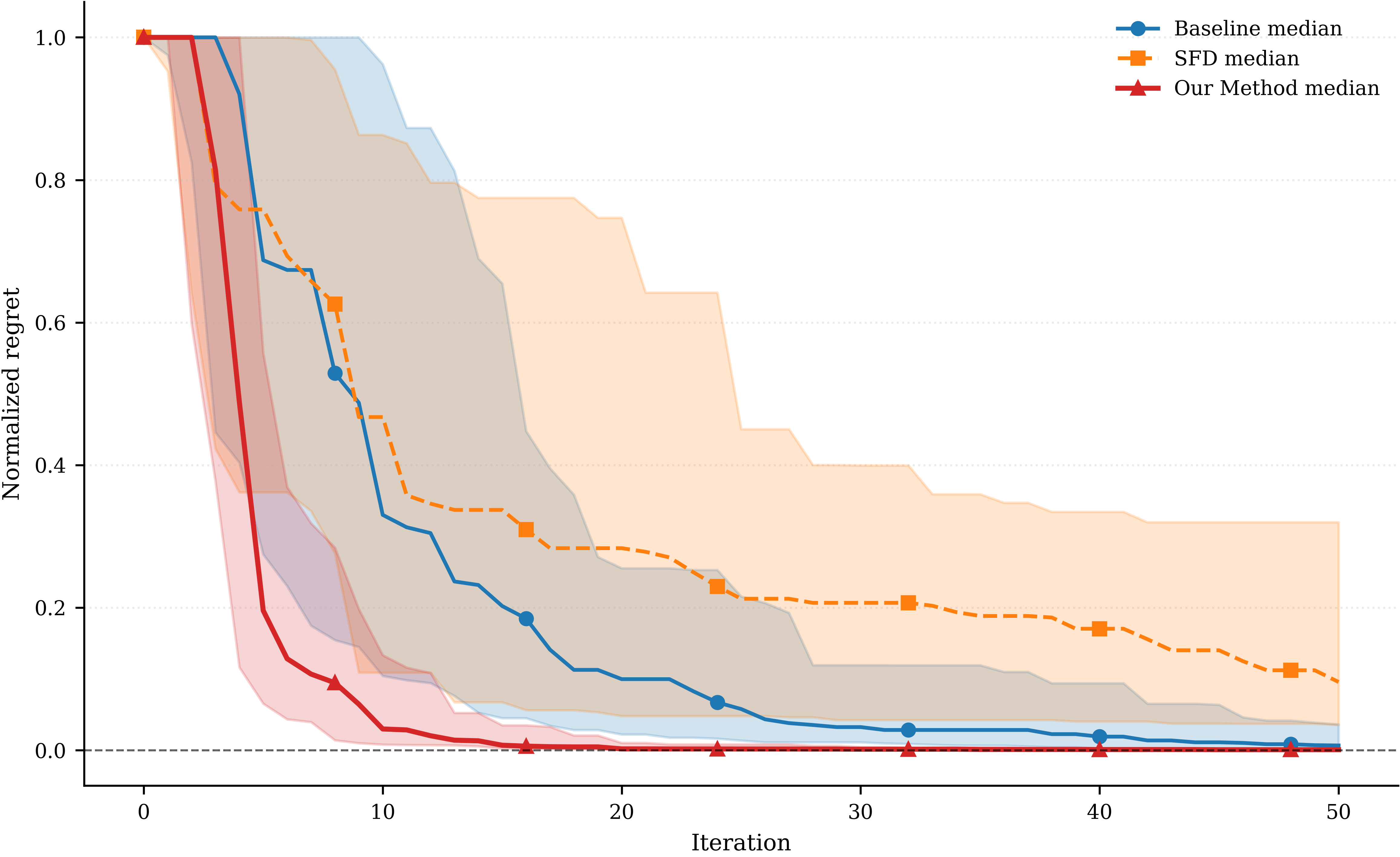}
  \caption{Lotka-Volterra}
\end{subfigure}\hfill
\begin{subfigure}[b]{.24\textwidth}
  \includegraphics[width=\linewidth]{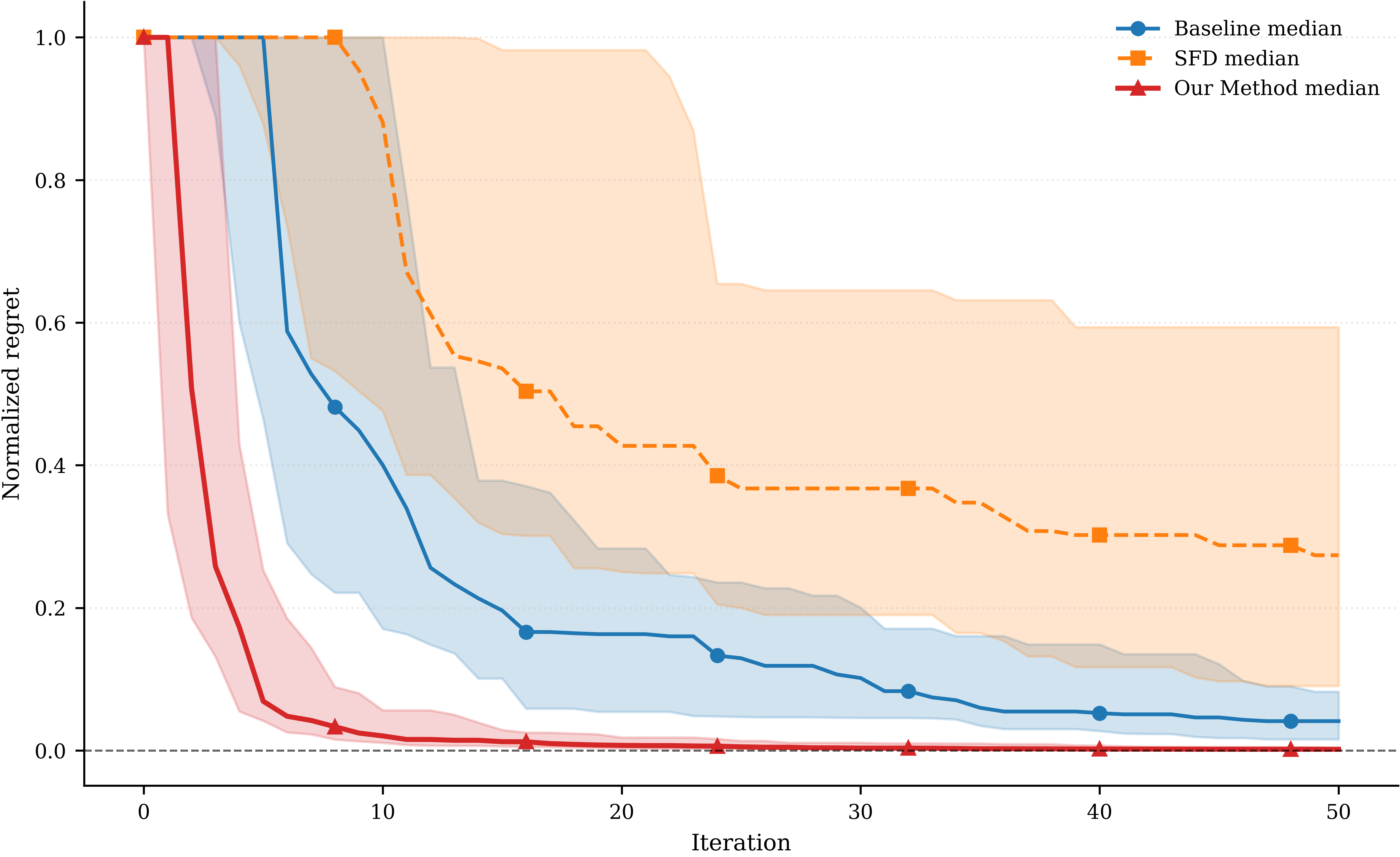}
  \caption{Heat diffusion}
\end{subfigure}
\caption{Normalized regret trajectories showing convergence speed and stability.}
\label{fig:sim_regret_norm}
\end{figure*}

\begin{figure*}[t]
\centering
\begin{subfigure}[b]{.24\textwidth}
  \includegraphics[width=\linewidth]{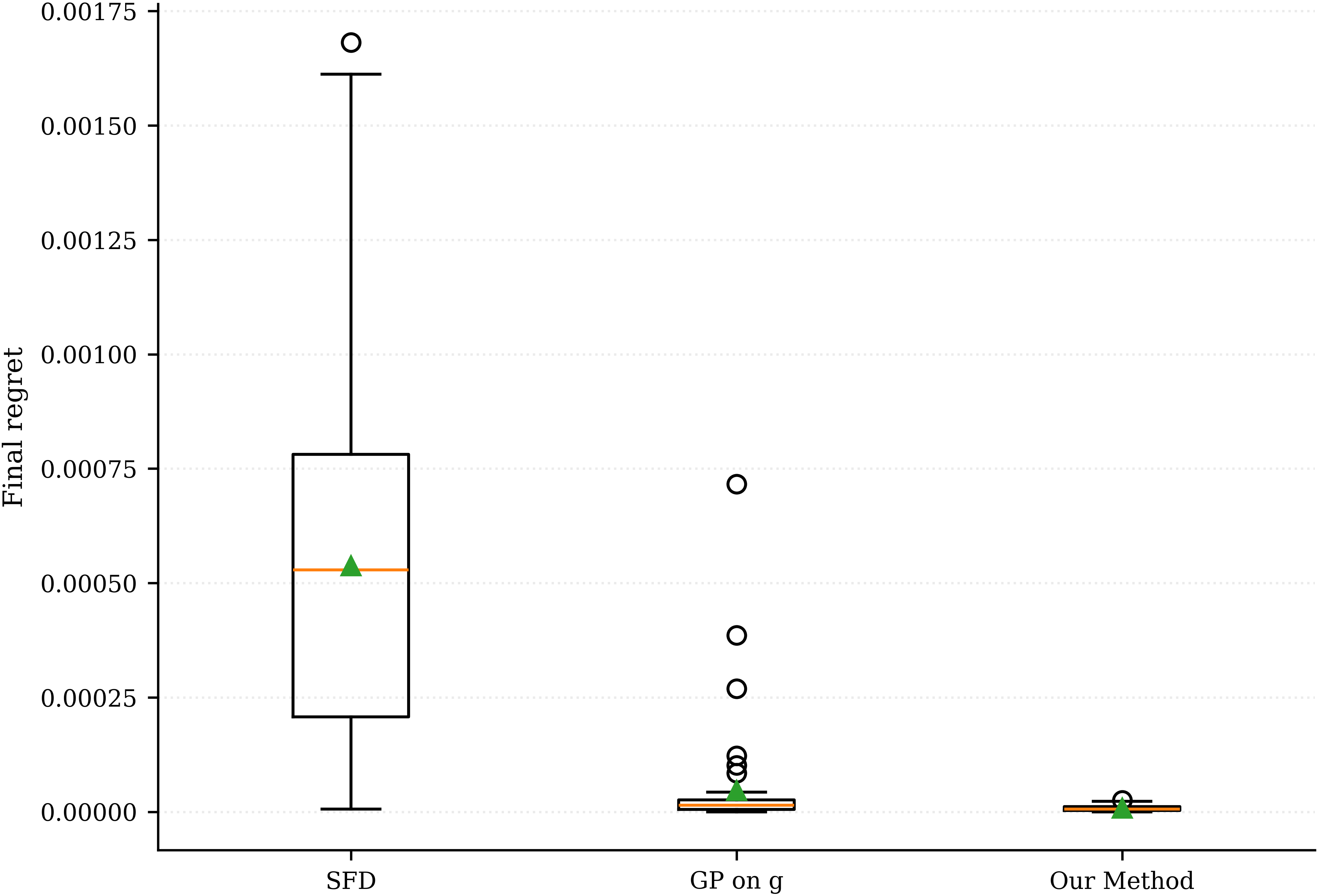}
  \caption{Mass-spring damper}
\end{subfigure}\hfill
\begin{subfigure}[b]{.24\textwidth}
  \includegraphics[width=\linewidth]{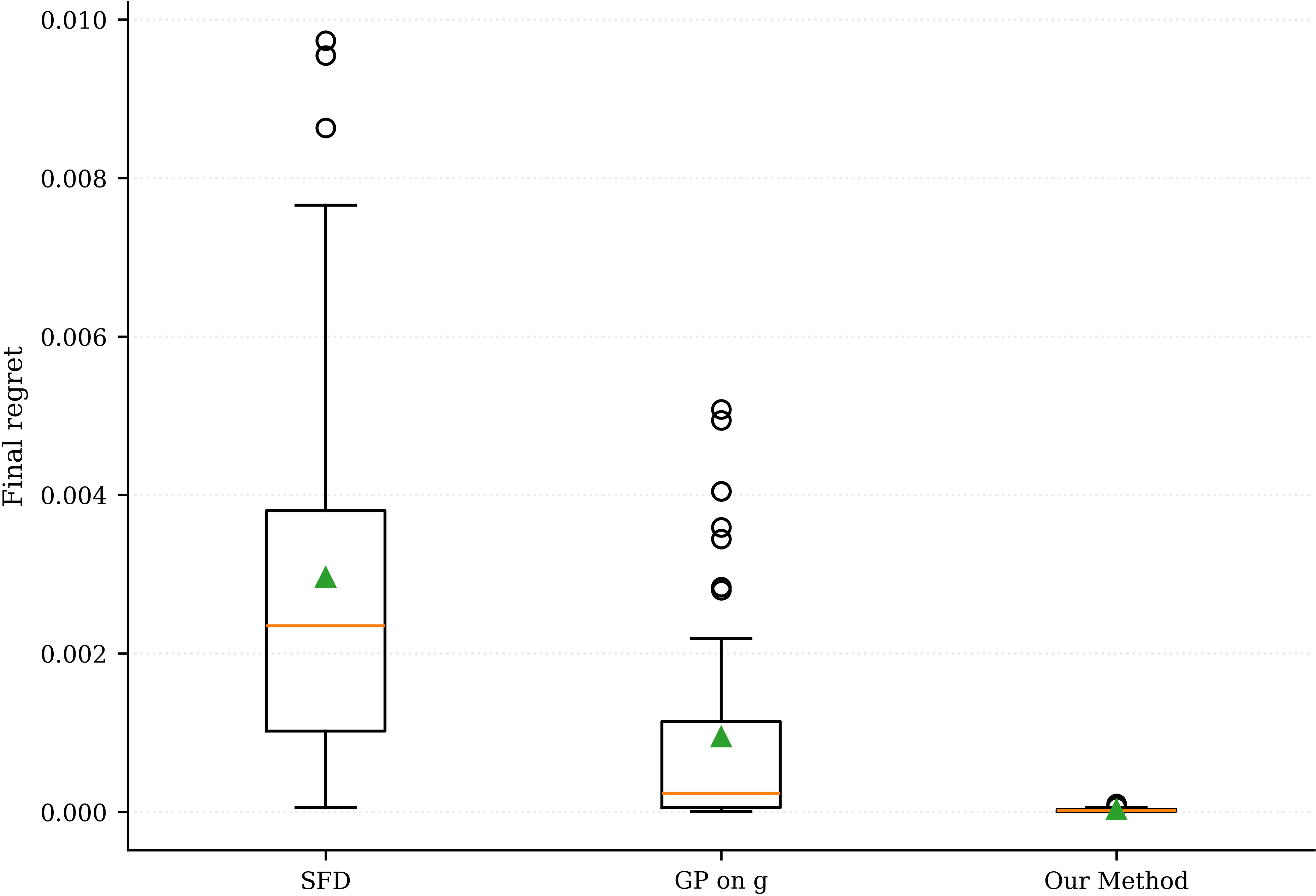}
  \caption{SIR}
\end{subfigure}\hfill
\begin{subfigure}[b]{.24\textwidth}
  \includegraphics[width=\linewidth]{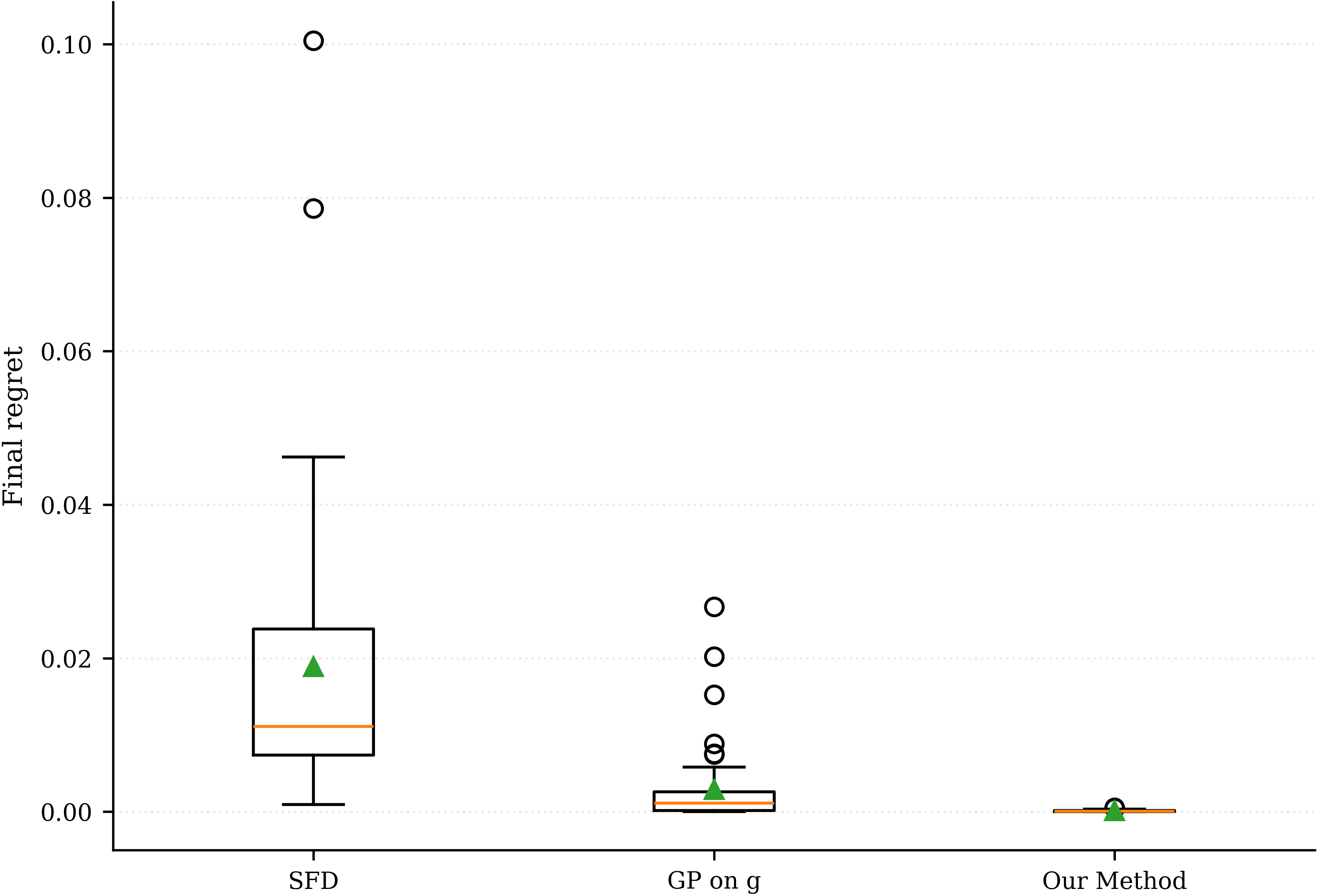}
  \caption{Lotka-Volterra}
\end{subfigure}\hfill
\begin{subfigure}[b]{.24\textwidth}
  \includegraphics[width=\linewidth]{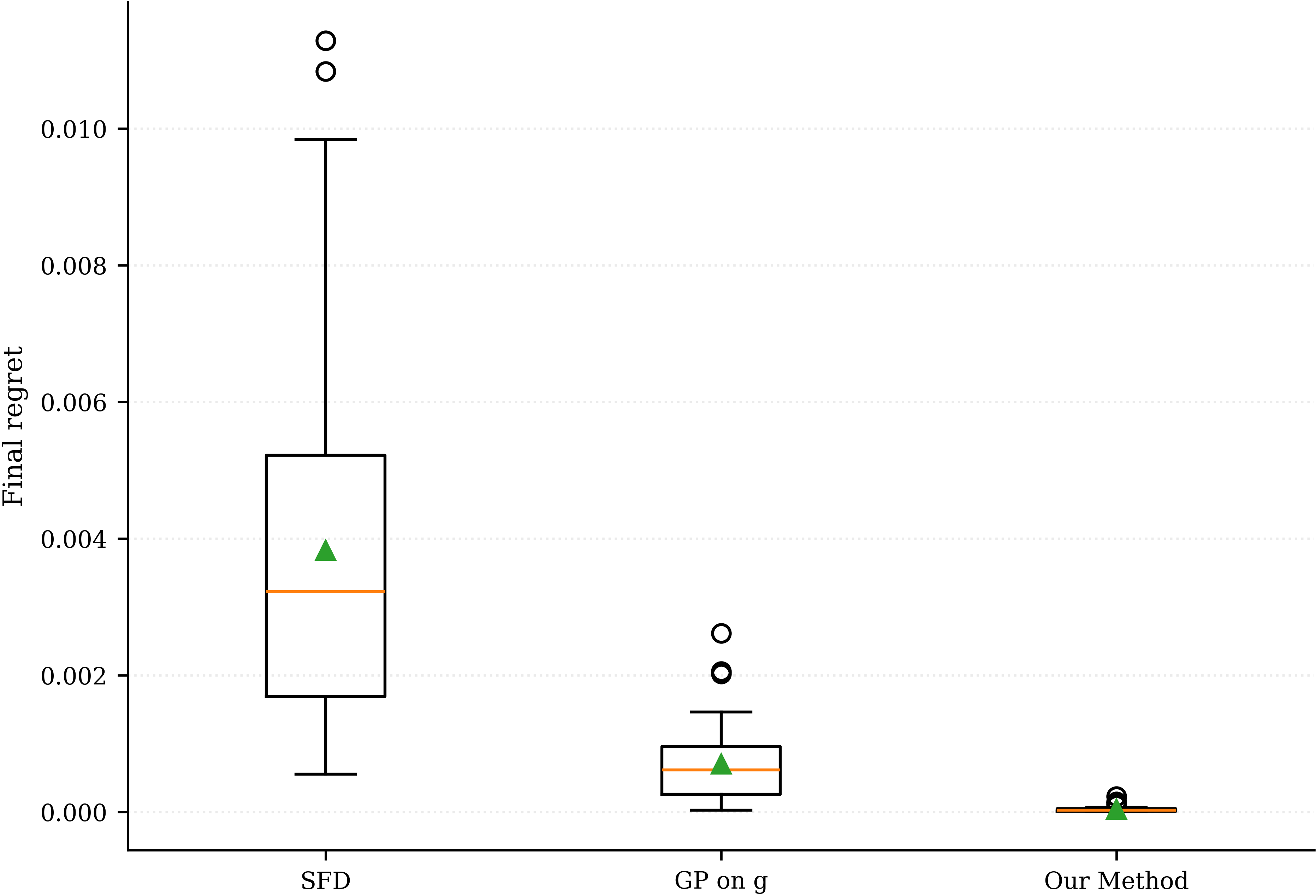}
  \caption{Heat diffusion}
\end{subfigure}
\caption{Box plots of final regret distributions across replications.}
\label{fig:sim_final}
\end{figure*}

\begin{figure*}[t]
\centering
\begin{subfigure}[b]{.24\textwidth}
  \includegraphics[width=\linewidth]{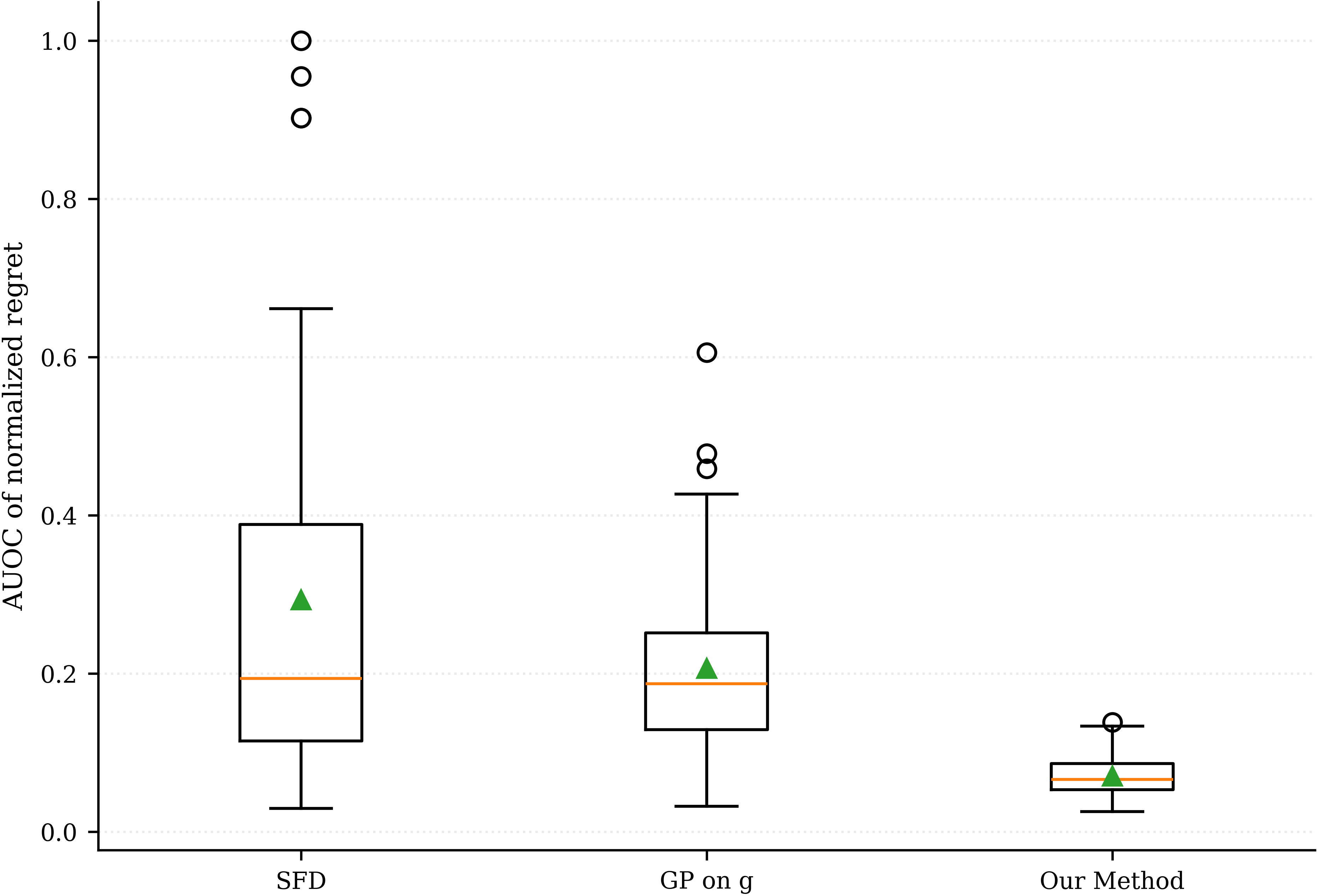}
  \caption{Mass-spring damper}
\end{subfigure}\hfill
\begin{subfigure}[b]{.24\textwidth}
  \includegraphics[width=\linewidth]{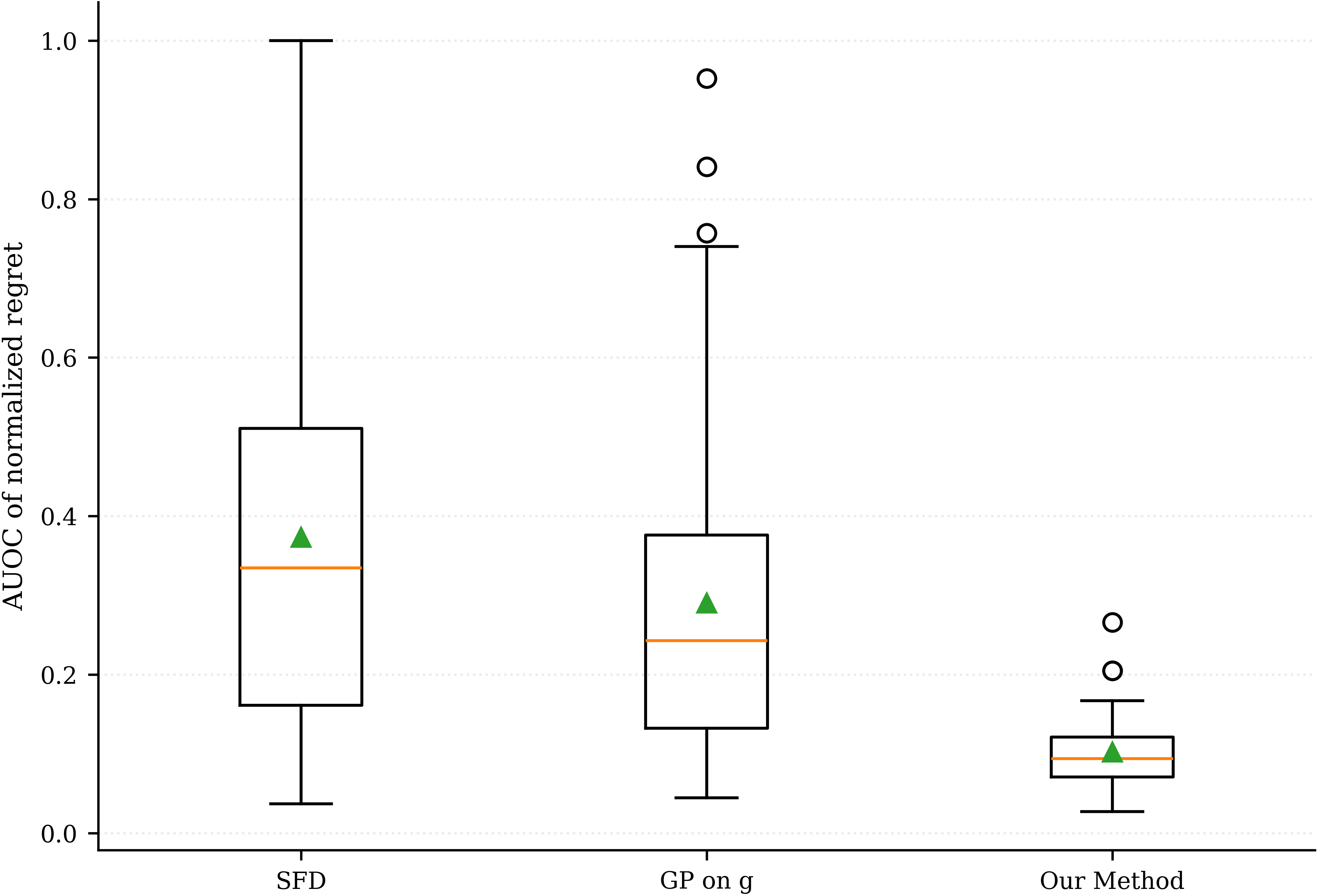}
  \caption{SIR}
\end{subfigure}\hfill
\begin{subfigure}[b]{.24\textwidth}
  \includegraphics[width=\linewidth]{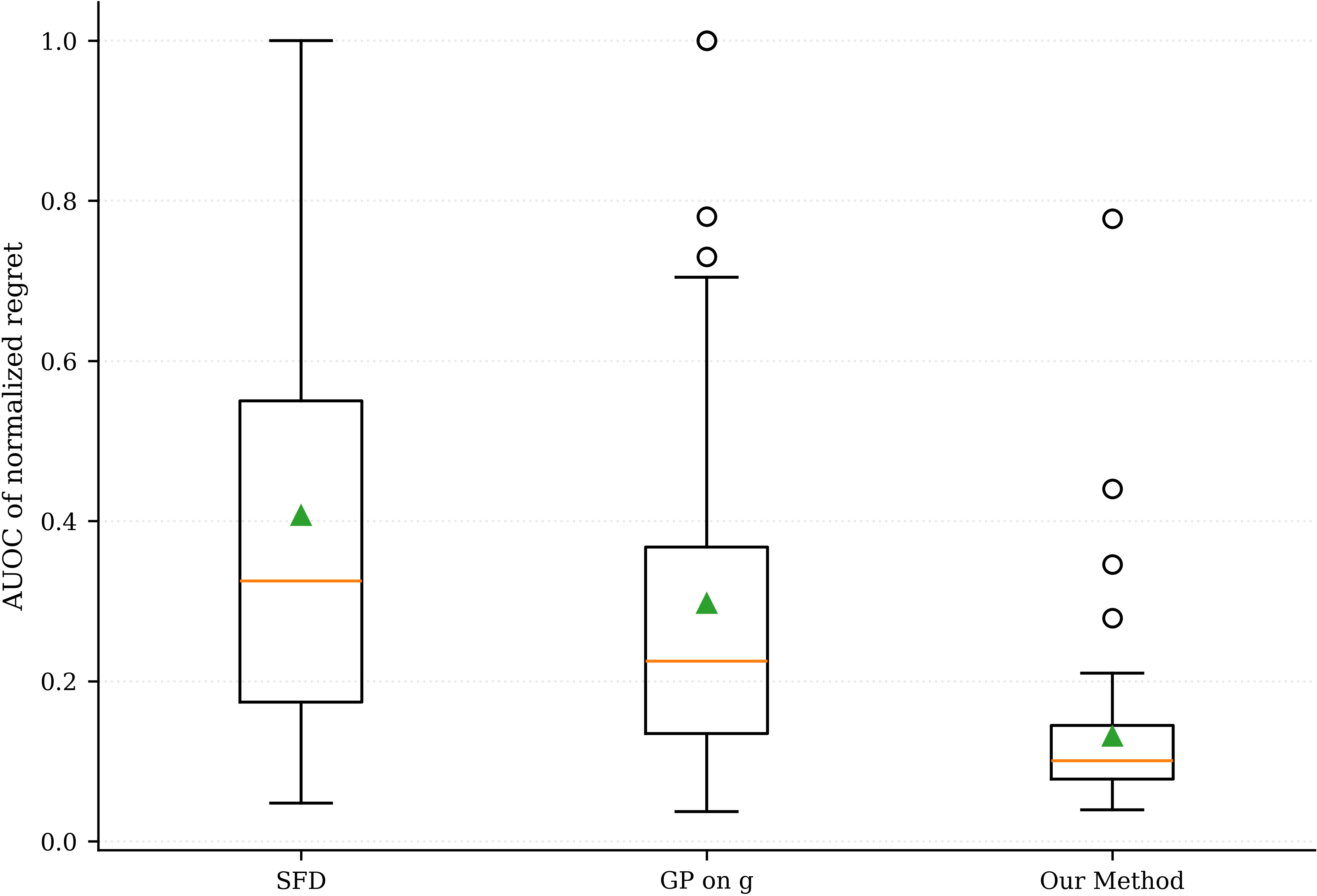}
  \caption{Lotka-Volterra}
\end{subfigure}\hfill
\begin{subfigure}[b]{.24\textwidth}
  \includegraphics[width=\linewidth]{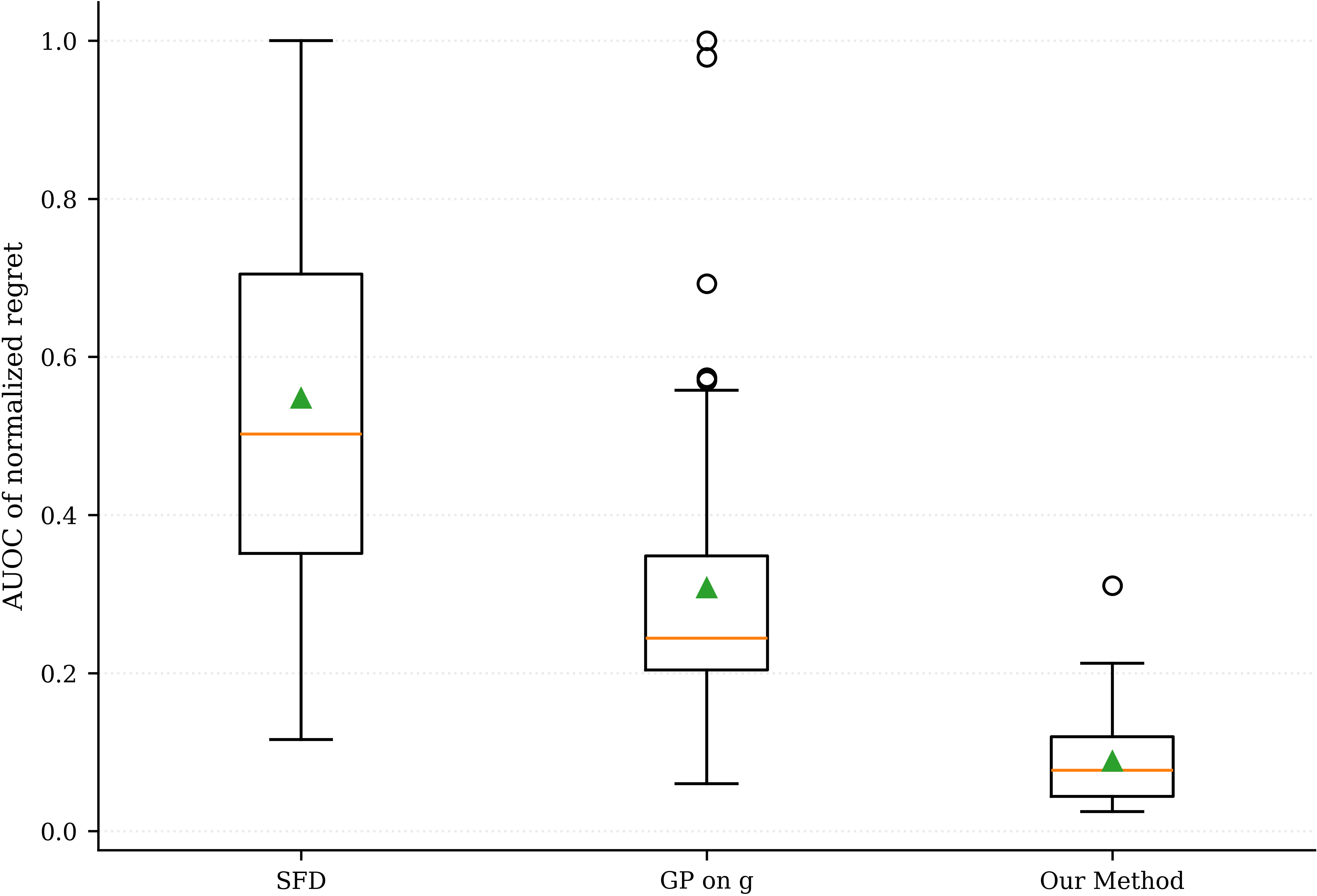}
  \caption{Heat diffusion}
\end{subfigure}
\caption{AUOC distributions summarizing full regret trajectories.}
\label{fig:sim_auoc}
\end{figure*}

The results are summarized in four panels. Figure~\ref{fig:sim_regret} presents regret versus iteration across the four problems. Our method consistently converges faster and reaches lower regret values compared with both baselines. The improvement is most pronounced in the mass–spring–damper and heat-diffusion systems, where clear separation appears within the first few iterations and is maintained throughout. In the SIR and Lotka–Volterra systems the advantage is still evident, but emerges more gradually as the dynamics introduce oscillatory or nonlinear interactions that pose greater challenges for scalar approaches. Figure~\ref{fig:sim_regret_norm} reports the normalized regret trajectories, which emphasize stability across replications. Normalization makes clear that our method not only converges earlier but also exhibits substantially tighter interquartile bands, reflecting more reliable performance across independent runs. Figure~\ref{fig:sim_final} shows the distribution of final regret after the evaluation budget is exhausted. Across all four problems, our method achieves markedly smaller medians and interquartile spreads, confirming that the best-found solutions are consistently closer to the reference function. Complementing this perspective, Figure~\ref{fig:sim_auoc} summarizes the area under the normalized regret curve (AUOC), which integrates performance over the entire trajectory. Here again our method achieves the smallest values across problems, underscoring superior sample efficiency. Finally, Tables~\ref{tab:tte10} and \ref{tab:tte05} report the time to threshold (TT) metric for $\epsilon=0.10$ and $\epsilon=0.05$, respectively. Each entry shows the fraction of successful runs together with the median iteration required to cross the threshold. The results demonstrate that our method reaches both thresholds in nearly all replications and does so in far fewer iterations than either baseline.

\begin{table}[t]
\centering
\caption{Time to threshold (TT) results for $\epsilon=0.10$. Each entry shows fraction of successful runs and median iteration.}
\label{tab:tte10}
\begin{tabular}{lccc}
\toprule
Problem & SFD & GP on g & MM-FBO \\
\midrule
Mass spring damper & 0.74 (15) & 0.98 (18) & 1.00 (5) \\
SIR              & 0.48 (14) & 0.84 (20) & 1.00 (7) \\
Lotka Volterra     & 0.52 (11) & 0.86 (18) & 0.98 (8) \\
Heat diffusion     & 0.32 (25) & 0.80 (25) & 1.00 (5) \\
\bottomrule
\end{tabular}
\end{table}

\begin{table}[t]
\centering
\caption{Time to threshold (TT) results for $\epsilon=0.05$. Each entry shows fraction of successful runs and median iteration.}
\label{tab:tte05}
\begin{tabular}{lccc}
\toprule
Problem & SFD & GP on g & MM-FBO \\
\midrule
Mass spring damper & 0.60 (23) & 0.94 (25) & 1.00 (5) \\
SIR              & 0.30 (19) & 0.78 (25) & 1.00 (10) \\
Lotka Volterra     & 0.34 (10) & 0.82 (20) & 0.98 (10) \\
Heat diffusion     & 0.12 (23) & 0.62 (29) & 0.96 (6) \\
\bottomrule
\end{tabular}
\end{table}

Across all metrics and problems, the proposed method is consistently superior. It converges faster in terms of regret trajectories, is more stable under normalization, achieves substantially smaller final regret, and exhibits lower AUOC values that confirm improved sample efficiency. The TT results further show that thresholds of both $\epsilon=0.10$ and $\epsilon=0.05$ are reached reliably and in far fewer iterations. Particularly in challenging problems such as heat diffusion and Lotka-Volterra, where functional responses display sharp features and oscillations, the benefits of balancing squared deviation with integrated uncertainty are most pronounced.

\subsection{Case Studies}

We next ground the methodology in physics-inspired case studies where the response is a function of wavelength $\lambda$ or time $t$, and the design vector $x$ collects geometric or process parameters. In both settings, the objective remains to minimize the worst case squared deviation from a prescribed target response. For the metasurface scattering problem, the target is the flat scattering spectrum $f^{\ast}(\lambda)\equiv 1$, reflecting the goal of broadband scattering with high efficiency as the high response. For the vapor phase infiltration process, the target $f^{\ast}(t)$ is a reference infiltration profile obtained from physical considerations, where fidelity to the entire curve rather than any single time point is critical. Throughout, we denote by $\Lambda=\{\lambda_{1},\ldots,\lambda_{T}\}$ or $\{t_{1},\ldots,t_{T}\}$ the discretized functional domain and by $x\subset\mathbb{R}^{p}$ a bounded hyper-rectangle of admissible designs.

\begin{figure*}[t]
\centering
\includegraphics[width=0.6\linewidth]{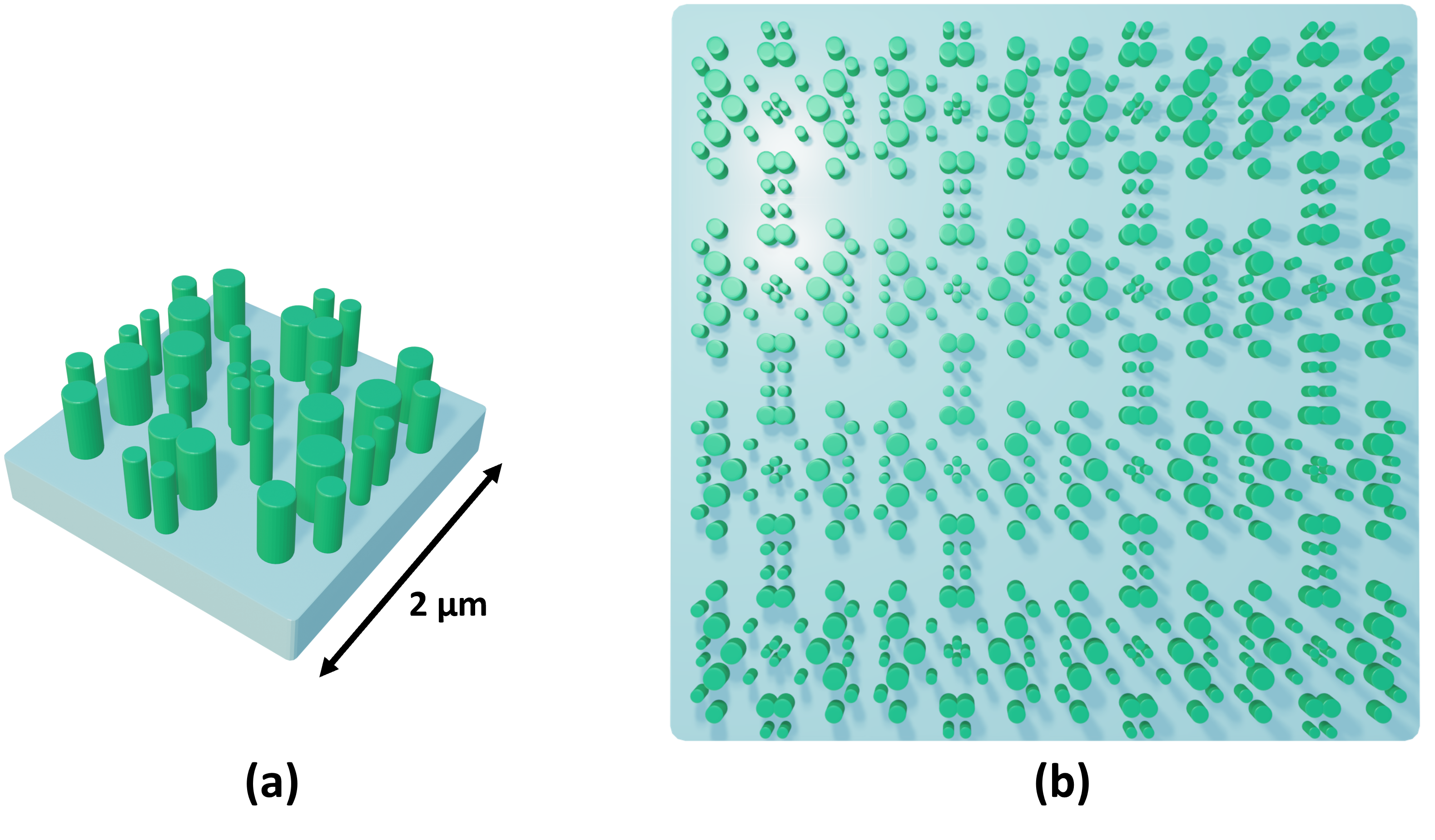}
\caption{Schematic of a metasurface generated through a basis-driven geometric parameterization. The structure is designed to maximize the broadband scattering cross-section and is composed of 200 nm-thick dielectric nanopillars on a silicon oxide (SiO$_2$) substrate. (a) Isometric view of the nanopillars within a 2~$\mu$m~$\times$~2~$\mu$m supercell. (b) Top view of the full metasurface, where the radius and lateral position of each nanopillar are modulated by a smooth basis function. This basis is defined by a compact set of geometric coefficients ($x$), enabling a low-dimensional representation of an otherwise complex, large-area design.}
\label{fig:metasurface}
\end{figure*}

\paragraph{Case Study 1: Design of Metasurfaces for Wideband Scattering in Photonics.}
This study considers photonic metasurfaces engineered to control an optical wavefront via subwavelength geometric patterning. The selected metasurface is a periodic structure with a supercell formed by \(16 \times 16\) cylindrical rods with arbitrary radii (larger than a minimum value assigned by fabrication limitations) located in a regular grid, as shown in Figure~\ref{fig:metasurface} (a). This results in an excessive number \((256)\) of design parameters for the radii of the rods. To reduce this number, we use a Fourier like parameterization of the pillar radius profile over the in plane coordinates \((x,y)\) of a unit cell with periods \((\mathrm{period}_x,\mathrm{period}_y)\). Writing \(\omega_x = \tfrac{2\pi}{\mathrm{period}_x}\) and 
\(\omega_y = \tfrac{2\pi}{\mathrm{period}_y}\), a representative 
truncated expansion is
\begin{align*}
R(u,v;\boldsymbol{c}) &= r_0 \big[A_0
  + A_1\sin(\omega_x u+\omega_y v)
  + A_2\cos(\omega_x u+\omega_y v) \\
  &\quad + A_3\sin\big(2(\omega_x u+\omega_y v)\big)
  + A_4\cos\big(2(\omega_x u+\omega_y v)\big) + \cdots \big],
\end{align*}
where $(u,v)$ denote the in-plane spatial coordinates of the unit 
cell, and the amplitudes \(\{A_\ell\}\) and base scale \(r_0\) are 
components of the design vector \(\boldsymbol{c} \in \mathcal{X}\). Together with additional layout parameters (for example unit cell period, aspect controls, and thickness) these coefficients define the super cell geometry from which the scattering spectrum \(H(\lambda;x)\) is generated by a full wave electromagnetic solver (see Figure~\ref{fig:metasurface} (b) for a visualization). This parameterization is based on the recently demonstrated GiBS framework, which represents the entire device geometry using a compact set of coefficients from smooth parametric bases (for example Fourier or Chebyshev) \citep{marzban2025gibs}. This input side representation drastically compresses the design space for full-wave optimization methods such as Bayesian optimization.

The goal of the inverse design is to find these geometrical parameters that define the shape of the metasurface with the best response. Each candidate design \(x \in \mathcal{X} \subset \mathbb{R}^{14}\) specifies a set of geometric coefficients that modulate a periodic array of nano pillars as shown in Figure~\ref{fig:metasurface}, producing a scattering spectrum \(H(\lambda;x)\) over a dense grid of wavelengths. Our target response is a broadband highly efficient uniform scattering spectrum; accordingly, we define the loss at design \(x\) as
\[
g(x)\;=\;\max_{\lambda \in \Lambda}\,\big(H(\lambda;x)-1\big)^{2},
\qquad \text{with } f^{\star}(\lambda)=1.
\]

We assembled a dataset of $N=4000$ designs, each paired with a measured/simulated spectrum sampled at $T=201$ wavelengths, i.e., $\{(x_i,\{H(\lambda_t;x_i)\}_{t=1}^{T})\}_{i=1}^{N}$. The wavelength grid $\Lambda$ spans the band of interest with increasing resolution where resonant structure is anticipated. Bounds for each coordinate of $x$ define a box-constrained domain $\mathcal{X}=[\ell_1,u_1]\times\cdots\times[\ell_{14},u_{14}]$ used in all optimization runs. For each candidate design we computed the scattering spectrum on a dense wavelength grid across the band of interest $\Lambda$, with design variables sampled uniformly within the admissible bounds described above. All full-wave electromagnetic simulations were performed using three-dimensional finite-difference time-domain (3D FDTD) technique (implemented using the commercial software Lumerical) on periodic supercell geometries of Figure~\ref{fig:metasurface} (b).

Because the dataset is fixed, we emulate the oracle $H(\lambda;x)$ with a fast, differentiable surrogate trained once on all $(x,\lambda)$ pairs and then used for on-the-fly evaluations. Concretely, we fit a low-rank separable neural model
\[
\widehat{H}(\lambda;x)\;=\;a(x)^{\top}b(\lambda),
\]
in which $a:x\!\to\!\mathbb{R}^{r}$ and $b:\Lambda\!\to\!\mathbb{R}^{r}$ are shallow networks and $\lambda$ is encoded with Fourier features to capture oscillatory structure. Inputs $x$ are standardized, wavelengths are scaled to $[0,1]$, and responses are standardized during training and mapped back to physical units at inference time.

\begin{figure*}[t]
\centering
\begin{subfigure}[b]{.24\textwidth}
  \centering\includegraphics[width=\linewidth]{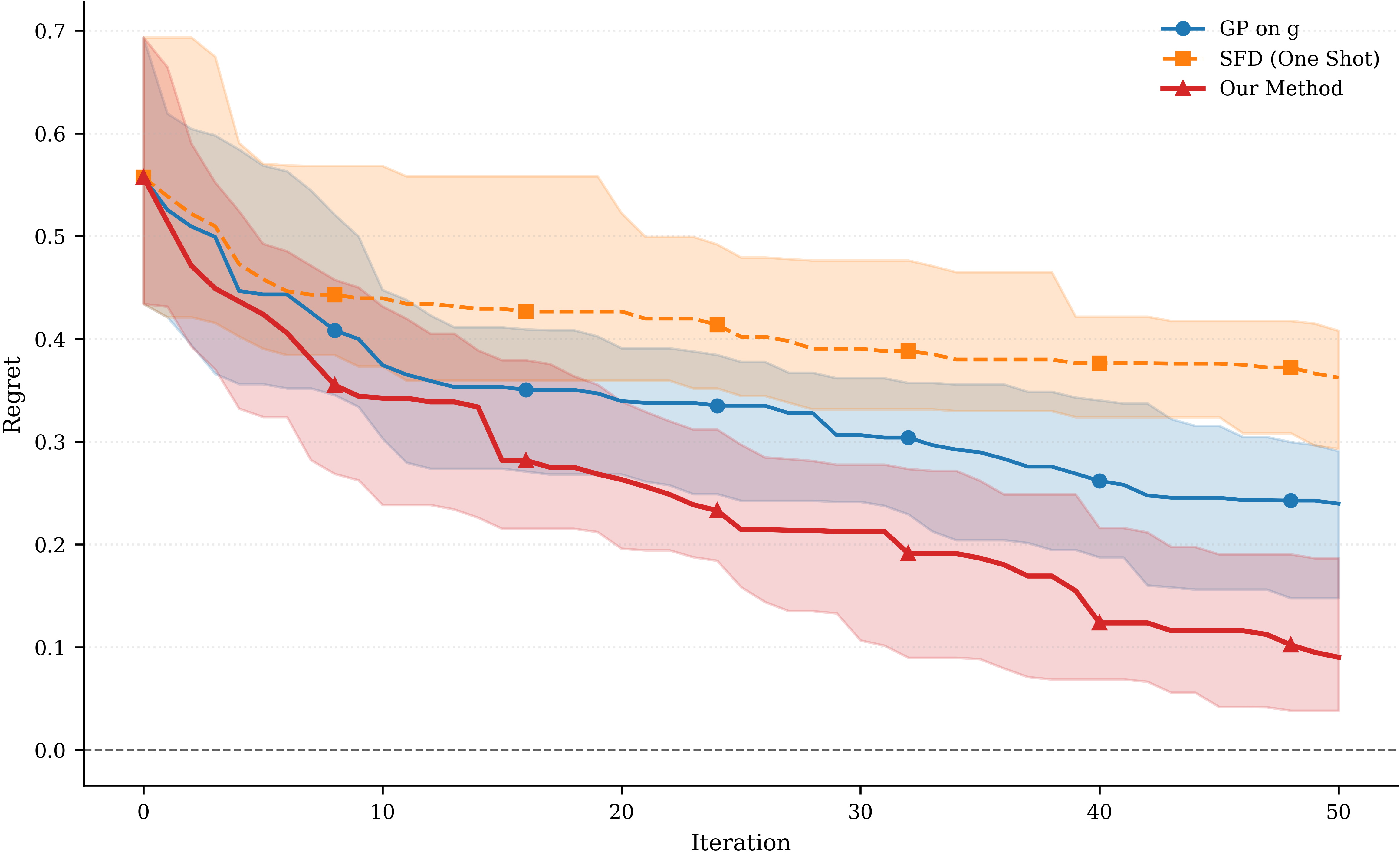}
  \caption{Regret trajectories ($m=50$)}
\end{subfigure}\hfill
\begin{subfigure}[b]{.24\textwidth}
  \centering\includegraphics[width=\linewidth]{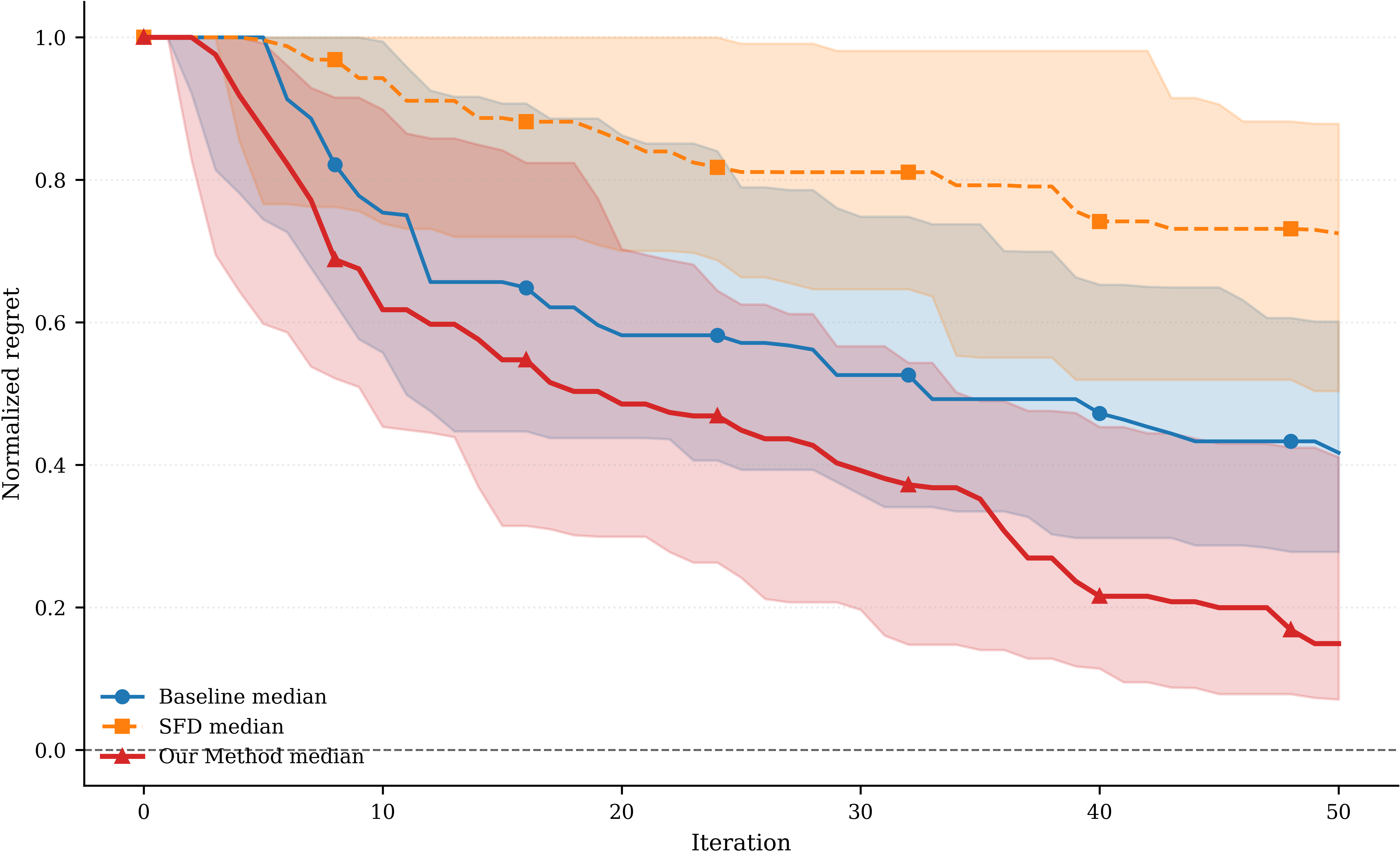}
  \caption{Normalized regret ($m=50$)}
\end{subfigure}\hfill
\begin{subfigure}[b]{.24\textwidth}
  \centering\includegraphics[width=\linewidth]{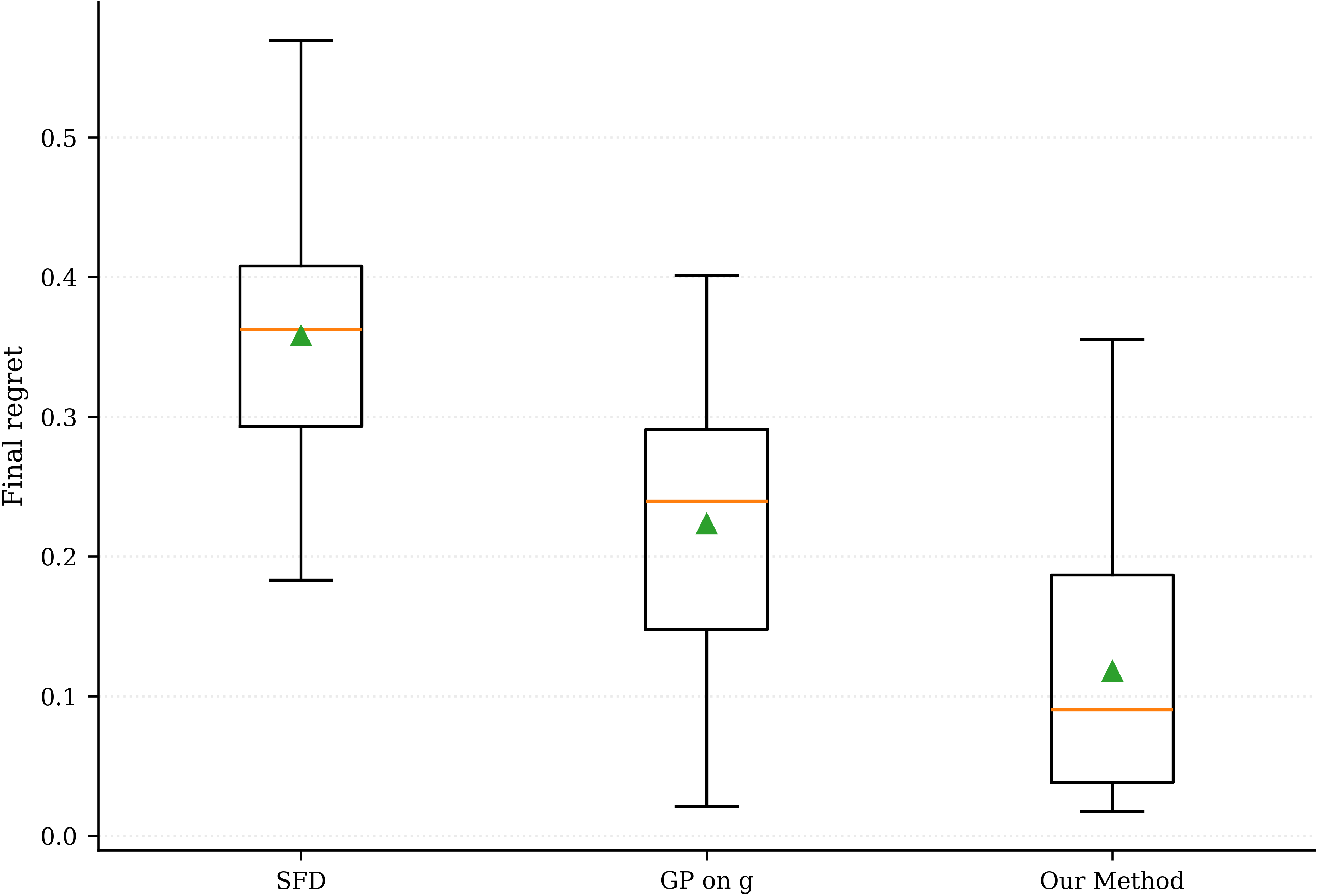}
  \caption{Final regret ($m=50$)}
\end{subfigure}\hfill
\begin{subfigure}[b]{.24\textwidth}
  \centering\includegraphics[width=\linewidth]{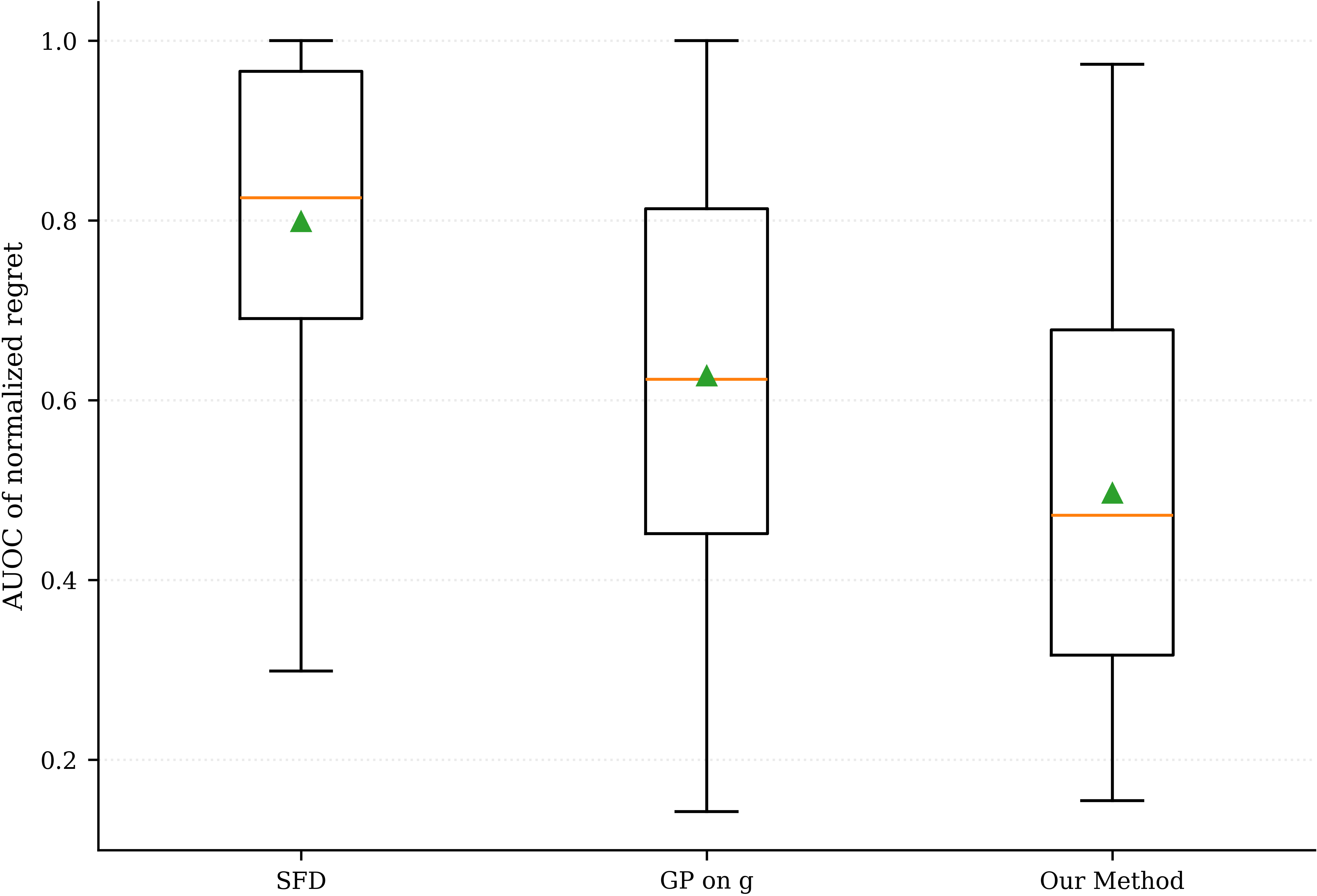}
  \caption{AUOC ($m=50$)}
\end{subfigure}
\caption{Electromagnetic scattering results with budget $m=50$. Our method achieves smaller and more stable regret trajectories, lower AUOC, and significantly better time-to-threshold performance compared with both baselines.}
\label{fig:em_results_m50}
\end{figure*}

\begin{figure*}[t]
\centering
\begin{subfigure}[b]{.24\textwidth}
  \centering\includegraphics[width=\linewidth]{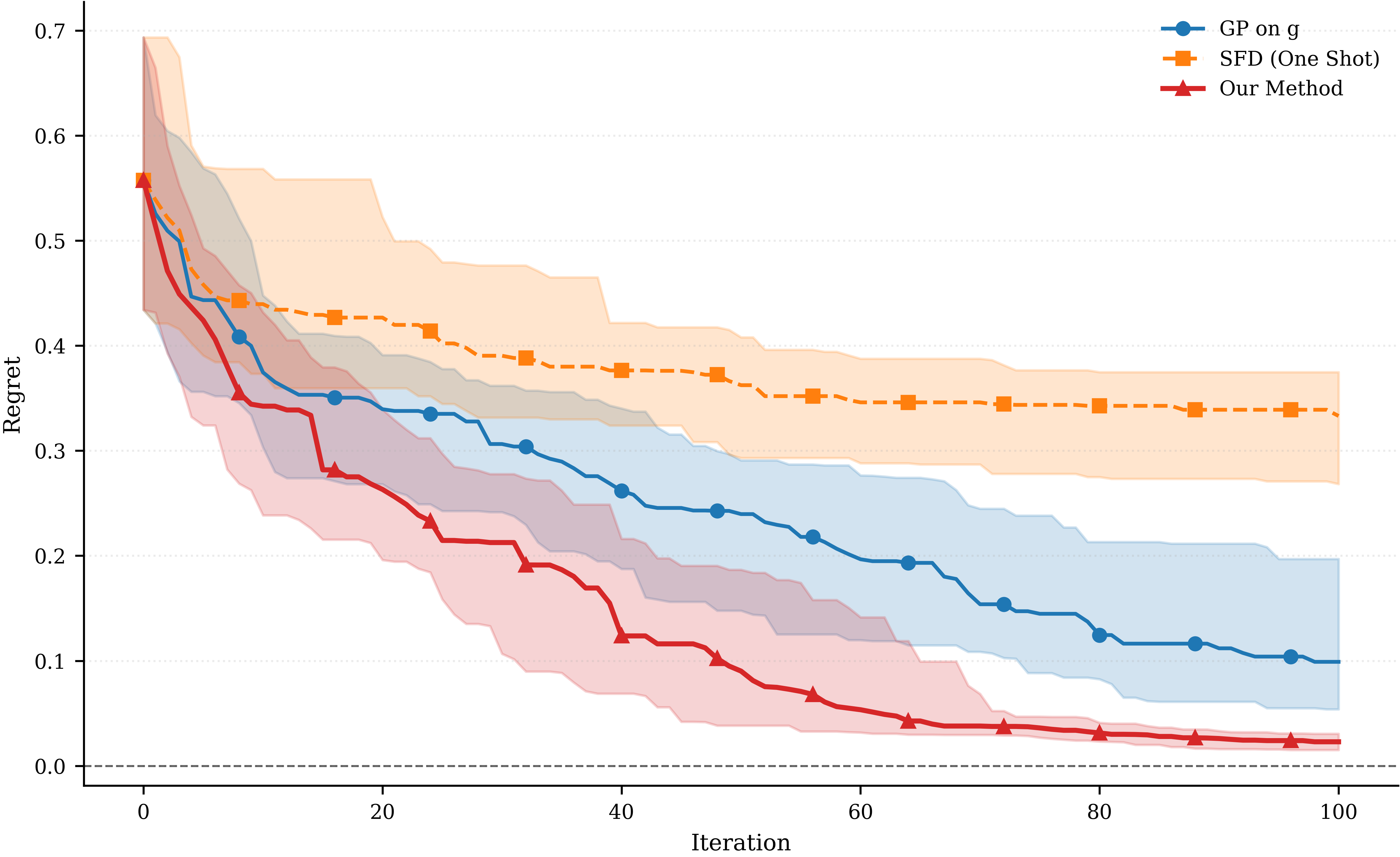}
  \caption{Regret trajectories ($m=100$)}
\end{subfigure}\hfill
\begin{subfigure}[b]{.24\textwidth}
  \centering\includegraphics[width=\linewidth]{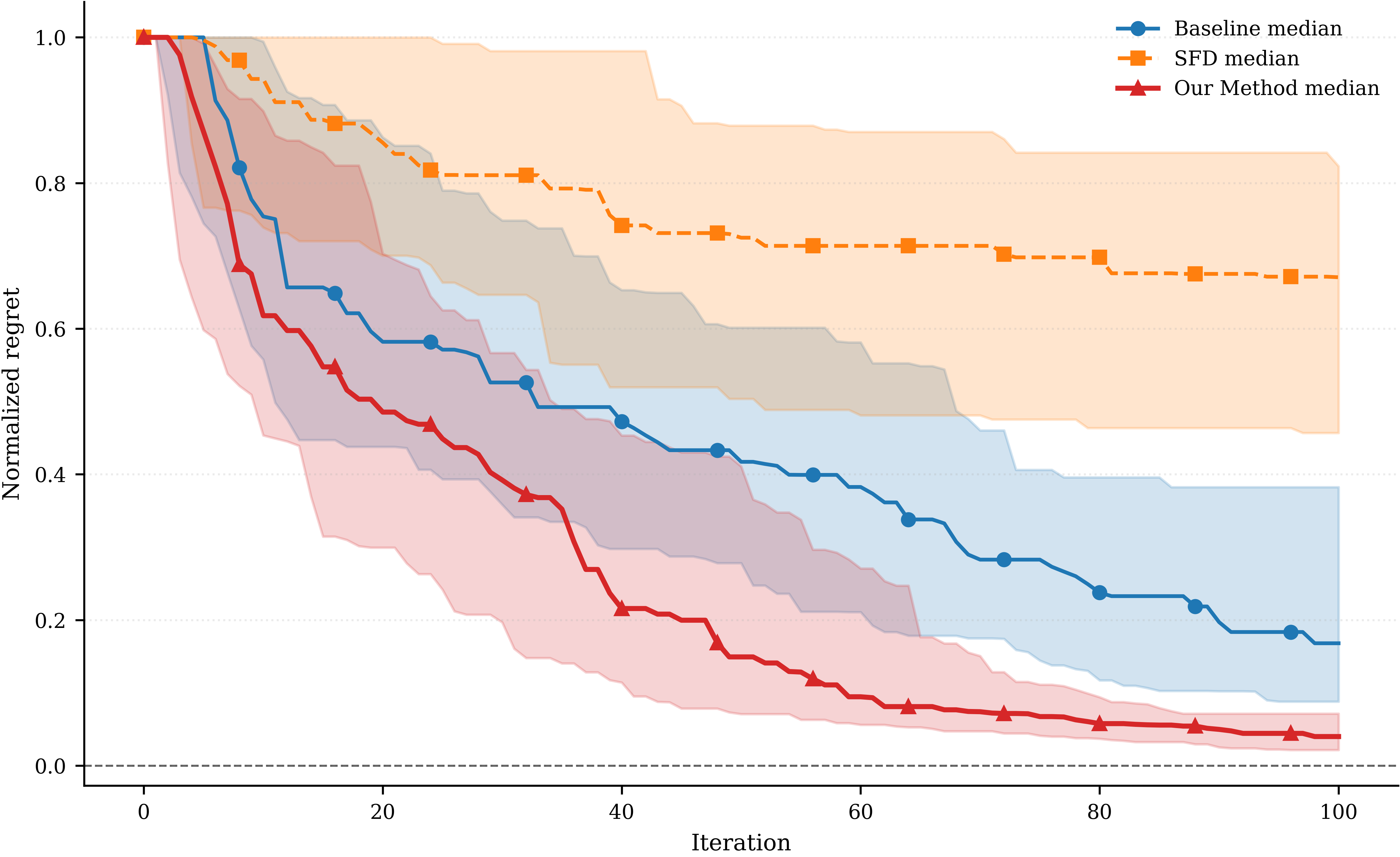}
  \caption{Normalized regret ($m=100$)}
\end{subfigure}\hfill
\begin{subfigure}[b]{.24\textwidth}
  \centering\includegraphics[width=\linewidth]{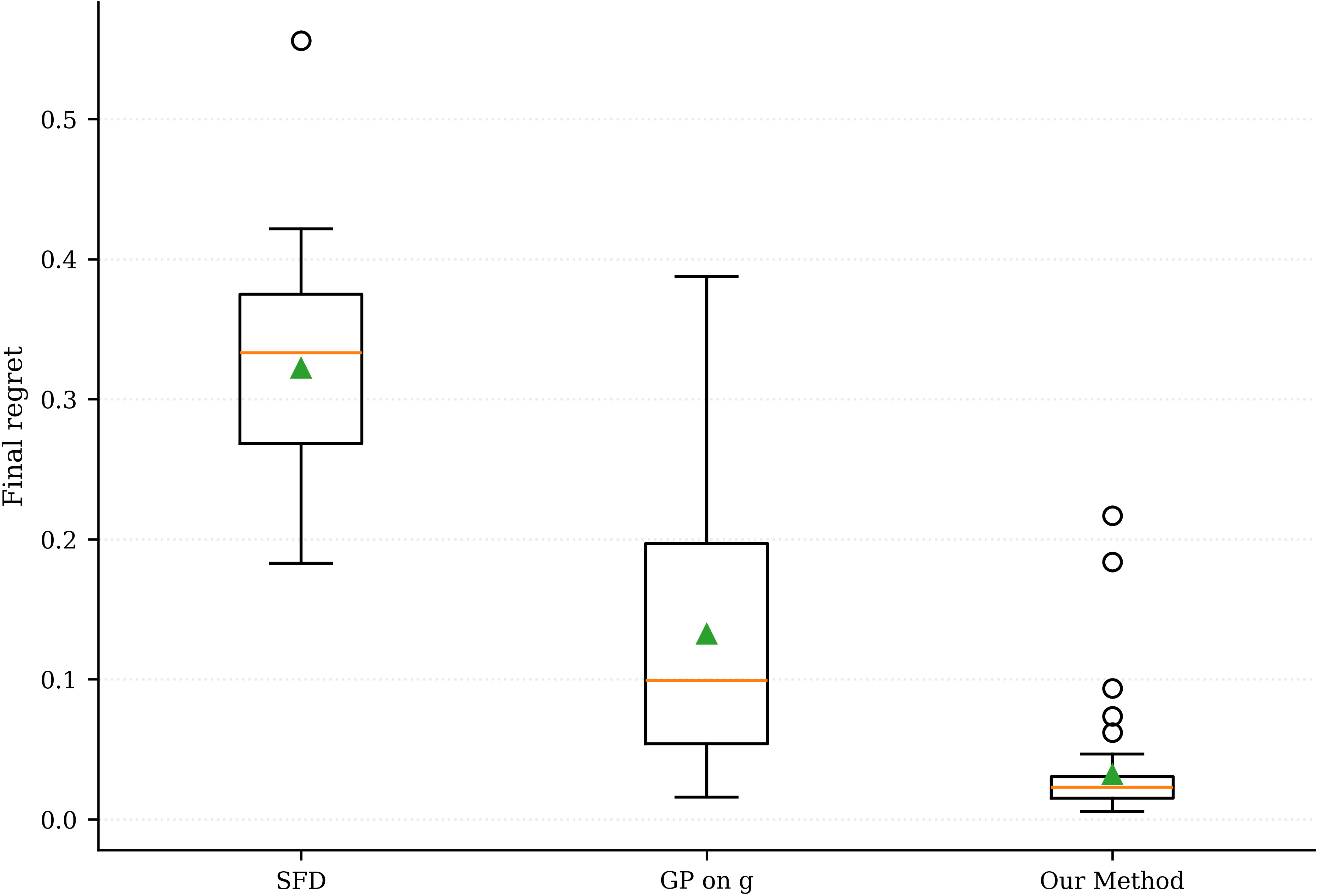}
  \caption{Final regret ($m=100$)}
\end{subfigure}\hfill
\begin{subfigure}[b]{.24\textwidth}
  \centering\includegraphics[width=\linewidth]{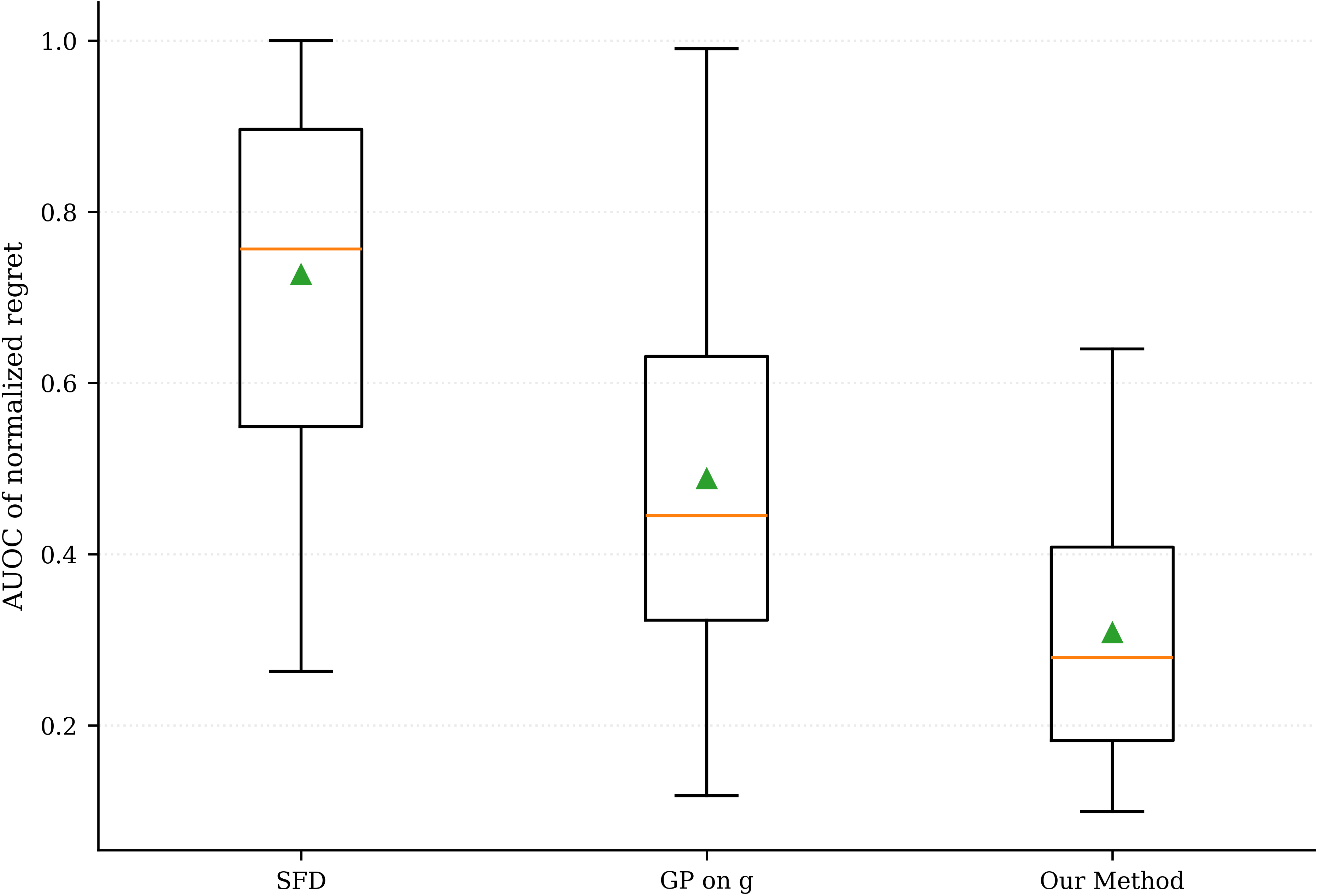}
  \caption{AUOC ($m=100$)}
\end{subfigure}
\caption{Electromagnetic scattering results with budget $m=100$. Larger budget accentuates the advantage of our method, which consistently dominates both baselines across all metrics.}
\label{fig:em_results_m100}
\end{figure*}

The optimization results for the electromagnetic scattering case study are summarized in Figures~\ref{fig:em_results_m50} and \ref{fig:em_results_m100}, with corresponding TT metrics reported in Tables~\ref{tab:em_tte_m50} and \ref{tab:em_tte_m100}. We first consider the low-budget setting with $m=50$ evaluations after seeding. Figure~\ref{fig:em_results_m50} (a) shows the regret trajectories across methods. Our approach rapidly drives regret toward zero, reaching very small values well within the budget, whereas the GP on $g$ baseline converges more slowly and the space-filling design (SFD) stagnates at substantially higher levels. Normalized regret curves in Figure~\ref{fig:em_results_m50} (b) reinforce this observation: our method exhibits a sharp and stable descent with minimal interquartile spread, in contrast to the wider bands of the two baselines. Distributional summaries are shown in panels Figures~\ref{fig:em_results_m50} (c) and \ref{fig:em_results_m50} (d): the boxplots indicate that our method achieves significantly smaller final regret and AUOC values, with both medians and interquartile ranges markedly below those of GP on $g$ and SFD. Specifically, the median final regret is $0.0902$ for our method compared with $0.2396$ for GP on $g$ and $0.3624$ for SFD, while AUOC values are $0.4720$, $0.6233$, and $0.8253$ respectively. The TT analysis in Table~\ref{tab:em_tte_m50} shows that at $\varepsilon=0.10$, our method succeeds in $42\%$ of runs with a median crossing at iteration $40$, while GP on $g$ manages only $8\%$ at iteration $41$ and SFD fails entirely. Even under the stricter $\varepsilon=0.05$, our method still crosses in $14\%$ of runs, whereas the two baselines remain almost uniformly unsuccessful.

\begin{table}[t]
\centering
\caption{TT results for electromagnetic scattering with $m=50$. Each entry shows fraction of successful runs and median iteration. “--” indicates that the threshold was not reached within the budget.}
\label{tab:em_tte_m50}
\begin{tabular}{lccc}
\toprule
$\varepsilon$ & SFD & GP on $g$ & MM-FBO \\
\midrule
0.10 & 0.00 (--) & 0.08 (41) & 0.42 (40) \\
0.05 & 0.00 (--) & 0.02 (42) & 0.14 (36) \\
\bottomrule
\end{tabular}
\end{table}

The high-budget setting with $m=100$ evaluations paints an even clearer picture (Figure~\ref{fig:em_results_m100}). Regret trajectories in Figure~\ref{fig:em_results_m100} (a) show that our method not only descends more steeply but also stabilizes at values close to zero, whereas GP on $g$ continues to oscillate and the SFD regret trajectory remains high. Normalized regret in Figure~\ref{fig:em_results_m100} (b) confirms this stability and shows that our method maintains a consistently smaller band of variation across replications. The boxplots in Figures~\ref{fig:em_results_m100} (c) and \ref{fig:em_results_m100} (d) demonstrate that both final regret and AUOC values are dramatically reduced, with much tighter interquartile ranges than in the low-budget case. Quantitatively, the median final regret is only $0.0229$ for our method, compared with $0.0990$ for GP on $g$ and $0.3331$ for SFD, while AUOC values fall to $0.2793$, $0.4451$, and $0.7567$, respectively. The TT results in Table~\ref{tab:em_tte_m100} now show $88\%$ of runs successfully crossing $\varepsilon=0.10$ with a median of just $55$ evaluations in our approach, compared with $32\%$ at iteration $80$ for GP on $g$ and no success for SFD. Even under $\varepsilon=0.05$, our method succeeds in $58\%$ of runs, far above the $12\%$ achieved by GP on $g$.

\begin{table}[t]
\centering
\caption{TT results for electromagnetic scattering with $m=100$. Each entry shows fraction of successful runs and median iteration. “--” indicates that the threshold was not reached within the budget.}
\label{tab:em_tte_m100}
\begin{tabular}{lccc}
\toprule
$\varepsilon$ & SFD & GP on $g$ & MM-FBO \\
\midrule
0.10 & 0.00 (--) & 0.32 (80) & 0.88 (55) \\
0.05 & 0.00 (--) & 0.12 (68) & 0.58 (72) \\
\bottomrule
\end{tabular}
\end{table}

Taken together, these results show that the proposed acquisition function achieves consistent and substantial improvements over scalar-objective baselines. The improvements manifest not only in the final regret but also in the entire optimization trajectory, as measured by AUOC and TT. The gains are particularly pronounced under tighter budgets, underscoring the value of balancing exploration and exploitation across the functional domain.

\paragraph{Case Study 2: Vapor-Phase Infiltration in Polymer Films.}
Vapor-phase infiltration (VPI) is a hybrid materials processing technique in which organic polymers are exposed to vapor-phase metal–organic precursors that sorb, diffuse, and eventually react within the polymer matrix, producing durable organic–inorganic composites with enhanced mechanical, optical, and thermal properties~\citep{Huang2021Bayesian,ren2021reaction}. Unlike traditional atomic layer deposition (ALD), which coats surfaces conformally, VPI achieves subsurface infiltration by coupling precursor transport with polymer–precursor reactions. This process has been applied to membranes, sensors, hybrid optics, and structural reinforcements.

From a modeling perspective, simple Fickian diffusion is inadequate to describe VPI since the infiltrating precursor both diffuses and reacts irreversibly with polymer functional groups. A reaction–diffusion framework captures these coupled effects~\citep{ren2021reaction}. Let $C_{\text{free}}(x,t)$ denote the local concentration of free precursor, $C_{\text{polymer}}(x,t)$ the concentration of accessible reactive groups, and $C_{\text{product}}(x,t)$ the concentration of immobilized product. The governing equations are
\begin{align}
\frac{\partial C_{\text{free}}}{\partial t} &= D(C_{\text{product}})\frac{\partial^2 C_{\text{free}}}{\partial x^2} - k\,C_{\text{free}}\,C_{\text{polymer}}, \\
\frac{\partial C_{\text{product}}}{\partial t} &= k\,C_{\text{free}}\,C_{\text{polymer}}, \\
D(C_{\text{product}}) &= D_0 \exp\!\left(-K' C_{\text{product}}\right), \\
\frac{\partial C_{\text{polymer}}}{\partial t} &= -k\,C_{\text{free}}\,C_{\text{polymer}},
\end{align}
with $D_0$ being the fresh diffusivity, $k$ the second-order rate constant, and $K'$ a hindering constant that encodes how immobilized product reduces free volume and transport. Nondimensionalization yields a set of key parameters: the Damköhler number $Da = k C_{\text{polymer}}^0 L^2/D_0$, the ratio $C_s/C_{\text{polymer}}^0$ of surface concentration to initial reactive groups, and the hindering number $K'C_{\text{polymer}}^0$, which jointly dictate whether infiltration is diffusion-limited, reaction-limited, or hindered.

We adopt this VPI case study following \cite{Huang2021Bayesian} as a benchmark for functional BO under reaction diffusion uncertainty. In this setting the functional index is time \(t\) with domain \(\mathcal T\), and the objective is
\[
g(x) \;=\; \max_{t \in \mathcal T}\,\big(H(t;x) - f^\ast(t)\big)^{2}.
\]
Here \(H(t;x)\) denotes the predicted infiltration uptake over time for design \(x\), and \(f^{\ast}(t)\) is the target infiltration profile. We set \(f^{\ast}(t)\) to the trajectory generated at a designated reference design \(x^{\ast}\), which represents a physically meaningful configuration of the process and yields an attainable performance level.

\begin{figure*}[t]
\centering
\begin{subfigure}[b]{.24\textwidth}
  \centering\includegraphics[width=\linewidth]{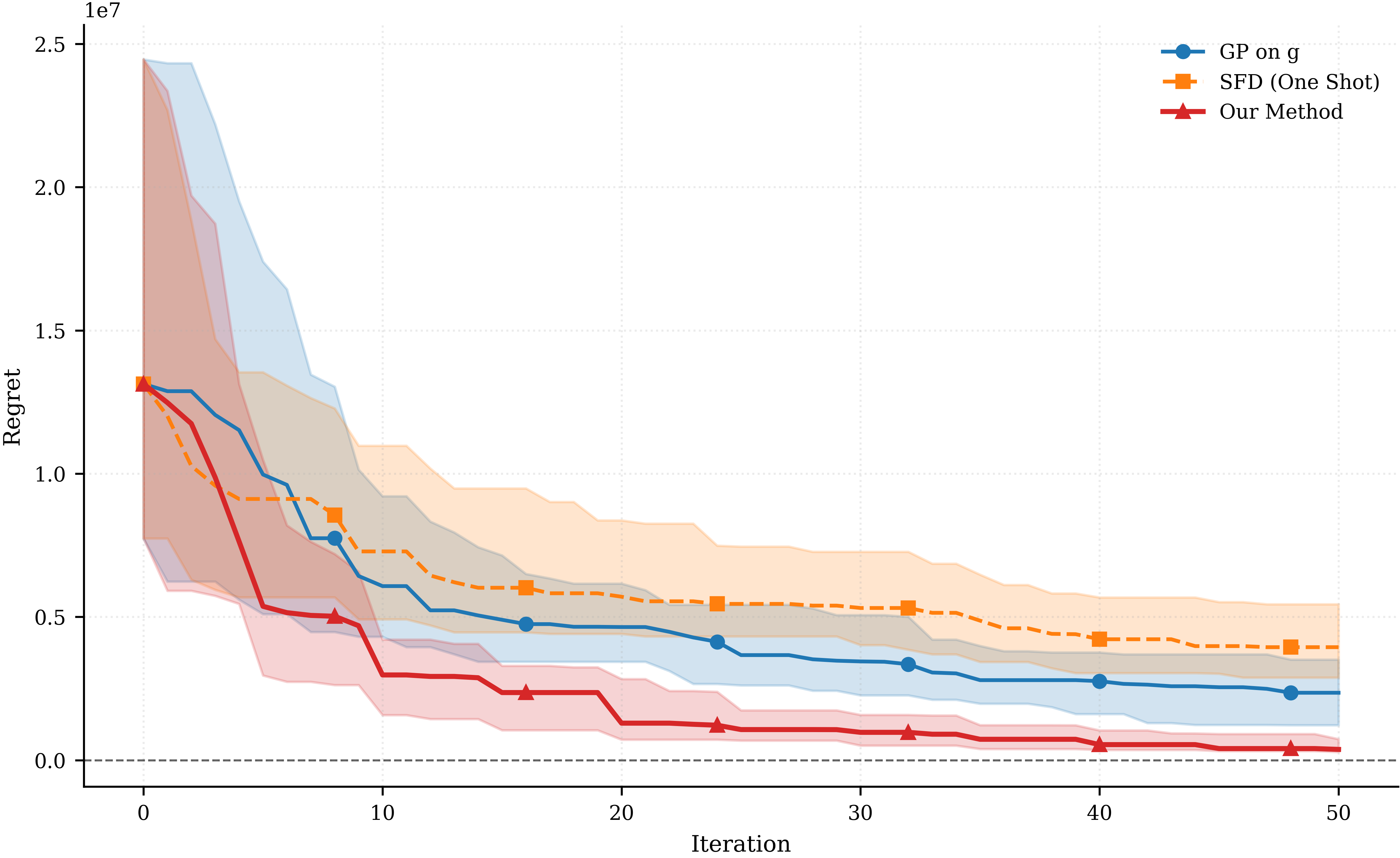}
  \caption{Regret trajectories}
\end{subfigure}\hfill
\begin{subfigure}[b]{.24\textwidth}
  \centering\includegraphics[width=\linewidth]{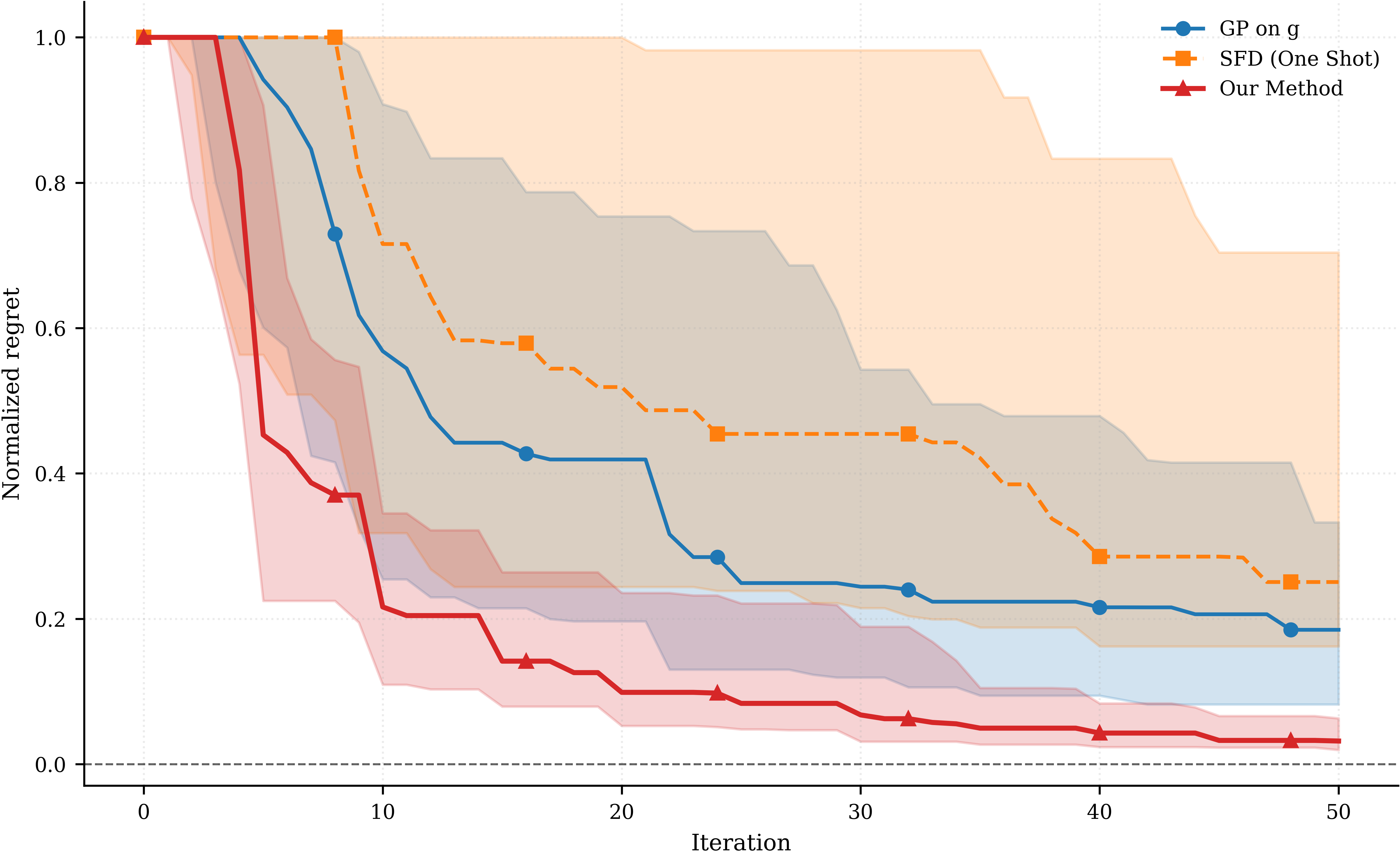}
  \caption{Normalized regret}
\end{subfigure}\hfill
\begin{subfigure}[b]{.24\textwidth}
  \centering\includegraphics[width=\linewidth]{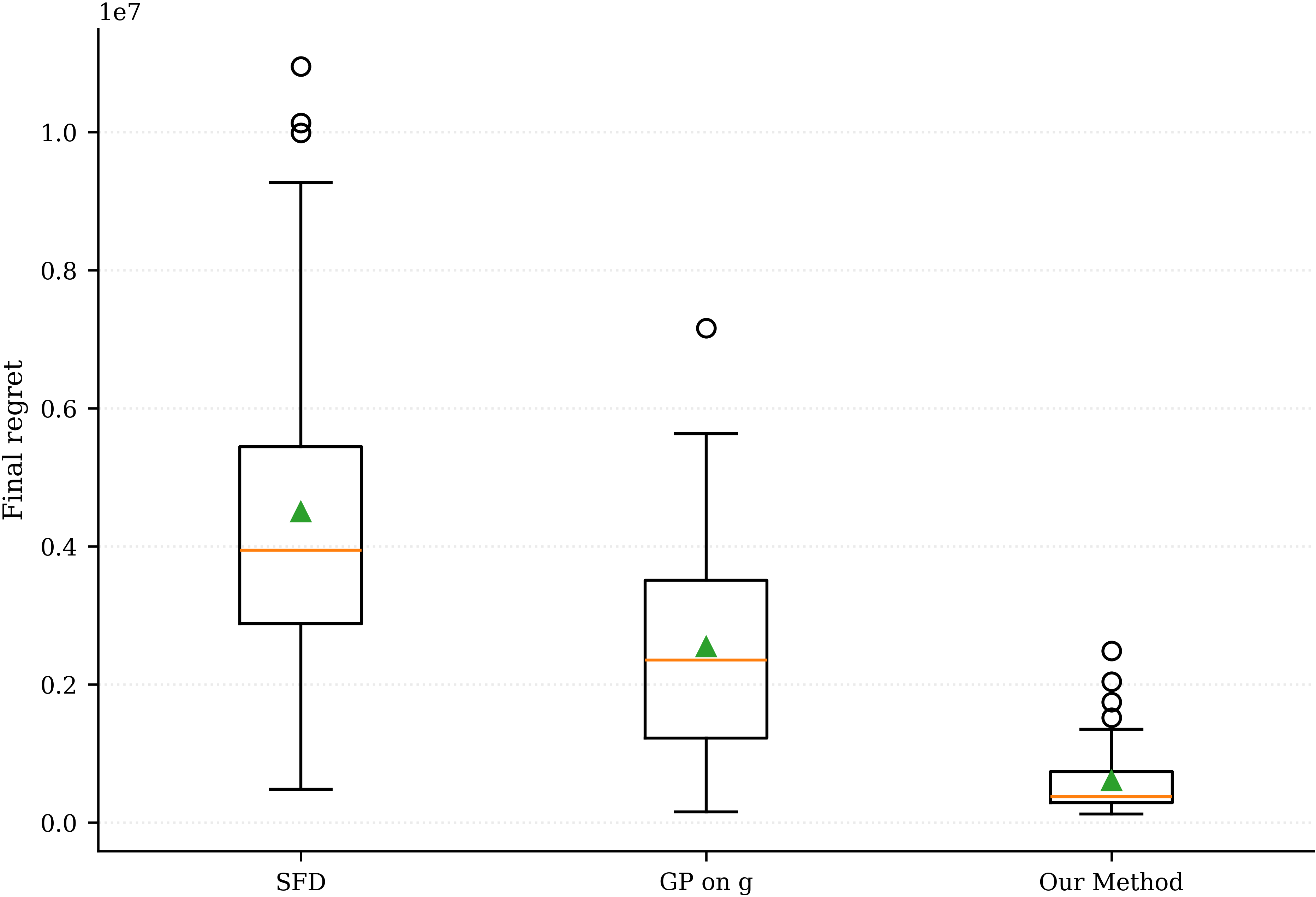}
  \caption{Final regret}
\end{subfigure}\hfill
\begin{subfigure}[b]{.24\textwidth}
  \centering\includegraphics[width=\linewidth]{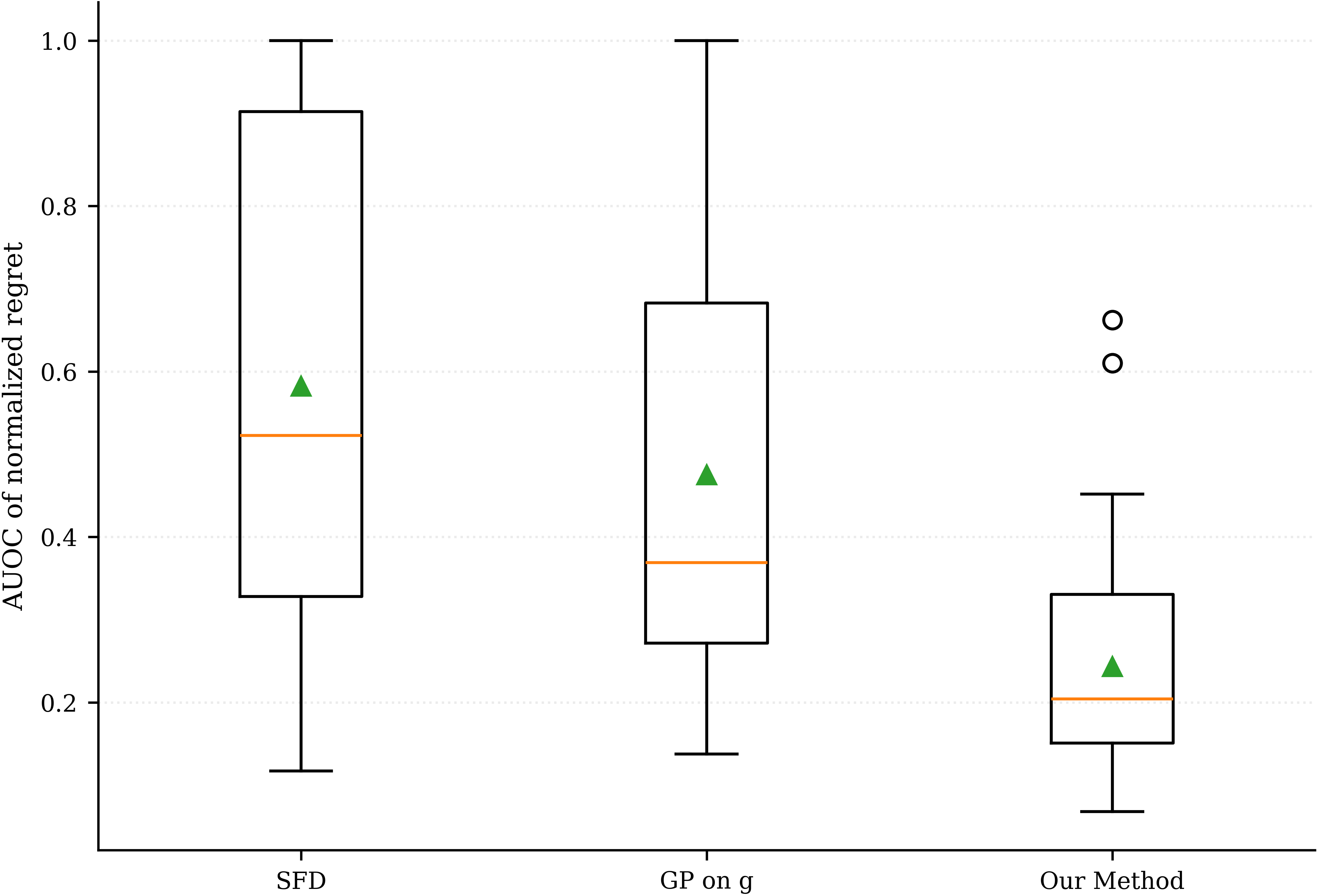}
  \caption{AUOC}
\end{subfigure}
\caption{VPI results. Our method achieves markedly smaller final regret and AUOC, with faster TT compared to both SFD and GP on $g$.}
\label{fig:vpi_results}
\end{figure*}

The optimization outcomes are summarized in Figure~\ref{fig:vpi_results}. Regret trajectories (Figure~\ref{fig:vpi_results} (a)) show that our method rapidly reduces the objective, while the baselines converge more slowly and with higher variability. Normalized regret (Figure~\ref{fig:vpi_results} (b)) confirms stability: the spread of our method is narrower and consistently below alternatives. Final performance comparisons in Figures~\ref{fig:vpi_results} (c) and \ref{fig:vpi_results} (d) highlight substantial improvements: median final regret drops to $3.74\times 10^{5}$ for our method versus $2.35\times 10^{6}$ (GP on $g$) and $3.94\times 10^{6}$ (SFD), while AUOC values are reduced to $0.2042$ compared with $0.3692$ and $0.5227$ respectively. These gains indicate that our method not only finds better solutions but does so more efficiently across the entire budget.

Time-to-threshold (TT) metrics in Table~\ref{tab:vpi_tte} further underscore this advantage. At $\varepsilon=0.10$, our method succeeds in $85\%$ of runs within a median of $20$ iterations, compared to $35\%$ at iteration $25$ for GP on $g$ and only $10\%$ at iteration $14$ for SFD. Under the stricter $\varepsilon=0.05$, our method still achieves $72\%$ success at iteration $33$, well above the $17\%$ for GP on $g$ and the negligible $3\%$ for SFD. Thus, our proposed approach consistently accelerates convergence toward high-quality infiltration designs.

\begin{table}[t]
\centering
\caption{Time to threshold (TT) results for VPI. Each entry shows fraction of successful runs and median iteration.}
\label{tab:vpi_tte}
\begin{tabular}{lccc}
\toprule
$\varepsilon$ & SFD & GP on $g$ & MM-FBO \\
\midrule
0.10 & 0.10 (14) & 0.35 (25) & 0.85 (20) \\
0.05 & 0.03 (17) & 0.17 (35) & 0.72 (33) \\
\bottomrule
\end{tabular}
\end{table}

Taken together, the VPI case study demonstrates that balancing exploration over functional responses yields dramatic performance gains. Unlike the baselines, which struggle under the coupled reaction–diffusion dynamics, our method leverages the full functional structure to achieve both lower regret and faster attainment of practically relevant error thresholds.

\section{Discussion and Conclusion}

This work introduced a framework for BO tailored to function-valued responses under a min--max criterion. By combining a Mercer based functional representation with Gaussian process surrogates for the coefficient processes, we derived an acquisition function that explicitly balances worst case exploitation with integrated exploration. Our experiments across synthetic oracles, physics-inspired simulations, and case studies in metaphotonics and VPI demonstrate that our proposed method consistently achieves lower regret, faster convergence, and improved stability compared with established baselines.

From a theoretical perspective, the results highlight the importance of moving beyond integrated deviation formulations toward explicit worst-case modeling. The min--max objective, though more challenging analytically, provides stronger guarantees of uniform fidelity across the functional domain. Our acquisition design leverages closed form squared deviation statistics to achieve computational tractability while ensuring asymptotic convergence of the acquisition to the target objective. These contributions extend the BO theory into settings where functional responses and robustness requirements are central.

Practically, the methodology offers clear benefits for engineering and scientific applications. In metaphotonic design, it identifies geometries that maintain high scattering response across the entire wavelength range rather than only in an average sense. This directly addresses the challenge of designing multifunctional and broadband nanophotonics devices, where hybrid design frameworks are already employing BO to solve worst-case objectives (e.g., maximizing minimum efficiency) in compressed latent spaces. Our MM-FBO framework provides the formal, efficient acquisition strategy needed for exactly these types of problems, moving beyond standard Expected Improvement used on scalarized objectives. In VPI, it accelerates the discovery of infiltration profiles that are robust to coupled reaction--diffusion dynamics. The approach is computationally efficient, relying on vectorized moment calculations and adaptive exploration strategies that avoid costly Monte Carlo approximations. At the same time, the framework is flexible and can be applied to a broad range of functional optimization problems beyond the case studies considered here.

The work also has limitations. The assumption of independence among the coefficient processes, while justified empirically in our data, may not hold in all settings. Extending the method to correlated scores through multi-output Gaussian processes remains an important direction, albeit at higher computational costs. Similarly, the scalability of the approach with respect to the number of retained components $M$ warrants further study, since squared deviation statistics grow quadratically with $M$. Finally, while our acquisition adapts dynamically, additional research is needed to better understand exploration--exploitation trade-offs in high-dimensional functional spaces.

In summary, this work advances BO by directly addressing functional responses under worst-case criteria. Our method unites functional representation, probabilistic modeling, and robust acquisition into a tractable framework that is both theoretically grounded and practically effective. We anticipate that these ideas will stimulate further research at the interface of functional data analysis, robust optimization, and machine learning for scientific discovery.

\bibliography{refrences}

\newpage

\appendix

\section{}
\label{app:coeff-processes}

\textbf{Proposition~\ref{prop:coeff-processes}} (Coefficient Processes of the
KL Expansion). \textit{Under the separable 
covariance~\eqref{eq:separable-kernel} with Mercer 
expansion~\eqref{eq:mercer-expansion}, the coefficient processes 
$\{\alpha_m(x)\}_{m=1}^\infty$ defined by the 
projection~\eqref{eq:coeff-projection} are mutually independent zero-mean 
Gaussian processes over $\mathcal{X}$ with cross-covariance}
\[
\mathrm{Cov}\!\left(\alpha_m(x),\,\alpha_{m'}(x')\right)
\;=\;
\gamma_m\, k_{x}(x,x')\,\delta_{mm'}.
\]

In this appendix, we prove Proposition~\ref{prop:coeff-processes}.

\begin{proof}
\emph{Step 1: Gaussianity.}
By the projection definition~\eqref{eq:coeff-projection},
\[
\alpha_m(x) 
\;=\; 
\int_{\Lambda} f(x,\lambda)\,\phi_m(\lambda)\,d\lambda,
\]
which is a continuous linear functional of the Gaussian process $f$. Linear 
functionals of Gaussian processes are Gaussian, so $\alpha_m(x)$ is a Gaussian 
random variable for each fixed $x\in\mathcal{X}$. More generally, any finite 
linear combination $\sum_{i,m} c_{i,m}\alpha_m(x_i)$ is a linear functional 
of $f$ and is therefore Gaussian, which establishes that 
$\{\alpha_m(x)\}_{m\ge 1}$ is jointly Gaussian over $\mathcal{X}$.

\emph{Step 2: Zero mean.}
The zero-mean assumption on $f$ together with linearity of expectation gives
\[
\mathbb{E}\bigl[\alpha_m(x)\bigr] 
\;=\; 
\int_{\Lambda} \mathbb{E}\bigl[f(x,\lambda)\bigr]\,\phi_m(\lambda)\,d\lambda 
\;=\; 
0,
\]
where the interchange of expectation and integration is justified by Fubini's 
theorem (finite second moments and compact $\Lambda$).

\emph{Step 3: Cross-covariance.}
For any $x,x'\in\mathcal{X}$ and $m,m'\ge 1$, applying 
definition~\eqref{eq:coeff-projection},
\begin{align*}
\mathrm{Cov}\!\left(\alpha_m(x),\,\alpha_{m'}(x')\right)
&\;=\;
\mathbb{E}\!\left[
\int_{\Lambda} f(x,\lambda)\,\phi_m(\lambda)\,d\lambda
\int_{\Lambda} f(x',\lambda')\,\phi_{m'}(\lambda')\,d\lambda'
\right] \\
&\;=\;
\int_{\Lambda}\int_{\Lambda}
\mathbb{E}\bigl[f(x,\lambda)\,f(x',\lambda')\bigr]\,
\phi_m(\lambda)\,\phi_{m'}(\lambda')\,
d\lambda\,d\lambda',
\end{align*}
where the second equality applies Fubini's theorem, valid because the 
integrand has finite expectation over the compact product domain. Since $f$ is 
zero-mean, $\mathbb{E}[f(x,\lambda)f(x',\lambda')]$ equals the joint 
covariance $k((x,\lambda),(x',\lambda'))$. Substituting the separable 
form~\eqref{eq:separable-kernel},
\[
\mathrm{Cov}\!\left(\alpha_m(x),\,\alpha_{m'}(x')\right)
\;=\;
k_x(x,x')
\int_{\Lambda}\int_{\Lambda}
k_\lambda(\lambda,\lambda')\,
\phi_m(\lambda)\,\phi_{m'}(\lambda')\,
d\lambda\,d\lambda'.
\]

\emph{Step 4: Apply the eigenfunction property.}
Using the eigenfunction property~\eqref{eq:eigenfunction-property} of $\phi_m$ 
to evaluate the inner integral,
\[
\int_{\Lambda} k_\lambda(\lambda,\lambda')\,\phi_{m'}(\lambda')\,d\lambda' 
\;=\; 
\gamma_{m'}\,\phi_{m'}(\lambda),
\]
the cross-covariance becomes
\[
\mathrm{Cov}\!\left(\alpha_m(x),\,\alpha_{m'}(x')\right)
\;=\;
k_x(x,x')\,\gamma_{m'}
\int_{\Lambda}
\phi_m(\lambda)\,\phi_{m'}(\lambda)\,
d\lambda.
\]

\emph{Step 5: Apply orthonormality.}
By the orthonormality condition~\eqref{eq:orthonormality}, the remaining 
integral equals $\delta_{mm'}$. When $m \neq m'$, the cross-covariance 
vanishes; when $m = m'$, the factor $\gamma_{m'}$ equals $\gamma_m$, giving
\[
\mathrm{Cov}\!\left(\alpha_m(x),\,\alpha_{m'}(x')\right)
\;=\;
\gamma_m\,k_x(x,x')\,\delta_{mm'}.
\]

\emph{Step 6: Independence and GP structure.}
Vanishing cross-covariance combined with joint Gaussianity (Step~1) implies 
mutual independence of the processes $\{\alpha_m\}_{m\ge 1}$. For each fixed 
$m$, the map $x \mapsto \alpha_m(x)$ is a zero-mean Gaussian process over 
$\mathcal{X}$ with covariance kernel $\gamma_m\,k_x(x,x')$; that is,
\[
\alpha_m(x)
\;\sim\;
\mathcal{GP}\!\left(0,\;\gamma_m\,k_x(x,x')\right),
\qquad m = 1, 2, \ldots,
\]
which completes the proof.
\end{proof}

\section{}
\label{app:truncation}
\textbf{Proposition~\ref{prop:truncation}} (Mercer Truncation Error).
\textit{Let $f(x,\lambda)$ be the Gaussian process with separable 
covariance~\eqref{eq:separable-kernel}, admitting the Karhunen--Lo\`eve 
representation~\eqref{eq:kl-expansion}, and let $f^{(M)}(x,\lambda)$ be its 
$M$-term truncation defined in~\eqref{eq:truncated-model}. Then for every 
$x\in\mathcal{X}$,}
\[
\mathbb{E}\!\left[
\int_{\Lambda}
\bigl(f(x,\lambda)-f^{(M)}(x,\lambda)\bigr)^2
\,d\lambda
\right]
\;=\;
k_x(x,x)\sum_{m=M+1}^{\infty}\gamma_m
\;=\;
k_x(x,x)\!\left(\sum_{m=1}^{\infty}\gamma_m\right)(1-r_M).
\]

In this appendix, we prove Proposition~\ref{prop:truncation}.

\begin{proof}
The truncation remainder is
\[
f(x,\lambda) - f^{(M)}(x,\lambda)
=
\sum_{m=M+1}^{\infty} \alpha_m(x)\,\phi_m(\lambda).
\]

\emph{Step 1: Apply Parseval's identity.}
Since $\{\phi_m\}_{m=1}^\infty$ is a complete orthonormal system in
$L^2(\Lambda)$, Parseval's identity applies on the event that
$\sum_{m=M+1}^{\infty}\alpha_m(x)^2 < \infty$, which holds almost surely
by the following argument. From Section~\ref{subsec:functional-rep},
$\mathbb{E}[\alpha_m(x)^2] = \gamma_m k_x(x,x)$, so
\[
\sum_{m=1}^\infty \mathbb{E}[\alpha_m(x)^2]
= k_x(x,x)\sum_{m=1}^\infty \gamma_m
= k_x(x,x)\int_\Lambda k_\lambda(\lambda,\lambda)\,d\lambda
< \infty,
\]
where finiteness follows from the finiteness of the trace of the integral
operator induced by $k_\lambda$. Hence $\sum_{m=M+1}^\infty \alpha_m(x)^2
< \infty$ almost surely, and Parseval's identity gives
\[
\int_{\Lambda}
\left(
\sum_{m=M+1}^{\infty} \alpha_m(x)\,\phi_m(\lambda)
\right)^2
\,d\lambda
=
\sum_{m=M+1}^{\infty} \alpha_m(x)^2.
\]

\emph{Step 2: Exchange expectation and summation.}
Since $\alpha_m(x)^2 \ge 0$ for all $m$, we may apply the monotone
convergence theorem to exchange expectation and sum,
\[
\mathbb{E}\!\left[
\sum_{m=M+1}^{\infty} \alpha_m(x)^2
\right]
=
\sum_{m=M+1}^{\infty} \mathbb{E}\!\left[\alpha_m(x)^2\right].
\]
From Section~\ref{subsec:functional-rep}, the prior of each coefficient
process satisfies $\alpha_m(x) \sim \mathcal{GP}(0,\gamma_m k_x(x,x'))$,
so $\mathbb{E}[\alpha_m(x)^2] = \gamma_m k_x(x,x)$ for every
$x\in\mathcal{X}$. Substituting,
\[
\mathbb{E}\!\left[
\int_{\Lambda}
\bigl(f(x,\lambda) - f^{(M)}(x,\lambda)\bigr)^2
\,d\lambda
\right]
=
\sum_{m=M+1}^{\infty} \gamma_m\, k_x(x,x)
=
k_x(x,x)\sum_{m=M+1}^{\infty} \gamma_m.
\]
Finally, writing $\sum_{m=M+1}^\infty \gamma_m =
\bigl(\sum_{m=1}^\infty \gamma_m\bigr)(1 - r_M)$
by definition of $r_M$ gives the second equality in the statement.
\end{proof}

\section{}
\label{app:posterior}

\textbf{Proposition~\ref{prop:posterior}} (Posterior Reconstruction of the
Functional Response).
\textit{Let $\mu_m(x)$ and $\sigma_m^2(x)$ denote the posterior mean and 
variance of the $m$-th coefficient process given by~\eqref{eq:coeff-post-mean} 
and~\eqref{eq:coeff-post-var}. Under the truncated separable GP 
model~\eqref{eq:truncated-model}, the marginal posterior of the functional 
response at any $(x,\lambda)\in\mathcal{X}\times\Lambda$ is Gaussian with mean 
and variance given by~\eqref{eq:post-f}. Furthermore, the posterior covariance 
between two functional locations $\lambda$ and $\lambda'$ is given 
by~\eqref{eq:post-cov-f}.}

In this appendix, we prove Proposition~\ref{prop:posterior}.

\begin{proof}
Under the $M$-term truncation from Section~\ref{subsec:functional-rep},
the functional surrogate at any $(x,\lambda)$ is the finite linear
combination
\[
f^{(M)}(x,\lambda)
=
\sum_{m=1}^{M} \alpha_m(x)\,\phi_m(\lambda).
\]

\emph{Step 1: Posterior independence of the coefficient processes.}
Prior to observing data, the coefficient processes
$\{\alpha_m\}_{m=1}^M$ are mutually independent Gaussian processes, as
established in Section~\ref{subsec:functional-rep}. Under the observation model~\eqref{eq:obs-model} of 
Section~\ref{subsec:surrogate}, the noisy coefficient vector 
$\boldsymbol{\alpha}_m$ depends only on the $m$-th coefficient process 
$\alpha_m$ (the observation noise is independent across modes by 
assumption), so the likelihood factorizes across modes:
\[
p(\boldsymbol{\alpha}_1,\dots,\boldsymbol{\alpha}_M \mid \alpha_1,\dots,\alpha_M)
\;=\;
\prod_{m=1}^{M} p(\boldsymbol{\alpha}_m \mid \alpha_m).
\]
Combined with the independent prior, this implies that the posterior over
the coefficient processes factorizes across $m$, so the processes
$\{\alpha_m\mid\mathcal{D}\}_{m=1}^M$ remain mutually independent given
$\mathcal{D}$. Each marginal posterior $\alpha_m(x)\mid\mathcal{D}$ is
Gaussian with mean $\mu_m(x)$ and variance $\sigma_m^2(x)$ as given in
Section~\ref{subsec:surrogate}.

\emph{Step 2: Gaussianity and posterior mean of $f^{(M)}(x,\lambda)$.}
Since $f^{(M)}(x,\lambda)$ is a finite linear combination of the
conditionally independent Gaussian variables
$\{\alpha_m(x)\mid\mathcal{D}\}_{m=1}^M$, it is itself Gaussian given
$\mathcal{D}$. Its posterior mean follows by linearity of expectation,
\[
\mathbb{E}\!\left[f^{(M)}(x,\lambda)\mid\mathcal{D}\right]
=
\sum_{m=1}^{M}
\mathbb{E}\!\left[\alpha_m(x)\mid\mathcal{D}\right]\phi_m(\lambda)
=
\sum_{m=1}^{M}\mu_m(x)\,\phi_m(\lambda)
=
\mu_f(x,\lambda).
\]

\emph{Step 3: Posterior covariance and marginal variance.}
For any two functional locations $\lambda,\lambda'\in\Lambda$, the
posterior covariance is
\begin{align*}
\mathrm{Cov}\!\left(f^{(M)}(x,\lambda),\,f^{(M)}(x,\lambda')\mid\mathcal{D}\right)
&=
\mathrm{Cov}\!\left(
\sum_{m=1}^{M}\alpha_m(x)\phi_m(\lambda),\,
\sum_{m'=1}^{M}\alpha_{m'}(x)\phi_{m'}(\lambda')
\;\Bigg|\;\mathcal{D}
\right)\\
&=
\sum_{m=1}^{M}\sum_{m'=1}^{M}
\phi_m(\lambda)\,\phi_{m'}(\lambda')\,
\mathrm{Cov}\!\left(\alpha_m(x),\alpha_{m'}(x)\mid\mathcal{D}\right).
\end{align*}
By Step~1, the coefficient processes are mutually independent given
$\mathcal{D}$, so $\mathrm{Cov}(\alpha_m(x),\alpha_{m'}(x)\mid\mathcal{D})
= \sigma_m^2(x)\,\mathbf{1}_{m=m'}$. All cross terms with $m\neq m'$
therefore vanish, leaving
\[
\mathrm{Cov}\!\left(f^{(M)}(x,\lambda),\,f^{(M)}(x,\lambda')\mid\mathcal{D}\right)
=
\sum_{m=1}^{M}\sigma_m^2(x)\,\phi_m(\lambda)\,\phi_m(\lambda').
\]
Setting $\lambda = \lambda'$ gives the marginal posterior variance
$\sigma_f^2(x,\lambda) = \sum_{m=1}^{M}\sigma_m^2(x)\,\phi_m(\lambda)^2$,
completing the proof.
\end{proof}

\section{}
\label{app:deviation}

\textbf{Proposition~\ref{prop:deviation}} (Distribution and Moments of the
Posterior Squared Deviation).
\textit{Suppose $h(x,\lambda)\sim\mathcal{N}(\mu_h(x,\lambda),\,
\sigma_h^2(x,\lambda))$ for fixed $(x,\lambda)$, with
$\sigma_h^2(x,\lambda)>0$. Then $d(x,\lambda)=h(x,\lambda)^2$ satisfies}
\[
d(x,\lambda)
\;\sim\;
\sigma_h^2(x,\lambda)\,\chi_1^{\prime\,2}\!\left(\delta(x,\lambda)\right),
\qquad
\delta(x,\lambda)
=
\frac{\mu_h^2(x,\lambda)}{\sigma_h^2(x,\lambda)},
\]
\textit{and its first two moments are}
\[
\mu_d(x,\lambda)
=
\mu_h^2(x,\lambda) + \sigma_h^2(x,\lambda),
\qquad
\sigma_d^2(x,\lambda)
=
2\sigma_h^4(x,\lambda) + 4\mu_h^2(x,\lambda)\,\sigma_h^2(x,\lambda).
\]

In this appendix, we prove Proposition~\ref{prop:deviation}. Throughout
the proof we suppress the arguments $(x,\lambda)$ and write $\mu_h$,
$\sigma_h^2$, $\delta$, $\mu_d$, $\sigma_d^2$ for brevity.

\begin{proof}
\emph{Step 1: Distribution of $d$.}
Since $h \sim \mathcal{N}(\mu_h, \sigma_h^2)$ with $\sigma_h^2 > 0$, we
can write $h = \sigma_h Z + \mu_h$ where $Z \sim \mathcal{N}(0,1)$.
Dividing by $\sigma_h$,
\[
\frac{h}{\sigma_h}
=
Z + \frac{\mu_h}{\sigma_h}
\;\sim\;
\mathcal{N}\!\left(\frac{\mu_h}{\sigma_h},\,1\right),
\]
which is a noncentral standard normal with noncentrality parameter
$\mu_h/\sigma_h$. Squaring both sides and multiplying by $\sigma_h^2$,
\[
d = h^2 = \sigma_h^2 \left(\frac{h}{\sigma_h}\right)^2
\;\sim\;
\sigma_h^2\,\chi_1^{\prime\,2}(\delta),
\qquad
\delta = \frac{\mu_h^2}{\sigma_h^2},
\]
where the last step uses the definition that the square of a noncentral
standard normal with noncentrality $\mu_h/\sigma_h$ follows a
$\chi_1^{\prime\,2}(\delta)$ distribution with
$\delta = (\mu_h/\sigma_h)^2 = \mu_h^2/\sigma_h^2$.

\emph{Step 2: Mean of $d$.}
The mean follows directly from the variance--bias decomposition,
\[
\mu_d
=
\mathbb{E}[h^2]
=
\mathrm{Var}(h) + \bigl(\mathbb{E}[h]\bigr)^2
=
\sigma_h^2 + \mu_h^2.
\]

\emph{Step 3: Variance of $d$.}
We compute $\mathrm{Var}(d) = \mathbb{E}[h^4] - (\mathbb{E}[h^2])^2$.
For a Gaussian random variable $h \sim \mathcal{N}(\mu_h, \sigma_h^2)$,
the fourth central moment is $\mathbb{E}[(h-\mu_h)^4] = 3\sigma_h^4$.
Expanding $h^4 = ((h-\mu_h)+\mu_h)^4$ using the binomial theorem and
taking expectations, noting that odd central moments of a Gaussian vanish,
\begin{align*}
\mathbb{E}[h^4]
&=
\mathbb{E}[(h-\mu_h)^4]
+ 4\mu_h\,\mathbb{E}[(h-\mu_h)^3]
+ 6\mu_h^2\,\mathbb{E}[(h-\mu_h)^2]
+ 4\mu_h^3\,\mathbb{E}[h-\mu_h]
+ \mu_h^4 \\
&=
3\sigma_h^4 + 0 + 6\mu_h^2\sigma_h^2 + 0 + \mu_h^4
=
\mu_h^4 + 6\mu_h^2\sigma_h^2 + 3\sigma_h^4.
\end{align*}
Therefore,
\begin{align*}
\sigma_d^2
&=
\mathbb{E}[h^4] - \bigl(\mathbb{E}[h^2]\bigr)^2 \\
&=
\mu_h^4 + 6\mu_h^2\sigma_h^2 + 3\sigma_h^4
-
\bigl(\mu_h^2 + \sigma_h^2\bigr)^2 \\
&=
\mu_h^4 + 6\mu_h^2\sigma_h^2 + 3\sigma_h^4
-
\mu_h^4 - 2\mu_h^2\sigma_h^2 - \sigma_h^4 \\
&=
2\sigma_h^4 + 4\mu_h^2\sigma_h^2,
\end{align*}
which completes the proof.
\end{proof}

\section{}
\label{app:consistency}

\textbf{Theorem~\ref{thm:consistency}} (Consistency of the Acquisition
Function and Its Minimizers).
\textit{Under the assumptions of Section~\ref{sec:formulation}, with 
$\mathcal{X}$ and $\Lambda$ compact and $g(x) = \sup_{\lambda\in\Lambda} 
d(x,\lambda)$ continuous on $\mathcal{X}$, suppose conditions 
\textnormal{(i)}--\textnormal{(iii)} of Theorem~\ref{thm:consistency} hold. 
Then $\sup_{x\in\mathcal{X}}|\alpha_t(x)-g(x)|\to 0$, and any sequence 
$\{x_t\}$ with $\alpha_t(x_t)\le\inf_x\alpha_t(x)+\varepsilon_t$ and 
$\varepsilon_t\to 0$ has every limit point in 
$\arg\min_{x\in\mathcal{X}}g(x)$.}

In this appendix, we prove Theorem~\ref{thm:consistency}. Define the
pointwise surrogate error and integrated uncertainty at iteration $t$ as
\[
\Delta_t(x)
\;=\;
\sup_{\lambda\in\Lambda}
\bigl|\mu_{d,t}(x,\lambda) - d(x,\lambda)\bigr|,
\qquad
U_t(x)
\;=\;
\int_{\Lambda}\sigma_{d,t}(x,\lambda)\,d\lambda.
\]
By conditions~(i) and~(ii) of Theorem~\ref{thm:consistency}, 
$\sup_{x}\Delta_t(x)\to 0$ and $\sup_{x}U_t(x)\to 0$ as $t\to\infty$.

\begin{proof}
\emph{Step 1: Pointwise bound on $|\alpha_t(x) - g(x)|$.}
Fix $x\in\mathcal{X}$. Since $-\kappa_t U_t(x) \le 0$,
\begin{align*}
\alpha_t(x) - g(x)
&\;=\;
\sup_{\lambda}\mu_{d,t}(x,\lambda)
- \kappa_t U_t(x)
- \sup_{\lambda} d(x,\lambda) \\
&\;\le\;
\sup_{\lambda}\mu_{d,t}(x,\lambda)
- \sup_{\lambda} d(x,\lambda) \\
&\;\le\;
\sup_{\lambda}
\bigl(\mu_{d,t}(x,\lambda) - d(x,\lambda)\bigr)
\;\le\;
\Delta_t(x).
\end{align*}
In the other direction,
\begin{align*}
g(x) - \alpha_t(x)
&\;=\;
\sup_{\lambda} d(x,\lambda)
- \sup_{\lambda}\mu_{d,t}(x,\lambda)
+ \kappa_t U_t(x) \\
&\;\le\;
\sup_{\lambda}
\bigl(d(x,\lambda) - \mu_{d,t}(x,\lambda)\bigr)
+ \kappa_t U_t(x)
\;\le\;
\Delta_t(x) + \kappa_t U_t(x).
\end{align*}
Combining the two inequalities,
\[
\bigl|\alpha_t(x) - g(x)\bigr|
\;\le\;
\Delta_t(x) + \kappa_t\, U_t(x).
\]

\emph{Step 2: Uniform convergence of $\alpha_t$ to $g$.}
Taking the supremum over $x\in\mathcal{X}$ in Step~1,
\[
\sup_{x\in\mathcal{X}}
\bigl|\alpha_t(x) - g(x)\bigr|
\;\le\;
\sup_{x\in\mathcal{X}}\Delta_t(x)
\;+\;
\Bigl(\sup_{t}\kappa_t\Bigr)
\sup_{x\in\mathcal{X}} U_t(x).
\]
The first term vanishes by condition~(i), and the second term vanishes 
because $\sup_t\kappa_t < \infty$ by condition~(iii) and $\sup_x U_t(x)\to 0$ 
by condition~(ii). Therefore
$\sup_{x\in\mathcal{X}}|\alpha_t(x)-g(x)|\to 0$.

\emph{Step 3: Convergence of near minimizers.}
Let $\{x_t\}$ satisfy $\alpha_t(x_t)\le\inf_x\alpha_t(x)+\varepsilon_t$
with $\varepsilon_t\to 0$. Since $\mathcal{X}$ is compact, every subsequence 
of $\{x_t\}$ has a further convergent subsequence; let $x_{t_k}\to\bar{x}$ be 
any such convergent subsequence. We show 
$\bar{x}\in\arg\min_{x\in\mathcal{X}}g(x)$.

Let $x^*\in\arg\min_{x\in\mathcal{X}}g(x)$ be any minimizer of $g$, which 
exists by continuity of $g$ and compactness of $\mathcal{X}$. Using the near 
optimality condition and Step~2,
\begin{align*}
\limsup_{k\to\infty} g(x_{t_k})
&\;\le\;
\limsup_{k\to\infty}
\Bigl(
\alpha_{t_k}(x_{t_k})
+
\bigl|\alpha_{t_k}(x_{t_k}) - g(x_{t_k})\bigr|
\Bigr) \\
&\;\le\;
\limsup_{k\to\infty}
\Bigl(
\inf_x \alpha_{t_k}(x) + \varepsilon_{t_k}
+
\|\alpha_{t_k} - g\|_\infty
\Bigr).
\end{align*}
Since $\inf_x\alpha_{t_k}(x)\le\alpha_{t_k}(x^*)\le g(x^*) + 
\|\alpha_{t_k}-g\|_\infty$, we have
\[
\limsup_{k\to\infty} g(x_{t_k})
\;\le\;
g(x^*)
+
\limsup_{k\to\infty}
\Bigl(
\varepsilon_{t_k} + 2\|\alpha_{t_k} - g\|_\infty
\Bigr)
\;=\;
\inf_{x\in\mathcal{X}} g(x),
\]
where the equality uses $\varepsilon_{t_k}\to 0$ and 
$\|\alpha_{t_k}-g\|_\infty\to 0$ from Step~2. By continuity of $g$,
\[
g(\bar{x})
\;=\;
\lim_{k\to\infty} g(x_{t_k})
\;\le\;
\inf_{x\in\mathcal{X}} g(x).
\]
Since $g(\bar{x})\ge\inf_x g(x)$ trivially, we conclude 
$g(\bar{x})=\inf_x g(x)$, so $\bar{x}\in\arg\min_x g(x)$. As $\bar{x}$ was an 
arbitrary limit point of $\{x_t\}$, the result follows.
\end{proof}

\section{}
\label{app:discretization}

\textbf{Proposition~\ref{prop:discretization}} (Discretization Error for the 
Supremum).
\textit{Let $\psi:\mathcal{X}\times\Lambda\to\mathbb{R}$ admit a modulus of 
continuity $\omega$ in $\lambda$ uniformly in $x$, as 
in~\eqref{eq:modulus-continuity}, and let $\psi^{*}(x)$ and $\psi_T^{*}(x)$ be 
the continuous and discrete suprema defined in~\eqref{eq:psi-suprema}, with 
fill distance $h_T = \sup_{\lambda\in\Lambda} \min_{1\le t\le T}|\lambda - 
\lambda_t|$. Then for every $x\in\mathcal{X}$,}
\[
0 \;\le\; \psi^{*}(x) - \psi_T^{*}(x) \;\le\; \omega(h_T).
\]
\textit{If $\psi(x,\lambda)$ is $L$ Lipschitz in $\lambda$ uniformly in $x$, 
then $\psi^{*}(x) - \psi_T^{*}(x) \le L\,h_T$, and 
$\psi_T^{*}(x) \to \psi^{*}(x)$ uniformly in $x$ as $h_T \to 0$.}

In this appendix, we prove Proposition~\ref{prop:discretization}.

\begin{proof}
Fix $x\in\mathcal{X}$ throughout.

\emph{Step 1: Lower bound $\psi^{*}(x) - \psi_T^{*}(x) \ge 0$.}
Since $\{\lambda_1,\dots,\lambda_T\}\subset\Lambda$, the discrete maximum is 
taken over a subset of the domain over which the continuous supremum is 
taken. Therefore $\psi_T^{*}(x) \le \psi^{*}(x)$, giving 
$\psi^{*}(x) - \psi_T^{*}(x) \ge 0$.

\emph{Step 2: Upper bound $\psi^{*}(x) - \psi_T^{*}(x) \le \omega(h_T)$.}
For any $\lambda\in\Lambda$, let $t(\lambda) \in \{1,\dots,T\}$ be the index 
of the nearest grid point, so that $|\lambda - \lambda_{t(\lambda)}| \le h_T$ 
by definition of the fill distance. Applying the modulus of continuity 
assumption~\eqref{eq:modulus-continuity},
\[
\psi(x,\lambda)
\;\le\;
\psi\!\left(x,\lambda_{t(\lambda)}\right) 
+ \omega\!\left(|\lambda - \lambda_{t(\lambda)}|\right)
\;\le\;
\psi_T^{*}(x) + \omega(h_T),
\]
where the second inequality uses 
$\psi(x,\lambda_{t(\lambda)}) \le \psi_T^{*}(x)$ and 
$|\lambda - \lambda_{t(\lambda)}| \le h_T$ together with the fact that 
$\omega$ is nondecreasing. Since this bound holds for every 
$\lambda\in\Lambda$, taking the supremum over $\lambda$ gives
\[
\psi^{*}(x) 
\;=\; 
\sup_{\lambda\in\Lambda} \psi(x,\lambda) 
\;\le\; 
\psi_T^{*}(x) + \omega(h_T),
\]
and therefore $\psi^{*}(x) - \psi_T^{*}(x) \le \omega(h_T)$. Combining with 
Step~1 gives $0 \le \psi^{*}(x) - \psi_T^{*}(x) \le \omega(h_T)$.

\emph{Step 3: Lipschitz specialization.}
If $\psi(x,\lambda)$ is $L$ Lipschitz in $\lambda$ uniformly in $x$, then 
the modulus of continuity can be taken as $\omega(r) = Lr$, which satisfies 
the required conditions. Substituting into the bound from Step~2 gives 
$\psi^{*}(x) - \psi_T^{*}(x) \le L\,h_T$, and the convergence 
$\psi_T^{*}(x) \to \psi^{*}(x)$ as $h_T\to 0$ is uniform in 
$x\in\mathcal{X}$ since the bound does not depend on $x$.
\end{proof}

\end{document}